\documentclass[a4paper,fleqn]{cas-sc}
\usepackage[numbers,sort&compress]{natbib}
\usepackage{float}
\usepackage{bm}
\usepackage{multirow}
\usepackage{amsmath,amssymb,amsfonts}
\usepackage{amsthm}
\usepackage{textcomp}
\usepackage{manyfoot}
\usepackage{booktabs}
\usepackage{algorithm}
\usepackage{algorithmicx}
\usepackage{algpseudocode}
\usepackage{listings}
\usepackage{placeins}

\newtheorem{theorem}{Theorem}[section]
\newtheorem{proposition}[theorem]{Proposition}
\newtheorem{corollary}[theorem]{Corollary}
\newtheorem{assumption}[theorem]{Assumption}
\theoremstyle{remark}

\ExplSyntaxOn
\RenewDocumentCommand \printorcid { }
  {
    \seq_if_empty:NF \g_stm_orcid_seq
      {
        \group_begin:
        \tex_let:D \thefootnote \relax \footnotetext
          {
            \raggedright
            \textsc{orcid}(s):\c_space_token
            \seq_use:Nn \g_stm_orcid_seq { ;~ }
          }
        \group_end:
      }
    \group_begin:
    \tex_let:D \thefootnote \relax \footnotetext
      { \raggedright Public~code:\c_space_token \url{https://github.com/liangzhangyong/CPDEM.git} }
    \group_end:
  }

\ExplSyntaxOff

\begin{document}
\let\WriteBookmarks\relax
\def\floatpagepagefraction{1}
\def\textpagefraction{.001}
\shorttitle{CPDEM for random material parameters}
\shortauthors{Z.Liang, J.Zhu, H.Gao}

\title[mode=title]{Constitutive parameterized deep energy method for solid mechanics problems with random material parameters}
\author[1]{Zhangyong Liang}
\credit{Conceptualization, Methodology, Software, Validation, Formal analysis, Investigation, Visualization, Writing -- original draft}
\author[2]{Jihong Zhu}
\credit{Conceptualization, Methodology, Supervision, Investigation, Formal analysis, Writing -- original draft}
\author[3]{Guikai Guo}
\credit{Conceptualization, Methodology, Supervision, Formal analysis, Visualization, Writing -- original draft}
\author[3]{Huanhuan Gao}[orcid=0000-0003-4463-6433]
\cormark[1]
\ead{gao_huanhuan@jlu.edu.cn}
\credit{Conceptualization, Methodology, Supervision, Project administration, Funding acquisition, Writing -- review \& editing}
\affiliation[1]{organization={National Center for Applied Mathematics, Tianjin University},
                city={Tianjin},
                postcode={300072},
                country={PR China}}

\affiliation[2]{organization={School of Mechanical Engineering, Northwestern Polytechnical
University},
                city={Xi’an},
                postcode={710072},
                country={China}}
                
\affiliation[3]{organization={School of Mechanical and Aerospace Engineering, Jilin University},
                city={Changchun},
                postcode={130025},
                country={China}}
\cortext[cor1]{Corresponding author}

\begin{abstract}
Material properties vary across manufacturing batches and operating conditions, creating a need for repeated constitutive-parameter queries in solid mechanics.
Such queries generally require material-dependent assembly and new finite-element method (FEM) solves, while fixed-parameter deep energy methods must be retrained from scratch whenever material parameters change.
To address this issue, we propose the \textbf{C}onstitutive \textbf{P}arameterized \textbf{D}eep \textbf{E}nergy \textbf{M}ethod (\textbf{CPDEM}) to learn the material parameter-to-solution map over continuous property ranges across static, finite-strain, and history-dependent solid-mechanics problems.
Coordinate and material encoders condition a shared displacement trial field, trained by minimizing the expected potential energy or incremental potential without labeled displacement solutions.
The pretrained model supports direct prediction at unseen in-range parameters without parameter-specific retraining.
For history-dependent responses, material-shared incremental models predict the displacement history while parameter-specific constitutive updates advance internal states.
For uniformly stable linear-elastic families, the product-space analysis establishes parameter-to-solution continuity, population-energy consistency, and equivalence between the population-energy gap and the squared parameter-averaged mechanical error.
Finite-training estimates account for approximation, sampling, quadrature, boundary enforcement, and optimization errors.
The theoretical framework further covers finite-strain hyperelasticity and constitutive-state error propagation in elastoplastic and viscoelastic updates.
Numerical examples span linear elasticity, finite-strain hyperelasticity, cyclic elastoplasticity, and viscoelastic creep and relaxation.
Comparisons with FEM reproduce displacement patterns and capture cyclic displacement reversals, plastic-strain accumulation, stress evolution, and material-dependent viscoelastic responses in the benchmarks.
These results support repeated in-range material queries through a shared constitutive-parameterized variational framework.
\end{abstract}

\begin{highlights}
\item CPDEM learns material parameter-to-solution maps by minimizing expected energy.
\item Energy gaps bound parameter-averaged errors in stable linear-elastic families.
\item Pretrained models predict unseen in-range material responses without retraining.
\item Benchmarks span elasticity, hyperelasticity, elastoplasticity, and viscoelasticity.
\end{highlights}


\maketitle
\section{Introduction}
\label{introduction}

Material properties used in structural analysis rarely coincide exactly with their nominal values.
Manufacturing tolerances, batch-to-batch variability, calibration updates, and changing operating conditions generate random variations in constitutive parameters, including Young's modulus, Poisson's ratio, hardening modulus, and viscosity.
For a given structural configuration and loading scenario, these constitutive variations define a parameterized family of closely related solid mechanics problems that require repeated mechanical evaluations.
The finite-element method (FEM) remains a principal numerical tool for such problems \cite{quarteroni2008,ames2014}.
Finite-volume and meshless formulations provide related discretization routes \cite{HE2025118264}, and the lattice-Boltzmann method offers another numerical strategy \cite{CHEN2025113831}.
Material and geometric nonlinearities can make each analysis expensive \cite{ZHOU2025104267,HeShuai}.
Contact and friction introduce further nonlinear iterations \cite{HE2025110666}, fluid--structure interaction requires coupled solution procedures \cite{VIVALDI2025114121, LEROGERON2026116274}, and multiphysics problems bring related numerical demands \cite{WU2025125513}.
A fixed mesh can generally be reused for a fixed geometry, but each material parameter set still requires assembly of the material-dependent weak form and solution of the associated equilibrium or incremental constitutive problem.
Classical stochastic finite-element methods propagate random fields or random variables through the mechanical model \cite{ghanem1991stochastic,stefanou2009review}.
Polynomial-chaos formulations represent parameter dependence through sampling or spectral expansions \cite{xiu2002wiener,blatman2010adaptive}, while multilevel and multifidelity methods combine models at different resolutions to reduce repeated computational effort \cite{giles2008mlmc,peherstorfer2018multifidelity}.
Surrogate-based reliability methods construct response surfaces for declared limit-state calculations \cite{echard2011akmcs}.
The need to evaluate solid mechanics responses across continuous material parameter ranges motivates the use of a shared physics-driven surrogate.

Physics-driven neural solvers offer a complementary route because the mechanical field can be determined from governing equations or variational principles without first generating labeled displacement solutions.
Physics-informed neural networks (PINNs) enforce strong-form residuals together with initial or boundary residuals at collocation points \cite{raissi2019phy,karniadakis2021phy, WOS:000784275100002}.
The Deep Ritz method represents the governing problem through minimization of a variational functional \cite{weinan2018deep}, and deep energy methods (DEM) use a variational functional to approximate a mechanical field \cite{samaniego2020energy,nguyen2020deep,nguyen2021}.
Conventional PINNs minimize strong-form PDE and boundary residuals \cite{sirignano2018dgm,raissi2019phy}, whereas DEM minimizes a first-derivative potential and enforces Dirichlet conditions via trial fields or penalties.
Physics-driven learning has been applied to linear elasticity and finite-strain hyperelasticity \cite{haghighat2021phy,abueidda2020}, as well as mixed stress-displacement formulations and inelastic constitutive behavior \cite{rao2020physics,fuhg2022mixed}.
PINN studies include small-strain elasticity and elastoplasticity \cite{haghighat2021phy}, coupled field equations and mixed displacement--stress settings \cite{rao2020physics}, and residual-based treatments of linear elasticity, hyperelasticity, and plasticity \cite{abueidda2020}.
Subsequent DEM work includes combined strong- and weak-form treatments \cite{abueidda2023}, systematic hyperparameter selection \cite{chadha2022}, adversarial multiphysics formulations \cite{lee2024adv}, and comparisons between residual- and energy-based approaches \cite{li2021phy}.
Related DEM formulations address elastoplasticity \cite{he2023elastoplasticity} as well as hyperelasticity and viscoelasticity \cite{abueidda2022viscoelasticity}.
In conventional fixed-parameter implementations, the material constants enter the energy or residual as fixed quantities, so a material perturbation defines a new training problem.
P$^2$INN addresses the repeated-training issue for residual-based parameterized PDEs by encoding PDE parameters into latent features and representing an entire PDE family with one network \cite{cho2024p2inn}.
The remaining mechanics-specific question is how to formulate the parameterization when material parameters define a constitutive energy, a material tangent, and, for inelastic materials, the evolution of internal states.

Beyond purely physics-driven solvers, data-driven and hybrid approaches learn mechanical responses from data.
Deep learning has entered computational mechanics through physical information, experimental observations, and simulation data \cite{WOS:000355286600030, WOS:001398286100002}.
Applications include nonlinear mechanical behavior \cite{nguyen2020deep,WOS:000784275100002}, mechanical dynamics \cite{GOYAL2024134158}, fluid mechanics \cite{WOS:001031995000007}, and heat transfer \cite{WOS:000877762800009}.
These studies are often grouped as data-driven, physics-driven, or hybrid data--physics-driven formulations \cite{WOS:000662865400006,raissi2019phy,karniadakis2021phy}.
Hybrid schemes place observations and governing relations in the same learning procedure \cite{WOS:001324824300011}.
A phase-space nearest-point projection can be constructed directly from material data without an explicit constitutive model \cite{kirchdoerfer2016data,eggersmann2019model}.
Dense and convolutional neural networks instead learn input--output relations from labelled examples \cite{WOS:001243991400001, WOS:000462895900002}.
Other hybrid approaches combine data fitting with mechanical constraints in a jointly defined optimization problem \cite{WOS:001649775500006, WOS:001104934700004}.
Neural operators address the many-query setting by learning solution maps over families of PDE inputs.
DeepONet combines branch and trunk representations for input functions and spatial query coordinates \cite{lu2021learn}, while Fourier neural operators use a global integral-kernel map on a chosen discretization \cite{li2020fourier}.
In their standard supervised forms, these approaches are trained from paired input and solution fields and require high-fidelity data to construct a pretrained surrogate.
Their function-space formulation is intentionally broad and is well suited to spatially varying coefficients, loads, initial conditions, and other high-dimensional inputs.
In contrast, constitutive parameterization targets the finite-dimensional material space via variational energy.

Physics-informed operator methods reduce the dependence on paired solution fields by incorporating governing equations directly into operator training \cite{wang2021learn,li2024pino}.
Along the same lines, variational physics-informed operator formulations show that a discretized energy functional can also serve as the training objective \cite{eshaghi2025vino}, while finite-operator learning further connects neural parameter-to-solution maps with discrete weak forms \cite{rezaei2025fol}.
In addition, parametric Ritz formulations minimize variational energies without requiring labeled solution fields \cite{fuhg2023dcrm,kaltenbach2026parametric}.
Moreover, transfer-learning and physics-informed operator models offer additional routes to handling repeated parameter queries \cite{desai2022one,wang2021learn,li2024pino}.
A related parametric deep-energy formulation has also been presented for elasticity with strain-gradient effects \cite{nguyen2021}.
Parameter conditioning supports repeated queries within a shared model; physics supervision and variational training provide routes to learning that model from governing equations or energies.
In general, their scope is a reusable solution operator for generic PDE data, which may include material properties along with geometry, loads, and boundary information.
In solid mechanics, however, each material parameter set indexes a mechanical energy or incremental potential and determines the equilibrium minimizers and local constitutive state updates.

Motivated by this variational structure, we propose the \textbf{C}onstitutive \textbf{P}arameterized \textbf{D}eep \textbf{E}nergy \textbf{M}ethod (\textbf{CPDEM}), in which the constitutive vector parameterizes a family of variational problems and their associated mechanical minimizers.
Specifically, CPDEM represents this family with a shared displacement trial field that is conditioned on both spatial coordinates and constitutive parameters.
The population objective averages the material-dependent potential energy (or the incremental potential) over spatial quadrature points and parameter samples.
For elastoplasticity and viscoelasticity, the formulation relies on the appropriate incremental potential and propagates the physically defined internal variables between increments.
Meanwhile, the random-parameter domain specifies the seen parameter sets used for parameterized pretraining, and the uniform sampling rule assigns numerical coverage over that domain.
Finally, the numerical study evaluates unseen in-range material parameter sets across linear-elastic, hyperelastic, elastoplastic, and viscoelastic benchmarks.

The main contributions of this work are summarized as follows:
\begin{itemize}
    \item \textbf{Constitutive-parameterized variational formulation.}
    We formulate CPDEM as a physics-driven approximation of the map $\boldsymbol\eta\mapsto\bm u^*(\cdot,\boldsymbol\eta)$ on the joint spatial-constitutive domain.
    The coordinates and constitutive parameters condition a shared neural trial field, and the training minimizes the expected potential energy, or the incremental potential for history-dependent problems.
    
    \item \textbf{Parametric well-posedness, stability, and variational consistency.}
    For uniformly stable linear-elastic families, we establish well-posedness and Lipschitz continuity of the constitutive parameter-to-solution map.
    We characterize the parameter-wise mechanical solution family as the minimizer of a population energy functional in a Bochner space over the material domain. This minimizer is unique up to sets of zero measure.
    
    \item \textbf{Nonlinear and history-dependent constitutive behavior.}
    We extend the parameter-conditioned construction to finite-strain compressible Neo-Hookean hyperelasticity, incremental small-strain associative $J_2$ elastoplasticity with linear isotropic hardening and radial return, and incremental small-strain standard-linear-solid viscoelasticity with a backward-Euler internal-variable update.
    The nonlinear estimate applies locally under stable energy growth, while the history-dependent statement accounts for step-to-step error propagation through the stored constitutive state across loading increments.
    
    \item \textbf{Parameterized pretraining and inference across constitutive regimes.}
    Parameterized pretraining over seen material parameter sets produces one shared model for direct inference at unseen in-range queries in place of parameter-specific retraining.
    The numerical study spans two- and three-dimensional elastic and hyperelastic problems together with incremental elastoplastic and viscoelastic benchmarks.
\end{itemize}

The remainder of this paper is organized as follows:
Section~\ref{problem} presents the random material parameters and constitutive formulations.
Section~\ref{methodology} introduces CPDEM, the discrete expected-energy objective, and the material and spatial encoders.
Section~\ref{theoretical-analysis} develops the product-space formulation, the linear-elastic stability results, and the finite-training error structure.
Section~\ref{numerical-examples} presents the numerical examples and reference comparisons.
Section~\ref{conclusions} summarizes the main findings and concludes the paper.

\section{Problem formulation}
\label{problem}

\subsection{Random material parameters}

\begin{figure}[pos=htbp]
    \centering
    \includegraphics[width=\linewidth]{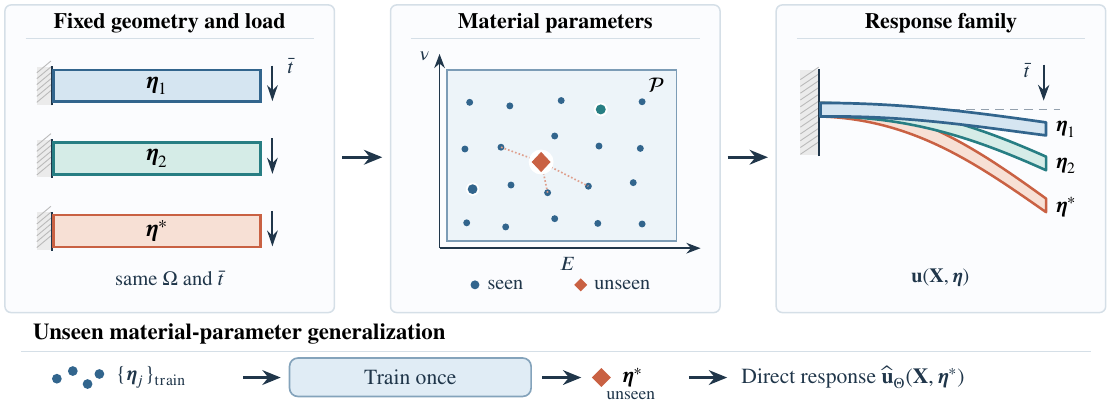}
    \par\vspace{-2pt}
    \caption{Schematic of material-parameter sampling and evaluation at unseen in-range parameters $(E,\nu)$ for a fixed mechanical setting.
    Circles indicate training samples; the diamond denotes an unseen query.}
    \label{fig:random-material-parameter-family}
\end{figure}

We consider a family of solid-mechanics problems with varying random constitutive parameters on a fixed reference domain $\Omega\subset\mathbb R^d$, under fixed boundary conditions and loading.
Fig.~\ref{fig:random-material-parameter-family} illustrates material-parameter sampling and an unseen in-range query for a fixed mechanical setting.

For each material model, the parameters are collected in a finite-dimensional vector
\begin{equation*}
    \boldsymbol{\eta}=[\eta_1,\eta_2,\ldots,\eta_{d_\eta}]^\top
    \in\mathcal P\subseteq
    \prod_{i=1}^{d_\eta}[\eta_{i,\min},\eta_{i,\max}],
\end{equation*}
where $\mathcal P$ is the admissible material-parameter domain.
Each parameter vector specifies spatially uniform material properties throughout $\Omega$, and parameter variations define different mechanical problems.

\begin{figure}[pos=htbp]
    \centering
    \includegraphics[width=.9\linewidth]{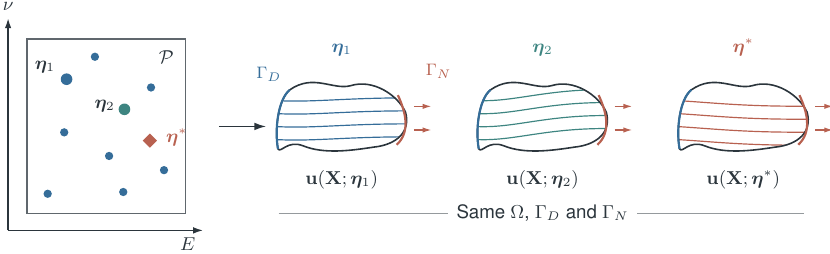}
    \par\vspace{-2pt}
    \caption{Schematic of the mechanical response family under varying material parameters.
    Vectors $\boldsymbol\eta_1$, $\boldsymbol\eta_2$, and $\boldsymbol\eta^*$ specify spatially uniform properties on the same domain $\Omega$ with fixed loading and boundaries $\Gamma_D$ and $\Gamma_N$.}
    \label{fig:material-response-boundary-family}
\end{figure}

Let $\Gamma_D$ and $\Gamma_N$ denote the fixed Dirichlet and Neumann parts of the boundary, and let $\mathcal U_D$ be the set of kinematically admissible displacement fields satisfying the imposed displacement on $\Gamma_D$.
For a static problem admitting an energy minimizer, the potential energy and an associated solution are
\begin{equation*}
\begin{aligned}
    \Pi_{\boldsymbol\eta}(\bm v)
      &=\int_\Omega \Psi\bigl(\mathcal K(\bm v);\boldsymbol\eta\bigr)
        \,\mathrm d\Omega-\mathcal W(\bm v),\\
    \bm u^*(\cdot,\boldsymbol\eta)
      &=\underset{\bm v\in\mathcal U_D}{\arg\min}\,
        \Pi_{\boldsymbol\eta}(\bm v),
\end{aligned}
\end{equation*}
where $\Psi$ is the constitutive energy density, $\mathcal W$ is the external work, and $\mathcal K(\bm v)$ is the kinematic measure, given by $\bm\varepsilon(\bm v)$ for small-strain elasticity or $\bm F(\bm v)=\bm I+\operatorname{Grad}\bm v$ for finite-strain hyperelasticity.
For finite-strain problems, $\mathcal U_D$ also includes constitutive admissibility conditions such as $\det\bm F>0$.
As illustrated in Fig.~\ref{fig:material-response-boundary-family}, the dependence of $\Pi_{\boldsymbol\eta}$ on the material vector generates a family of displacement fields on the same domain $\Omega$ under fixed boundary conditions and loading.
A parameterized displacement field is represented in Eq.~\eqref{eq:parametric-displacement-field}:
\begin{equation}
    \label{eq:parametric-displacement-field}
    \bm u:\Omega\times\mathcal P\longrightarrow\mathbb R^d,
    \qquad (\bm X,\boldsymbol\eta)\longmapsto
    \bm u(\bm X,\boldsymbol\eta).
\end{equation}
The strain and stress fields depend on $\boldsymbol\eta$ through the displacement field and the constitutive law.

Let $p(\boldsymbol\eta)$ be a normalized sampling density on $\mathcal P$, with $\int_{\mathcal P}p(\boldsymbol\eta)\,\mathrm d\boldsymbol\eta=1$.
The benchmarks use uniform sampling over the admissible parameter domain, with $p$ specifying the numerical sampling measure.
The expected potential energy of the static problem family is defined in Eq.~\eqref{eq:expected-potential-energy}:
\begin{equation}
    \label{eq:expected-potential-energy}
    \begin{aligned}
    \mathcal J(\bm u)
      =\mathbb E_{\boldsymbol\eta\sim p}
        \!\left[\Pi_{\boldsymbol\eta}\bigl(\bm u(\cdot,\boldsymbol\eta)\bigr)\right]=\int_{\mathcal P}
        \Pi_{\boldsymbol\eta}\bigl(\bm u(\cdot,\boldsymbol\eta)\bigr)
        p(\boldsymbol\eta)\,\mathrm d\boldsymbol\eta.
    \end{aligned}
\end{equation}
The objective averages the energies evaluated for complete spatial displacement fields at each material parameter.
For elastoplasticity and viscoelasticity, the incremental potential and parameter-specific internal state define the energy problem at each increment, as detailed in the constitutive formulations below.

Here, an unseen in-range parameter vector $\boldsymbol\eta^*\in\mathcal P$ is one used in neither training nor model selection; direct prediction at $\boldsymbol\eta^*$ evaluates $\widehat{\bm u}_{\Theta^*}(\bm X,\boldsymbol\eta^*)$ with the trained weights $\Theta^*$ held fixed (written compactly as $\Theta$ in Fig.~\ref{fig:random-material-parameter-family}).

\subsection{Constitutive formulation}
Below we state the four constitutive formulations considered in this study: small-strain linear elasticity, finite-deformation hyperelasticity, and incremental small-strain $J_2$ elastoplasticity and viscoelasticity.

\textbf{Linear elasticity problem}
In linear elasticity, the variational minimization of the total potential energy $\mathcal{E}(\bm{u})$ takes the form
\begin{equation}
    \label{eq:linear-elastic-bvp}
    \begin{split}
        \text{Minimize:}\;\;\;\; \mathcal{E}(\bm{u}) &= \int_{\Omega} \Psi(\bm{\varepsilon}) d\Omega - \int_{\Omega} \bm{f}_b \cdot \bm{u} d\Omega - \int_{\Gamma_N} \bar{\bm{t}} \cdot \bm{u} d\Gamma,\\
        \text{where:}\;\;\;\; \Psi(\bm{\varepsilon}) &= \frac{1}{2}\bm{\varepsilon}:\mathbb{C}:\bm{\varepsilon},\\
        \text{subject to:}\;\;\;\; \bm{u} &= \bar{\bm{u}} \text{ on } \Gamma_D,\\
        \text{and}\;\;\;\; \bm{\sigma}\cdot \bm{n} &= \bar{\bm{t}} \text{ on } \Gamma_N \\
    \end{split},
\end{equation}
where $\Psi(\bm{\varepsilon})$ is the elastic strain-energy density, $\bm{\varepsilon}(\bm{u})$ is the small-strain tensor, $\mathbb{C}$ is the fourth-order elasticity tensor, $\bm{f}_b$ is the body force, $\bm{\sigma}$ is the Cauchy stress tensor, and $\bm{n}$ is the outward unit normal on $\Gamma_N$.
Here, $\bar{\bm{u}}$ denotes the imposed displacement on the Dirichlet boundary $\Gamma_D$, and $\bar{\bm{t}}$ denotes the applied traction on the Neumann boundary $\Gamma_N$.

\textbf{Hyperelasticity problem}
Under finite deformation, the motion is described by the deformation mapping $\bm{\varphi}(\bm{X}):=\bm{x}=\bm{X}+\bm{u}$ with deformation gradient $\bm{F}=\mathrm{Grad}\,\bm{\varphi}(\bm{X})$, and the first Piola--Kirchhoff stress follows from $\bm{P}=\partial\Psi/\partial\bm{F}$.
The hyperelastic benchmarks adopt the compressible Neo-Hookean strain-energy density
\begin{equation*} 
    \Psi(I_1,J)=\frac{1}{2}\lambda[\ln(J)]^2-\mu\ln(J)+\frac{1}{2}\mu(I_1-3),
\end{equation*}
where $I_1=\operatorname{tr}(\bm{F}^\top\bm{F})$ is the first invariant of the right Cauchy--Green tensor, and the Jacobian determinant $J=\det\bm{F}>0$ excludes orientation-reversing deformations.
The Lam\'e parameters $\lambda$ and $\mu$ are related to the material vector $\boldsymbol{\eta}=[E,\nu]^\top$ by
\begin{equation}
    \label{eq:lame-parameters}
    \begin{split}
        \lambda &= \frac{E\nu}{(1+\nu)(1-2\nu)} \\
        \mu &= \frac{E}{2(1+\nu)},
    \end{split}
\end{equation}
with $E$ and $\nu$ denoting Young's modulus and Poisson's ratio, respectively.

The total potential energy reads
\begin{equation}
    \label{eq:hyperelastic-potential}
    \mathcal{L}(\bm{\varphi}(\bm{X}))=\int_\Omega\Psi(\bm{F})d\Omega-\int_\Omega \bm{f}_b \cdot\bm{\varphi}(\bm{X}) d\Omega-\int_{\Gamma_N} \bar{\bm{t}}\cdot\bm{\varphi}(\bm{X})d\Gamma,
\end{equation}
where the external-work integrals in the reference configuration are understood as work-conjugate to the displacement $\bm{u}=\bm{\varphi}-\bm{X}$.
With imposed boundary deformation $\bar{\bm{\varphi}}(\bm{X})=\bm{X}+\bar{\bm{u}}(\bm{X})$ on $\Gamma_D$, the equilibrium configuration is obtained by minimizing $\mathcal{L}$,
\begin{equation}
    \label{eq:hyperelastic-minimization}
    \begin{split}
        \text{min}:\hspace{5mm}&\mathcal{L}(\bm{\varphi}(\bm{X}))\\
        \text{s.t.}: \hspace{5mm}&\bm{\varphi}(\bm{X}) = \bar{\bm{\varphi}}(\bm{X}) \hspace{7mm} \text{on}\ \Gamma_D \\
    \end{split}.
\end{equation}

\textbf{Elastoplasticity problem}
For the $J_2$ elastoplastic formulation, we use a small-strain BVP solved incrementally over load steps.
The constitutive response depends on the history variables $\bm{\alpha}_i = [\bm{\varepsilon}_i^p,\bar{\varepsilon}_i^p]$, where $\bm{\varepsilon}^p$ is the plastic strain tensor and $\bar{\varepsilon}^p$ is the equivalent plastic strain.
In this work, associative $J_2$ plasticity with linear isotropic hardening is considered, and the constitutive update at each loading increment is performed by radial return.
Given the committed state $[\bm{\sigma}_i,\bm{\varepsilon}_i^p,\bar{\varepsilon}_i^p]$ and the total strain increment $\Delta\bm{\varepsilon}$, the elastic trial stress and trial yield function are
\begin{equation}
    \label{eq:trial-state}
    \begin{aligned}
        \bm{\sigma}_{\mathrm{trial}} &= \bm{\sigma}_i + \mathbb C:\Delta\bm{\varepsilon}, \\
        \bm{\sigma}'_{\mathrm{trial}} &= \operatorname{dev}\bm{\sigma}_{\mathrm{trial}}, \\
        f_{\mathrm{trial}} &= \|\bm{\sigma}'_{\mathrm{trial}}\|_F - \sqrt{\frac{2}{3}}\,\sigma_y(\bar{\varepsilon}_i^p),
    \end{aligned}
\end{equation}
where the prime denotes the deviatoric part,
\begin{equation}
    \bm A'=\operatorname{dev}\bm A
    =\bm A-\frac{1}{3}\operatorname{tr}(\bm A)\bm I,
    \label{eq:plastic-deviatoric-part}
\end{equation}
and $\|\bm A\|_F=\sqrt{\bm A:\bm A}$ is the Frobenius norm.
Linear isotropic hardening is defined by
\begin{equation}
    \sigma_y(\bar{\varepsilon}^p)=\sigma_{y,0}+\kappa\bar{\varepsilon}^p,
    \label{eq:plastic-hardening-law}
\end{equation}
where $\sigma_{y,0}$ is the initial yield stress and $\kappa$ is the isotropic hardening modulus.

If $f_{\mathrm{trial}}\leq0$, the increment is elastic:
\begin{equation}
    \bm{\sigma}_{i+1}=\bm{\sigma}_{\mathrm{trial}},\qquad
    \bm{\varepsilon}_{i+1}^p=\bm{\varepsilon}_i^p,\qquad
    \bar{\varepsilon}_{i+1}^p=\bar{\varepsilon}_i^p.
    \label{eq:plastic-elastic-branch}
\end{equation}
If $f_{\mathrm{trial}}>0$, radial return gives
\begin{equation}
    \begin{aligned}
        \bm N_{\mathrm{trial}}
        &=\frac{\bm{\sigma}'_{\mathrm{trial}}}{\|\bm{\sigma}'_{\mathrm{trial}}\|_F},
        \qquad \Delta\gamma=\frac{f_{\mathrm{trial}}}{2\mu+\frac{2}{3}\kappa},\\
        \bm{\varepsilon}_{i+1}^p
        &=\bm{\varepsilon}_i^p+\Delta\gamma\,\bm N_{\mathrm{trial}},\\
        \bar{\varepsilon}_{i+1}^p
        &=\bar{\varepsilon}_i^p+\sqrt{\frac{2}{3}}\,\Delta\gamma,\\
        \bm{\sigma}_{i+1}
        &=\bm{\sigma}_{\mathrm{trial}}-2\mu\Delta\gamma\,\bm N_{\mathrm{trial}}.
    \end{aligned}
    \label{eq:plastic-radial-return}
\end{equation}
The plastic multiplier increment $\Delta\gamma$ is normalized such that
$\Delta\bm{\varepsilon}^p=\Delta\gamma\,\bm N_{\mathrm{trial}}$, where
$\|\bm N_{\mathrm{trial}}\|_F=1$ and $\mu$ denotes the elastic shear modulus of the isotropic medium.

In the variational formulation, the displacement $\bm u_{i+1}$ serves as the global unknown while the plastic state is eliminated locally at each quadrature point via the radial-return map $\bm\alpha_{i+1}=\mathcal R(\bm\varepsilon(\bm u_{i+1}),\bm\alpha_i;E,\nu,\kappa)$ in Eqs.~\eqref{eq:plastic-elastic-branch}--\eqref{eq:plastic-radial-return}, which enforces the discrete yield and consistency conditions across both elastic and plastic branches.
The reduced incremental potential at load step $i+1$ is
\begin{equation}
    \label{eq:elastoplastic-free-energy}
    \begin{split}
        \mathcal{L}_{ep}^{i+1}(\bm u_{i+1};\bm\alpha_i) = \int_{\Omega} \left[\frac{1}{2}(\bm{\varepsilon}_{i+1}-\bm{\varepsilon}_{i+1}^{p}):\mathbb{C}:(\bm{\varepsilon}_{i+1}-\bm{\varepsilon}_{i+1}^{p}) + \frac{1}{2}\kappa(\bar{\varepsilon}_{i+1}^{p})^2 + \sigma_{y,0}(\bar{\varepsilon}_{i+1}^{p}-\bar{\varepsilon}_{i}^{p})\right]d\Omega + \mathcal{P}_{ext}(\bm u_{i+1}),
    \end{split}
\end{equation}
with the external potential
\begin{equation}
    \label{eq:external-potential}
    \mathcal{P}_{ext}(\bm u_{i+1}) = -\int_{\Omega}\bm f_b^{i+1}\cdot\bm u_{i+1}\,d\Omega-\int_{\Gamma_N}\bar{\bm t}^{i+1}\cdot\bm u_{i+1}\,d\Gamma.
\end{equation}
Evaluating the local return map prior to differentiating $\mathcal{L}_{ep}^{i+1}$ with respect to the displacement field yields the incremental minimization problem
\begin{equation}
    \label{eq:elastoplastic-minimization}
    \bm u_{i+1}^* = \underset{\bm u\in\mathcal U(\bar{\bm u}_{i+1})}{\operatorname{argmin}}\;\mathcal{L}_{ep}^{i+1}(\bm u;\bm\alpha_i), \bm{\varphi}_{i+1}(\bm{X}) = \bar{\bm{\varphi}}(\bm{X}) \hspace{7mm} \text{on}\ \Gamma_D,
\end{equation}
where $\mathcal U(\bar{\bm u})=\{\bm v\in H^1(\Omega;\mathbb R^d):\bm v=\bar{\bm u}\text{ on }\Gamma_D\}$.
After each global minimization, the updated state $[\bm\sigma_{i+1},\bm\varepsilon^p_{i+1},\bar\varepsilon^p_{i+1}]$ is committed and carried forward to the next load step.

\textbf{Viscoelasticity problem}
For the viscoelastic benchmarks, the rate-dependent response is described through an additive decomposition $\bm{\varepsilon} = \bm{\varepsilon}^{e} + \bm{\varepsilon}^{v}$ into the recoverable elastic strain $\bm{\varepsilon}^{e}$ and the viscous internal variable $\bm{\varepsilon}^{v}$, governed by a two-potential standard-linear-solid framework.
For an isotropic material, the free-energy density $\psi$ and dissipation potential $\phi$ are
\begin{equation}
    \label{eq:viscoelastic-potentials}
    \begin{aligned}
        \psi &= \frac{1}{2} \bm{\varepsilon} : \mathbb{L}^{\infty} : \bm{\varepsilon} + \frac{1}{2} (\bm{\varepsilon} - \bm{\varepsilon}^{v}) : \mathbb{L}^{1} : (\bm{\varepsilon} - \bm{\varepsilon}^{v}), \\
        \phi &= \frac{1}{2} \dot{\bm{\varepsilon}}^{v} : \mathbb{M} : \dot{\bm{\varepsilon}}^{v},
    \end{aligned}
\end{equation}
where $\mathbb{L}^{\infty}$, $\mathbb{L}^{1}$, and $\mathbb{M}$ denote the equilibrium, non-equilibrium, and viscosity tensors, respectively, and $\dot{\bm{\varepsilon}}^{v}$ is the viscous strain rate.
With common Poisson's ratio $\nu$, both elastic branches take the isotropic form $\mathbb L^r=\lambda_r\bm I\otimes\bm I+2\mu_r\mathbb I_{\mathrm{s}}$ ($r\in\{\infty,1\}$), where $\mathbb I_{\mathrm{s}}$ is the symmetric fourth-order identity, $\lambda_r=E_r\nu/[(1+\nu)(1-2\nu)]$, and $\mu_r=E_r/[2(1+\nu)]$, with equilibrium modulus $E_\infty$ and decaying-branch modulus $E_1$.
Setting $\mathbb M=\tau\mathbb L^1$ with viscosity $\xi$ and relaxation time $\tau=\xi/E_1$ enforces a single characteristic time in volumetric and deviatoric modes.



Discretizing the dissipation potential over the time step $\Delta t = t^{n+1} - t^{n}$ yields the incremental potential
\begin{equation}
    \label{eq:viscoelastic-incremental-potential}
    \mathcal{L}_{ve}^{n+1} = \int_{\Omega} \psi(\bm{\varepsilon}^{n+1}, \bm{\varepsilon}^{v,n+1}) \, d\Omega + \Delta t \int_{\Omega} \phi\!\left(\frac{\bm{\varepsilon}^{v,n+1}-\bm{\varepsilon}^{v,n}}{\Delta t}\right) d\Omega - \int_{\Omega}\bm f_b^{n+1}\cdot\bm u^{n+1}\,d\Omega-\int_{\Gamma_N} \bar{\bm{t}}^{n+1}\cdot\bm{u}^{n+1}\,d\Gamma,
\end{equation}
starting from the initial internal state $\bm\varepsilon^{v,0}=\bm 0$ unless another equilibrated state is specified.
Backward-Euler discretization of the viscous evolution gives $\mathbb{M}:(\bm{\varepsilon}^{v,n+1}-\bm{\varepsilon}^{v,n})/\Delta t=\mathbb{L}^{1}:(\bm{\varepsilon}^{n+1}-\bm{\varepsilon}^{v,n+1})$, or equivalently $[\mathbb M/\Delta t+\mathbb L^1]:\bm\varepsilon^{v,n+1}=\mathbb L^1:\bm\varepsilon^{n+1}+(\mathbb M/\Delta t):\bm\varepsilon^{v,n}$.
At each time step, $(\bm u^{n+1},\bm\varepsilon^{v,n+1})$ minimizes the incremental potential,
\begin{equation*}
    (\bm u^{n+1},\bm\varepsilon^{v,n+1})=\underset{\substack{\bm u\in\mathcal U(\bar{\bm u}^{n+1})\\ \bm\varepsilon^v\in L^2_{\mathrm{sym}}(\Omega)}}{\operatorname{argmin}}\;\mathcal{L}_{ve}^{n+1}(\bm u,\bm\varepsilon^v;\bm\varepsilon^{v,n}), \bm{\varphi}^{n+1}(\bm{X}) = \bar{\bm{\varphi}}(\bm{X}) \hspace{7mm} \text{on}\ \Gamma_D,
\end{equation*}
where $L^2_{\mathrm{sym}}(\Omega)$ denotes the space of square-integrable symmetric tensor fields, and the resulting internal state $\bm\varepsilon^{v,n+1}$ is passed to the next increment.

\section{Methodology}
\label{methodology}
\subsection{The deep energy method (DEM)}
The DEM \cite{samaniego2020energy} employs neural networks to approximate the displacement field by reformulating the governing problem within a variational framework.
This reformulation casts the learning task as an unconstrained optimization problem in which the potential energy serves as the loss function written as Eq.~\eqref{eq:dem-energy}:
\begin{equation}
    \label{eq:dem-energy}
    \underbrace{\mathcal{L}(\hat{\bm{u}};\bm{\theta})}_{\text{Potential energy}}
    = \underbrace{U(\hat{\bm{u}};\bm{\theta})}_{\text{Internal energy}}
    - \underbrace{W(\hat{\bm{u}};\bm{\theta})}_{\text{External energy}},
\end{equation}
where $\bm{\theta}$ denotes the vector of trainable neural-network parameters optimized using randomly sampled collocation points.
The term $U(\hat{\bm{u}};\bm{\theta})$ is the internal deformation energy, whereas $W(\hat{\bm{u}};\bm{\theta})$ represents the work performed by the applied body forces and surface tractions.

For the stable static problems considered here, the total potential energy is minimized with respect to the trainable network parameters $\bm{\theta}$, yielding Eq.~\eqref{eq:dem-optimization}:
\begin{equation}
    \label{eq:dem-optimization}
    \bm{\theta}^*=\underset{\bm{\theta}}{\text{argmin}}\mathcal{L}(\hat{\bm{u}};\bm{\theta}).
\end{equation}

The principal limitation of a conventional fixed-parameter DEM lies in its treatment of material parameters.
Properties such as Young's modulus $E$ and Poisson's ratio $\nu$ are embedded as constants in the internal-energy functional, and the trained network is associated with one specified material configuration. 

\subsection{Constitutive Parameterized Deep Energy Method (CPDEM)}
\begin{figure}[pos=htbp]
    \centering
    \includegraphics[width=1.0\columnwidth]{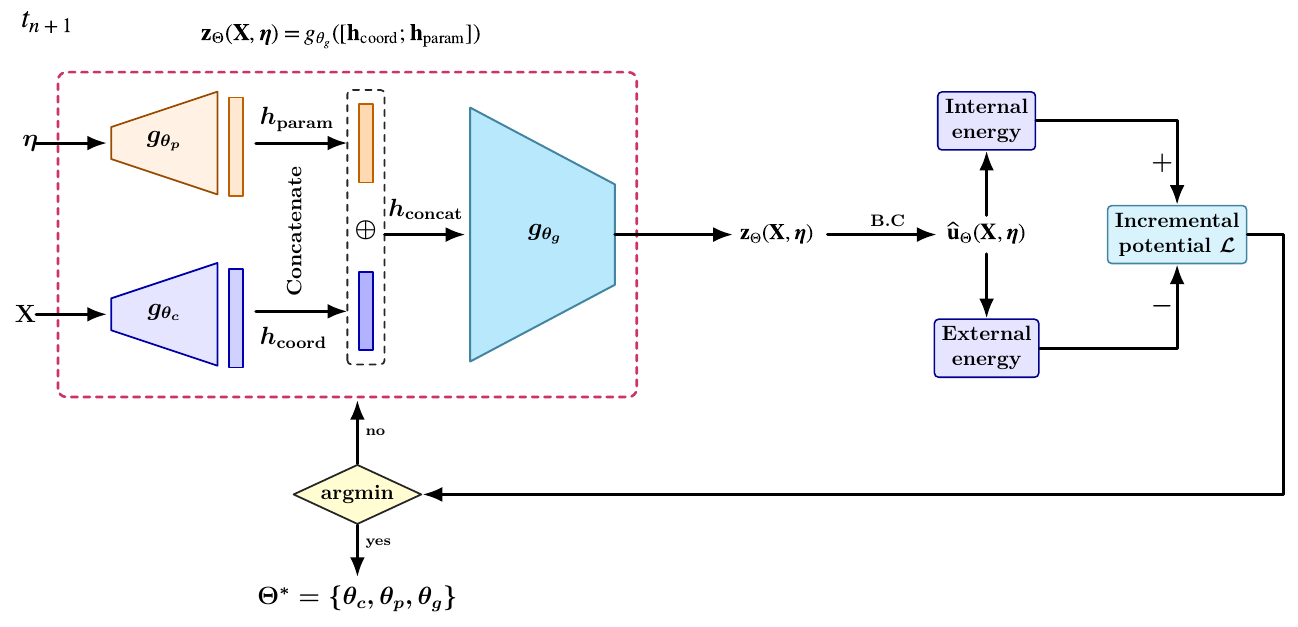}
    \par\vspace{-2pt}
    \caption{CPDEM architecture.
    The material encoder $g_{\theta_p}$ maps the constitutive vector to latent features, and the coordinate encoder $g_{\theta_c}$ maps the collocation coordinates to spatial features.
    The fusion network $g_{\theta_g}$ maps the concatenated features to $\bm z_\Theta$, and the boundary treatment gives the trial displacement $\widehat{\bm u}_\Theta$.}
    \label{fig:cpdem}
\end{figure}

In this section, we introduce the \textbf{C}onstitutive \textbf{P}arameterized \textbf{D}eep \textbf{E}nergy \textbf{M}ethod (\textbf{CPDEM}), as shown in Fig.~\ref{fig:cpdem}.
The objective is to construct a neural architecture that approximates the solution function $\bm{u}(\bm{X},\boldsymbol{\eta})$ of the governing mechanical equations when the random material parameters are sampled from the admissible parameter ranges.

In CPDEM, the material parameters are represented by $\boldsymbol{\eta}=[\eta_1,\eta_2,\ldots,\eta_{d_\eta}]^{\mathrm T}\in\mathcal P\subset\mathbb R^{d_\eta}$, where each component denotes one constitutive parameter.
The raw network map $\bm z_\Theta(\bm X,\boldsymbol\eta)$ has a modular architecture consisting of the separate material and coordinate encoders $g_{\theta_p}$ and $g_{\theta_c}$ together with a fusion network $g_{\theta_g}$ such that
\begin{align*}
    \bm z_\Theta(\bm X,\boldsymbol\eta) = g_{\theta_g}([g_{\theta_c}(\bm X);g_{\theta_p}(\boldsymbol\eta)]),
\end{align*}
where $\Theta = \{\theta_c, \theta_p, \theta_g\}$ denotes the set of model parameters.
The coordinate encoder maps the collocation coordinates $\bm{X}$ to $\bm{h}_{\text{coord}} = g_{\theta_c}(\bm{X})$, whereas the material encoder maps $\boldsymbol{\eta}$ to $\bm{h}_{\text{param}} = g_{\theta_p}(\boldsymbol{\eta})$.
Applying the boundary treatment $\mathfrak B$ to the raw network output gives $\widehat{\bm u}_\Theta=\mathfrak B[\bm z_\Theta]$, with $\mathfrak B=\mathfrak B_{n+1}$ at increment $n+1$ and $\mathfrak B$ equal to the identity when the boundary condition is imposed through a penalty alone.

The architectural choice is to encode the material parameters into a hidden representation before concatenating that representation with the coordinate features.
Using the spatial coordinate and encoded material features, we write the CPDEM mapping as
\begin{equation}
    \label{eq:cpdem-latent-form}
    \bm z_\Theta(\bm X,\boldsymbol\eta) = \bm z_{\{\theta_c,\theta_g\}}(\bm X,\bm h_{\text{param}}),
\end{equation}
where $\bm{h}_{\text{param}}$ encodes the constitutive parameters supplied to the network.

The representation argument for CPDEM combines two complementary viewpoints.
Neural networks approximate continuous mappings, while networks equipped with the rectified linear unit $\operatorname{ReLU}(x)=\max(0,x)$ can express continuous piecewise-linear finite-element functions under the constructions studied in \cite{opschoor2020relu}.
The following derivation reviews this correspondence on a fixed one-dimensional mesh before introducing material parameters as additional network inputs.
The resulting algebraic observation motivates the separation between spatial and material features in the proposed architecture.
The derivation establishes a fixed-mesh, one-dimensional representation identity.
Extending this representation to multidimensional finite-element discretizations and recovering coefficients through training require dedicated mathematical analysis beyond the one-dimensional setting.

Consider a fixed discretization of the interval $[0, l]$: $x_0 < x_1 < \dots < x_n$, where $x_0 = 0$ and $x_n = l$.
A continuous piecewise-linear FE approximation of $u(x; \boldsymbol{\eta})$ on this mesh takes the form:
\begin{equation}
\label{uh_FE}
u^h(x, \boldsymbol{\eta}) = \sum_{i=0}^{n} u_i(\boldsymbol{\eta}) \, N_i(x),
\end{equation}
where $N_i(x)$ denotes the standard spatial hat function associated with node $i$, and $u_i(\boldsymbol{\eta})$ denotes the nodal displacement as a function of $\boldsymbol{\eta}$, reflecting how the deformation field changes across the material parameter space.
For intermediate nodes, $N_i(x)$ can be expressed in terms of ReLU functions as Eq.~\eqref{eq:hat-function-relu}:
\begin{equation}
\label{eq:hat-function-relu}
N_i(x) = \frac{\text{ReLU}(x - x_{i-1}) - \text{ReLU}(x - x_i)}{h_i}
- \frac{\text{ReLU}(x - x_i) - \text{ReLU}(x - x_{i+1})}{h_{i+1}},
\end{equation}
where $h_i = x_i - x_{i-1}$ and $i=1,\ldots,n-1$ for the interior nodal basis functions.
Expanding and collecting terms gives:
\begin{equation}
\label{Ni_ReLU2}
N_i(x) = \frac{1}{h_i} \text{ReLU}(x - x_{i-1})
- \left(\frac{1}{h_i} + \frac{1}{h_{i+1}}\right) \text{ReLU}(x - x_i)
+ \frac{1}{h_{i+1}} \text{ReLU}(x - x_{i+1}).
\end{equation}

Substituting Eq.
\eqref{Ni_ReLU2} into Eq.
\eqref{uh_FE} and collecting coefficients of $\text{ReLU}(x - x_i)$ yields Eq.~\eqref{eq:uh-relu-expanded}:
\begin{equation}
\label{eq:uh-relu-expanded}
u^h(x, \boldsymbol{\eta}) = u_0(\boldsymbol{\eta}) + \frac{\Delta_1(\boldsymbol{\eta})}{h_1}\text{ReLU}(x-x_0)+\sum_{i=1}^{n-1}\left(\frac{\Delta_{i+1}(\boldsymbol{\eta})}{h_{i+1}}-\frac{\Delta_i(\boldsymbol{\eta})}{h_i}\right)\text{ReLU}(x-x_i),
\end{equation}
where $\Delta_i(\boldsymbol{\eta}) = u_i(\boldsymbol{\eta}) - u_{i-1}(\boldsymbol{\eta})$ for $i = 1, 2, \dots, n$.
The endpoint contribution is written separately using the defined mesh intervals; $h_0$ falls outside the range of interval indices.
Defining the $\boldsymbol{\eta}$-dependent weight functions
\begin{equation*}
    w_0(\boldsymbol{\eta}) = \frac{\Delta_1(\boldsymbol{\eta})}{h_1},\qquad w_i(\boldsymbol{\eta}) = \frac{\Delta_{i+1}(\boldsymbol{\eta})}{h_{i+1}} - \frac{\Delta_i(\boldsymbol{\eta})}{h_i},\quad i=1,\ldots,n-1,
\end{equation*}
we arrive at the compact parameterized ReLU representation:
\begin{equation}
\label{uh_ReLU2}
u^h(x, \boldsymbol{\eta}) = u_0(\boldsymbol{\eta}) + \sum_{i=0}^{n-1} w_i(\boldsymbol{\eta}) \, \text{ReLU}(x - x_i).
\end{equation}

Equation \eqref{uh_ReLU2} gives the parameterized counterpart of the standard one-dimensional correspondence between a finite-element function and a shallow network.
The identity shows that $u^h$ is a shallow neural network with input $(x, \boldsymbol{\eta})$, spatial biases at node positions $x_i$, and weights $w_i(\boldsymbol{\eta})$.
For any fixed $\boldsymbol{\eta}=\boldsymbol{\eta}_0$, Eq.~\eqref{uh_ReLU2} reduces to the ReLU representation of the fixed-mesh finite-element function.
Continuous dependence of $w_i$ on $\boldsymbol{\eta}$ additionally requires continuous dependence of the displacement solution on the material parameters, as ensured by uniform coercivity and regularity in linear elasticity.
For nonlinear and history-dependent constitutive models, parameter dependence is formulated within the incremental variational framework developed in Section~\ref{problem}.

The FE--ReLU equivalence in Eq.~\eqref{uh_ReLU2} connects directly to the CPDEM architecture: the material encoder maps $\boldsymbol{\eta}$ to a latent representation $\bm{h}_{\text{param}} = g_{\theta_p}(\boldsymbol{\eta})$, which modulates the spatial features generated by $g_{\theta_c}(\bm{X})$.
In correspondence with Eq.~\eqref{uh_ReLU2}, the effective nodal weights $w_i(\boldsymbol{\eta})$ are realized as parameter-conditioned functions $w_i(\boldsymbol{\eta}) = \phi_i(g_{\theta_p}(\boldsymbol{\eta}))$ within the fusion network $g_{\theta_g}$.
In this view, the spatial encoder $g_{\theta_c}$ produces the hidden representation motivated by the shared spatial bases $\{\text{ReLU}(x - x_i)\}$, while the material encoder $g_{\theta_p}$ drives the coefficient functions $\{\phi_i\}$.
The CPDEM approximation on a one-dimensional domain is given in Eq.~\eqref{eq:cpdem-1d-network}:
\begin{equation}
\label{eq:cpdem-1d-network}
u_{\Theta}^h(x, \boldsymbol{\eta}) = u_0(\boldsymbol{\eta}) + \sum_{i=0}^{n-1} \phi_i(g_{\theta_p}(\boldsymbol{\eta})) \, \text{ReLU}(x - x_i)
= g_{\theta_g}\!\big( [g_{\theta_c}(x); \, g_{\theta_p}(\boldsymbol{\eta}) ] \big).
\end{equation}

This reformulation shows that CPDEM can be interpreted as a parameter-conditioned spatial approximation: a fixed set of coordinate features is shared across all material configurations, and the coefficients are obtained from the latent representation $\bm{h}_{\text{param}}$ produced by the material encoder.
The correspondence concerns the representation of a displacement field, while a finite-element solver determines that field through the discrete solution process.

The CPDEM parameterization defines a shared approximation over the random-parameter domain, whereas standard finite-element analysis requires assembling and solving a discrete system for each material configuration even though the same mesh can normally be reused when the geometry is unchanged.
CPDEM shares coordinate features across the parameter values and conditions them through $\bm{h}_{\text{param}}$.

The FE--ReLU correspondence represents a piecewise-linear spatial approximation with material-dependent coefficients.
The correspondence explains why a network can represent a piecewise-linear spatial approximation whose coefficients depend on the material input.
Coefficient recovery from energy minimization depends on the selected widths, activation functions, and optimizer.
In particular, a single multilayer perceptron that receives the concatenated input $[\bm X;\boldsymbol\eta]$ has the same basic approximation property.
The separate encoders systematically organize the spatial coordinate and material-parameter inputs prior to feature fusion.

The material encoder maps the numerical values in $\boldsymbol\eta$ to a learned representation, while the constitutive relation remains in the strain-energy density or incremental potential evaluated during optimization.
Inputs of different dimensions across the elastic, plastic, and viscoelastic examples can be represented with the same network form, and each constitutive family specifies the energy expression, internal variables, and boundary conditions.

Over a random-parameter domain, the target is a field-valued map whose accurate approximation requires both a spatial representation of boundary layers and stress concentrations and a parameter representation of changes in stiffness or history-dependent response.
Errors can originate from either source, as well as from quadrature or boundary enforcement.
The numerical section evaluates direct field-level comparisons against high-fidelity reference solutions across the parameter domain.

With the parameterized approximation established, the one-dimensional linear-elastic energy functional takes the explicit variational form shown in Eq.~\eqref{eq:parametric-elastic-energy}:
\begin{equation}
\label{eq:parametric-elastic-energy}
\mathcal{E}(\Theta; \mathcal{P})
= \mathbb{E}_{\boldsymbol{\eta} \sim p(\boldsymbol{\eta})}\!\left[\int_{0}^{l}\!\left\{\Psi\!\big(\bm{\varepsilon}(u_{\Theta}^h(x,\boldsymbol{\eta}));\boldsymbol{\eta}\big)-b(x)u_{\Theta}^h(x,\boldsymbol{\eta})\right\}\,dx-\bar t\,u_{\Theta}^h(l,\boldsymbol{\eta})\right],
\end{equation}
where $b$ and $\bar t$ denote the one-dimensional body-force and endpoint-traction data.
The expectation defines a population objective over the selected parameter measure.
The exact product-space minimizer of this objective recovers the parameter-wise linear-elastic solutions under the assumptions of Section~\ref{theoretical-analysis}, whereas a finite network trained with finite samples generally retains approximation, sampling, quadrature, and optimization errors.
The population-averaged training objective over the admissible parameter domain reads Eq.~\eqref{eq:parametric-training-objective}:
\begin{equation}
\label{eq:parametric-training-objective}
\Theta_{\mathrm{pop}} = \arg\min_{\Theta} \; \mathcal{E}(\Theta; \mathcal{P}).
\end{equation}

In Eq.~\eqref{eq:parametric-training-objective}, $\Theta_{\mathrm{pop}}$ denotes an ideal minimizer of the population energy over the full material domain.
The parameterized objective optimizes one shared field over the material domain, whereas a per-configuration FE solve determines the discrete solution for one material parameter set.

The material encoder $g_{\theta_p}$ reads the constitutive vector $\boldsymbol{\eta}$ and generates the hidden representation $\bm{h}_{\text{param}}$.
The encoder has the following fully connected structure:
\begin{align*}
    \bm{h}_{\text{param}} = \sigma(\mathrm{FC}_{D_p} \cdots (\sigma(\mathrm{FC}_2(\sigma(\mathrm{FC}_1(\boldsymbol{\eta})))))),
\end{align*} 
where $\sigma$ denotes a nonlinear activation, such as ReLU or tanh, $\mathrm{FC}_i$ denotes the $i$-th fully connected layer of the encoder, and $D_p$ denotes the total number of fully connected layers.

The latent dimension can exceed the dimension of $\boldsymbol{\eta}$ so that the subsequent fusion network receives a higher-dimensional representation of the constitutive input.
The dimension and components of $\boldsymbol{\eta}$ remain determined by the selected constitutive model.
For example, the vector can contain Young's modulus $E$ and Poisson's ratio $\nu$ for an elastic problem, whereas the history-dependent examples require additional constitutive quantities.

The coordinate encoder $g_{\theta_c}$ reads the collocation coordinates $\bm{X}$ and generates the hidden representation $\bm{h}_{\text{coord}}$.
This encoder has the following fully connected layer structure:
\begin{align*}
    \bm{h}_{\text{coord}} = \sigma(\mathrm{FC}_{D_c} \cdots (\sigma(\mathrm{FC}_2(\sigma(\mathrm{FC}_1(\bm{X})))))),
\end{align*}
where $\mathrm{FC}_i$ and $D_c$ denote the $i$-th fully connected layer of the coordinate encoder and the total number of fully connected layers in that encoder, respectively.

The fusion network $g_{\theta_g}$ receives the hidden representations $\bm{h}_{\text{param}}$ and $\bm{h}_{\text{coord}}$ and returns the raw field $\bm z_\Theta(\bm X,\boldsymbol\eta)$.
The network has the form
\begin{align*}
    \bm z_\Theta(\bm X,\boldsymbol\eta) = \mathrm{FC}_{D_g}\bigl(\sigma(\mathrm{FC}_{D_g-1}(\cdots\sigma(\mathrm{FC}_1(\bm{h}_{\text{concat}}))\cdots))\bigr),
\end{align*}
where $\bm{h}_{\text{concat}} = \bm{h}_{\text{coord}} \oplus \bm{h}_{\text{param}}$, $\oplus$ is the concatenation of the two vectors, and $D_g$ denotes the number of FC layers.
The fusion network takes the concatenated outputs of the two encoders.
Table~\ref{tab:cpdem-config} lists the layer widths.

\subsection{Parameterized energy formulation}
Building upon the parameterized architecture, we reformulate the energy-based loss so that the material parameters are incorporated as explicit network inputs.
In contrast to a fixed-parameter DEM, which embeds the material properties as constants in the energy functional, CPDEM evaluates the constitutive law directly at the sampled parameter vectors $\boldsymbol{\eta}$ within one shared network model.

For a sampled material vector, the parameterized CPDEM energy loss can be shown in Eq.~\eqref{eq:cpdem-energy}:
\begin{equation}
    \label{eq:cpdem-energy}
    \underbrace{\mathcal{L}_{\text{CPDEM}}(\hat{\bm{u}},\boldsymbol\eta; \Theta)}_{\text{Parameterized potential energy}}
    = \underbrace{U(\hat{\bm{u}}, \boldsymbol{\eta}; \Theta)}_{\text{Internal energy}}
    - \underbrace{W(\hat{\bm{u}}, \boldsymbol{\eta}; \Theta)}_{\text{External energy}},
\end{equation}
where $\Theta = \{\theta_c, \theta_p, \theta_g\}$ is the full set of trainable network parameters.
In CPDEM, the neural network $\hat{\bm{u}}(\bm{X},\boldsymbol{\eta})$ accepts the spatial coordinate $\bm{X}$ and a model-specific material vector $\boldsymbol{\eta}\in\mathcal P\subset\mathbb R^{d_\eta}$ as inputs.
The trial displacement field depends on the material parameters supplied during training and evaluation.


For a fixed material vector $\boldsymbol\eta$, the ideal network weights satisfy
\begin{equation}
    \Theta_{\boldsymbol\eta} = \underset{\Theta}{\operatorname{argmin}} \; \mathcal{L}_{\text{CPDEM}}(\hat{\bm{u}},\boldsymbol\eta; \Theta),
    \quad \text{s.t.}\quad \hat{\bm u}(\bm X,\boldsymbol\eta)=\mathfrak B[\bm z_\Theta](\bm X,\boldsymbol\eta),
    \label{eq:cpdem_opt}
\end{equation}
where $\Theta_{\boldsymbol\eta}$ denotes an ideal minimizer for one material vector.
Section~\ref{sec:train} averages the single-material energies to train the shared model, whose computed weights are frozen as $\Theta^*$ in Algorithm~\ref{alg:cpdem-training-inference}.

\subsection{Parameterized pretraining over seen material parameters}
\label{sec:train}

\begin{figure}[pos=htbp]
    \centering
    \includegraphics[width=\linewidth]{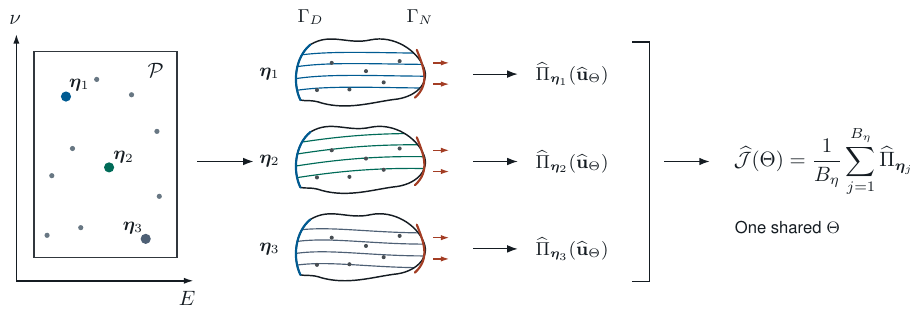}
    \par\vspace{-2pt}
    \caption{Schematic of material sampling and energy averaging in CPDEM.
    Each $\widehat\Pi_{\boldsymbol\eta_j}$ denotes the spatially discretized potential for material sample $\boldsymbol\eta_j$.
Their average is the training objective $\widehat{\mathcal J}(\Theta)$ in Eq.~\eqref{eq:cpdem-discrete-objective}.}
    \label{fig:cpdem-material-sampling}
\end{figure}

In CPDEM, parameterized pretraining denotes unsupervised optimization over seen material parameter sets sampled from the random-parameter domain $\mathcal{P}$.
Each Adam step draws a mini-batch of material configurations from $p(\boldsymbol{\eta})=\mathcal{U}(\mathcal{P})$, together with interior and Neumann-boundary quadrature points.
Let $\widehat\Pi_j(\Theta)$ denote the spatially discretized potential $\widehat\Pi_{\boldsymbol\eta_j}$ for material sample $\boldsymbol\eta_j$ in Fig.~\ref{fig:cpdem-material-sampling}.
The discrete estimator of the expected objective $\mathcal J$ is
\begin{equation}
\label{eq:cpdem-discrete-objective}
\begin{aligned}
    \widehat{\mathcal J}(\Theta)
    &=\frac{1}{B_\eta}\sum_{j=1}^{B_\eta}\widehat\Pi_j(\Theta),\\
    \widehat\Pi_j(\Theta)
    &=\frac{|\Omega|}{B_\Omega}\sum_{k=1}^{B_\Omega}
      \left(\Psi_{jk}-\bm f_{b,jk}\cdot\hat{\bm u}_{jk}\right)
      -\frac{|\Gamma_N|}{B_\Gamma}\sum_{\ell=1}^{B_\Gamma}
      \bar{\bm t}_{j\ell}\cdot\hat{\bm u}_{j\ell}
      +\mathcal L_{D,j},
\end{aligned}
\end{equation}
where $\mathcal{L}_{D,j}$ vanishes for hard Dirichlet enforcement and otherwise denotes the stated boundary penalty, while $B_\eta$, $B_\Omega$, and $B_\Gamma$ are the parameter, interior, and boundary batch sizes. 

The indices $j$, $k$, and $\ell$ identify material samples $\boldsymbol\eta_j$, interior points $\bm X_k$, and boundary points $\bm X_\ell^\Gamma$, respectively.
Thus $\hat{\bm u}_{jk}=\hat{\bm u}(\bm X_k,\boldsymbol\eta_j)$, $\hat{\bm u}_{j\ell}=\hat{\bm u}(\bm X_\ell^\Gamma,\boldsymbol\eta_j)$, and $\Psi_{jk}$ is the constitutive energy density at $\bm X_k$ for material sample $\boldsymbol\eta_j$.
Fig.~\ref{fig:cpdem-material-sampling} illustrates the two levels of integration in Eq.~\eqref{eq:cpdem-discrete-objective}.
Spatial quadrature approximates the mechanical potential for each sampled material vector $\boldsymbol\eta_j$ over the interior domain and loaded traction boundary.
Averaging these energies forms the objective used to optimize the shared network parameters $\Theta$.

Algorithm~\ref{alg:cpdem-training-inference} summarizes parameterized pretraining and direct inference for static elasticity and hyperelasticity.
Each sampled material vector enters both the displacement network and the constitutive energy; averaging the parameter-wise losses updates the shared weights.
Adam uses mini-batch estimates, followed by L-BFGS refinement with material samples and quadrature points fixed during repeated objective evaluations within each step.
The trained weights are held fixed for direct evaluation at unseen in-range parameters.
The training settings include the optimizer schedule, sample sizes, and stopping criteria. Displacement errors are evaluated against the reference fields.

\begin{algorithm}[tbp]
\caption{CPDEM training and direct prediction for static problems}
\label{alg:cpdem-training-inference}
\footnotesize
\begin{algorithmic}[1]
\Require Material samples $\bm\eta\sim\mathcal U(\mathcal P)$; energy $\Psi$; load and boundary data; networks $g_{\theta_c},g_{\theta_p},g_{\theta_g}$; Adam/L-BFGS steps $K_A,K_B$
\Ensure Shared trained weights $\Theta^*$ and displacement at a held-out
$\bm\eta^*\in\mathcal P$
\State Initialize shared weights $\Theta=\{\theta_c,\theta_p,\theta_g\}$
\Statex \textit{Offline parameterized training}
\For{$k=1,\ldots,K_A$}
    \State Sample $\{\bm\eta_j\}_{j=1}^{B_\eta}$ from $p$ and interior/boundary quadrature points
    \For{$j=1,\ldots,B_\eta$}
        \State $\bm z_j(\bm X)\gets g_{\theta_g}([g_{\theta_c}(\bm X);g_{\theta_p}(\bm\eta_j)])$
        \State Apply the Dirichlet boundary treatment to obtain $\widehat{\bm u}_j$
        \State Compute strain or deformation gradient from $\nabla_{\bm X}\widehat{\bm u}_j$
        \State Evaluate $\widehat\Pi_j(\Theta)$ using Eq.~\eqref{eq:cpdem-discrete-objective}
    \EndFor
    \State $\widehat{\mathcal J}(\Theta)\gets B_\eta^{-1}\sum_{j=1}^{B_\eta}\widehat\Pi_j(\Theta)$
    \State Update the shared $\Theta$ by Adam using $\nabla_\Theta\widehat{\mathcal J}$
\EndFor
\Statex \textit{L-BFGS refinement}
\For{$k=1,\ldots,K_B$}
    \State Select training material samples and quadrature points for this step
    \State Evaluate Eq.~\eqref{eq:cpdem-discrete-objective} with these samples fixed within the step
    \State Update the shared $\Theta$ by L-BFGS on that sampled objective
\EndFor
\State Set $\Theta^*\gets\Theta$ and freeze all network weights
\Statex \textit{Direct inference at an unseen in-range parameter}
\State $\bm z^*(\bm X)\gets \bm z_{\Theta^*}(\bm X,\bm\eta^*)$
\State Apply the same Dirichlet treatment to obtain $\widehat{\bm u}_{\Theta^*}(\bm X,\bm\eta^*)$
\State \Return $\Theta^*$ and $\widehat{\bm u}_{\Theta^*}(\cdot,\bm\eta^*)$
\end{algorithmic}
\end{algorithm}

For history-dependent materials, we formulate incremental CPDEM using the elastoplastic and viscoelastic potentials in Eqs.~\eqref{eq:elastoplastic-minimization} and \eqref{eq:viscoelastic-incremental-potential}.
Algorithm~\ref{alg:cpdem-incremental} specifies a fixed training parameter pool and one material-shared checkpoint $\Theta_{n+1}^*$ per load or time increment.
Each material parameter set stores an internal state $\bm s_n^{(j)}$, collecting $\bm\alpha_n=[\bm\varepsilon_n^p,\bar\varepsilon_n^p]$ for plasticity and $\bm\varepsilon^{v,n}$ for viscoelasticity.
Optimization holds the preceding committed states fixed and evaluates candidate updates. After optimization, the updated states are committed.
The increment is $\Delta t_n=t_{n+1}-t_n$.
The algorithm uses $N_t$ increments and $K_{A,n}$ Adam updates at increment $n+1$.
The update map $\mathcal R_{\boldsymbol\eta}^{n+1}$ uses radial return for plasticity and backward Euler for viscoelasticity, and $\mathfrak B_{n+1}$ applies the displacement boundary treatment at that increment.
In the training loops of Algorithm~\ref{alg:cpdem-incremental}, $\mathcal R_j^{n+1}$ denotes this update for $\bm\eta_j$.

For an unseen in-range parameter, the initial internal state starts an independent history.
The frozen sequence $\{\Theta_{n+1}^*\}$ predicts displacement at successive increments, and the constitutive map advances the query's internal state.
In a time-conditioned notation, the incremental field is $\widehat{\bm u}(\bm X,t_{n+1};\boldsymbol\eta)=\widehat{\bm u}_{\Theta_{n+1}^*}(\bm X;\boldsymbol\eta)$.
This formulation describes history propagation through stored incremental models; the fixed-time relaxation comparison uses the stated snapshot energy.

\begin{algorithm}[tbp]
\caption{CPDEM training and history prediction over load and time increments}
\label{alg:cpdem-incremental}
\footnotesize
\begin{algorithmic}[1]
\Require Training set $\mathcal S_{\rm tr}=\{\bm\eta_j\}_{j=1}^{N_\eta}\subset\mathcal P$; load/time stations $\{t_n\}_{n=0}^{N_t}$; quadrature $\mathcal Q_\Omega$; incremental energy $\mathcal L_{\bm\eta}^{n+1}$; state update $\mathcal R_{\bm\eta}^{n+1}$
\Require Boundary operator $\mathfrak B_{n+1}$; raw CPDEM output $\bm z_\Theta$;
held-out query $\bm\eta^*\in\mathcal P$
\Ensure Material-shared checkpoints $\{\Theta_{n+1}^*\}$ and predicted unseen history
\State Set $\mathcal R_{\bm\eta}^{n+1}$ to radial return (plasticity) or backward Euler (viscoelasticity)
\State Initialize separate internal states $\{\bm s_0^{(j)}\}_{j=1}^{N_\eta}$ on $\mathcal Q_\Omega$
\Statex \textit{Offline training across material parameters}
\For{$n=0,\ldots,N_t-1$}
    \State Initialize shared weights $\Theta$ for increment $n+1$
    \State Keep committed $\{\bm s_n^{(j)}\}$ fixed during optimization
    \For{$k=1,\ldots,K_{A,n}$}
        \State Draw indices $\mathcal I_\eta\subseteq\{1,\ldots,N_\eta\}$
        \For{$j\in\mathcal I_\eta$}
            \State $\widehat{\bm u}_j\gets\mathfrak B_{n+1}[\bm z_\Theta(\bm X,\bm\eta_j)]$
            \State $\widetilde{\bm s}_{n+1}^{(j)}\gets
              \mathcal R_j^{n+1}(\bm\varepsilon(\widehat{\bm u}_j),\bm s_n^{(j)};\Delta t_n)$
            \State Evaluate $\widehat{\mathcal L}_j^{n+1}$ using
              $\widehat{\bm u}_j$, $\widetilde{\bm s}_{n+1}^{(j)}$, $\bm s_n^{(j)}$ and quadrature
        \EndFor
        \State $\widehat{\mathcal J}_{n+1}(\Theta)\gets
          |\mathcal I_\eta|^{-1}\sum_{j\in\mathcal I_\eta}\widehat{\mathcal L}_j^{n+1}$
        \State Update $\Theta$ by Adam using $\nabla_\Theta\widehat{\mathcal J}_{n+1}$
    \EndFor
    \State Optionally refine $\Theta$ by L-BFGS with samples fixed across repeated closures within each step
    \State Store $\Theta_{n+1}^*\gets\Theta$ and freeze this checkpoint
    \For{$j=1,\ldots,N_\eta$}
        \State Re-evaluate $\widehat{\bm u}_{n+1}^{(j)}$ with $\Theta_{n+1}^*$
        \State Commit $\bm s_{n+1}^{(j)}\gets
          \mathcal R_j^{n+1}(\bm\varepsilon(\widehat{\bm u}_{n+1}^{(j)}),\bm s_n^{(j)};\Delta t_n)$
    \EndFor
\EndFor
\Statex \textit{Direct evaluation at an unseen material parameter}
\State Fix the query at $\bm\eta^*$ and initialize $\widehat{\bm s}_0$ from the initial internal state
\For{$n=0,\ldots,N_t-1$}
    \State $\widehat{\bm u}_{n+1}\gets
      \mathfrak B_{n+1}[\bm z_{\Theta_{n+1}^*}(\bm X,\bm\eta^*)]$
    \State $\widehat{\bm s}_{n+1}\gets
      \mathcal R_{\bm\eta^*}^{n+1}(\bm\varepsilon(\widehat{\bm u}_{n+1}),\widehat{\bm s}_n;\Delta t_n)$
\EndFor
\State \Return $\{\widehat{\bm u}_{n+1},\widehat{\bm s}_{n+1}\}_{n=0}^{N_t-1}$
\end{algorithmic}
\end{algorithm}

\subsection{Fine-tuning for a specified configuration}
\label{sec:tune}

For a specified in-range material configuration $\boldsymbol{\eta}^{*}$, the parameterized pretrained network can be further optimized by holding the parameter value fixed in the discrete energy objective.
A practical option is to freeze the coordinate and material encoders and update only the fusion-network parameters for a limited number of L-BFGS iterations.
The encoder-freezing procedure serves as a local post-processing optimization distinct from direct zero-retraining inference using the frozen pretrained surrogate model.

Fine-tuning may be useful when a particular material parameter set requires higher accuracy than the pretrained surrogate provides, serving as an optional post-processing step following the direct inference procedure in Algorithm~\ref{alg:cpdem-training-inference}.

\section{Theoretical analysis of CPDEM}
\label{theoretical-analysis}

For linear elasticity, uniform coercivity yields an energy-norm estimate accounting for approximation, parameter sampling, spatial quadrature, boundary enforcement, and optimization errors.
The analysis builds on Deep Ritz and parametric variational convergence results \cite{muller2022deepRitzError,lu2021deepRitzGeneralization,dondl2022deepRitzNonlinear,kaltenbach2026parametric}.
Extensions to finite-strain hyperelasticity and incremental history-dependent problems assume local stability.

\subsection{Stability of parametric solutions}

For the theorem below, consider homogeneous essential data and let
\begin{equation*}
    V=H_D^1(\Omega;\mathbb R^d)
    =\left\{\bm v\in H^1(\Omega;\mathbb R^d)\mid \bm v=\bm 0\ \text{on}\ \Gamma_D\right\},
\end{equation*}
where $\Gamma_D$ has positive measure and $\|\cdot\|_V$ denotes the energy-space norm.
For a material parameter set $\boldsymbol\eta\in\mathcal P$, define
\begin{align*}
    a_{\boldsymbol\eta}(\bm w,\bm v)
    &=\int_\Omega \bm\varepsilon(\bm w):\mathbb C(\boldsymbol\eta):\bm\varepsilon(\bm v)\,d\Omega,
\\
    \Pi_{\boldsymbol\eta}(\bm v)
    &=\frac12a_{\boldsymbol\eta}(\bm v,\bm v)-\ell(\bm v),
\end{align*}
where $\ell\in V^*$ contains the body-force and traction work and is parameter independent.
A fixed lifting handles nonhomogeneous essential data; the resulting parameter-dependent load requires a uniformly bounded dual norm together with a parameter-continuity estimate.

\begin{assumption}[Uniform mechanical stability]
\label{ass:uniform-stability}
There are constants $0<\alpha\leq\beta<\infty$, independent of $\boldsymbol\eta$, such that
\begin{equation}
    a_{\boldsymbol\eta}(\bm v,\bm v)\geq\alpha\|\bm v\|_V^2,
    \qquad
    |a_{\boldsymbol\eta}(\bm w,\bm v)|\leq\beta\|\bm w\|_V\|\bm v\|_V,
    \label{eq:analysis-uniform-stability}
\end{equation}
for every $\bm v,\bm w\in V$ and every $\boldsymbol\eta\in\mathcal P$.
\end{assumption}

\begin{assumption}[Constitutive parameter regularity]
\label{ass:parameter-regularity}
Equip $\mathcal P$ with $\|\boldsymbol\eta-\boldsymbol\eta'\|_{\mathrm{par}}=\|D_\eta^{-1}(\boldsymbol\eta-\boldsymbol\eta')\|_2$, $D_\eta=\operatorname{diag}(\eta_{r,\max}-\eta_{r,\min})$, for all parameter distances and Lipschitz constants in this section. Assume there exists $L_a<\infty$ such that
\begin{equation*}
    |a_{\boldsymbol\eta_1}(\bm w,\bm v)-a_{\boldsymbol\eta_2}(\bm w,\bm v)|
    \leq L_a\|\boldsymbol\eta_1-\boldsymbol\eta_2\|_{\mathrm{par}}\,\|\bm w\|_V\|\bm v\|_V,
\end{equation*}
for all $\boldsymbol\eta_1,\boldsymbol\eta_2\in\mathcal P$ and all $\bm v,\bm w\in V$.
\end{assumption}

For isotropic linear elasticity, the ranges of $E$ and $\nu$ must ensure uniform stiffness and parameter-independent norm bounds.
These conditions define the admissible material parameter domain.

\begin{theorem}[Well-posedness and parameter-to-solution continuity]
\label{thm:parameter-stability}
Under Assumptions~\ref{ass:uniform-stability} and \ref{ass:parameter-regularity}, every $\boldsymbol\eta\in\mathcal P$ has a unique solution $\bm u_{\boldsymbol\eta}\in V$ satisfying
\begin{equation}
    a_{\boldsymbol\eta}(\bm u_{\boldsymbol\eta},\bm v)=\ell(\bm v)
    \qquad \text{for all }\bm v\in V.
    \label{eq:analysis-weak-solution}
\end{equation}

Moreover,
\begin{equation}
    \|\bm u_{\boldsymbol\eta_1}-\bm u_{\boldsymbol\eta_2}\|_V
    \leq \frac{L_a\|\ell\|_{V^*}}{\alpha^2}
    \|\boldsymbol\eta_1-\boldsymbol\eta_2\|_{\mathrm{par}}.
    \label{eq:analysis-solution-lipschitz}
\end{equation}
\end{theorem}

\begin{proof}
The uniform continuity and coercivity estimates give existence, uniqueness, and $\|\bm u_{\boldsymbol\eta}\|_V\leq\|\ell\|_{V^*}/\alpha$.
Let $\bm w=\bm u_{\boldsymbol\eta_1}-\bm u_{\boldsymbol\eta_2}$.
Subtracting Eq.~\eqref{eq:analysis-weak-solution} at $\boldsymbol\eta_1$ and $\boldsymbol\eta_2$ and testing with $\bm w$ gives
\begin{equation*}
\alpha\|\bm w\|_V^2
\leq L_a\|\boldsymbol\eta_1-\boldsymbol\eta_2\|_{\mathrm{par}}\,
\|\bm u_{\boldsymbol\eta_2}\|_V\|\bm w\|_V.
\end{equation*}

The stability estimate for $\bm u_{\boldsymbol\eta_2}$ yields Eq.~\eqref{eq:analysis-solution-lipschitz}.
\end{proof}

Theorem~\ref{thm:parameter-stability} establishes a Lipschitz-continuous solution family in the mechanical energy space, providing the regularity underlying the one-dimensional FE--ReLU representation in Section~\ref{methodology}.

\subsection{Population-energy formulation on the product space}

Let $p$ be the normalized training measure on $\mathcal P$ and introduce the Bochner space
\begin{equation*}
    \mathbb V_p=L^2(\mathcal P,p;V),
    \qquad
    \|\bm v\|_{\mathbb V_p}^2
    =\int_{\mathcal P}\|\bm v(\boldsymbol\eta)\|_V^2p(\boldsymbol\eta)\,d\boldsymbol\eta.
\end{equation*}

The exact population functional is
\begin{equation*}
    \mathcal J(\bm v)
    =\int_{\mathcal P}\Pi_{\boldsymbol\eta}(\bm v(\boldsymbol\eta))
    p(\boldsymbol\eta)\,d\boldsymbol\eta.
\end{equation*}

The measure $p$ specifies training weights.
Probabilistic interpretation requires identifying $p$ independently as the underlying physical material-parameter distribution.

\begin{theorem}[Equivalence of population and parameter-wise problems]
\label{thm:population-equivalence}
Under Assumptions~\ref{ass:uniform-stability} and \ref{ass:parameter-regularity}, define $\bm u^*(\boldsymbol\eta)=\bm u_{\boldsymbol\eta}$, where $\bm u_{\boldsymbol\eta}$ solves Eq.~\eqref{eq:analysis-weak-solution}.
Then $\bm u^*\in\mathbb V_p$ is the unique minimizer of $\mathcal J$ up to sets of zero $p$-measure, and
\begin{equation*}
    \bm u^*=\operatorname*{argmin}_{\bm v\in\mathbb V_p}\mathcal J(\bm v),
\end{equation*}
the slice $\bm u^*(\cdot,\boldsymbol\eta)$ equals the unique mechanical solution for $p$-almost every $\boldsymbol\eta$.
\end{theorem}

\begin{proof}
The uniform estimate $\|\bm u_{\boldsymbol\eta}\|_V\leq\|\ell\|_{V^*}/\alpha$ places the solution map in $\mathbb V_p$.
For every $\bm v\in\mathbb V_p$, parameter-wise minimality gives
\begin{equation*}
    \Pi_{\boldsymbol\eta}(\bm v(\boldsymbol\eta))
    \geq\Pi_{\boldsymbol\eta}(\bm u_{\boldsymbol\eta}),
\end{equation*}
for $p$-almost every $\boldsymbol\eta$.
Integration over $\mathcal P$ proves global minimality.
Strict convexity induced by uniform coercivity gives uniqueness up to a set of zero $p$-measure.
\end{proof}

The exact minimizer recovers the parameter-wise solutions $p$-almost everywhere.
A computed network introduces approximation, sampling, quadrature, and optimization errors.

\begin{corollary}[Population energy-error equivalence]
\label{cor:energy-error}
For every $\bm v\in\mathbb V_p$,
\begin{equation}
    \frac{\alpha}{2}\|\bm v-\bm u^*\|_{\mathbb V_p}^2
    \leq \mathcal J(\bm v)-\mathcal J(\bm u^*)
    \leq \frac{\beta}{2}\|\bm v-\bm u^*\|_{\mathbb V_p}^2.
    \label{eq:analysis-energy-error-equivalence}
\end{equation}
\end{corollary}

\begin{proof}
The parameter-wise weak equation cancels the cross term in the quadratic potential and gives
\begin{equation*}
    \Pi_{\boldsymbol\eta}(\bm v)-\Pi_{\boldsymbol\eta}(\bm u_{\boldsymbol\eta})
    =\frac12a_{\boldsymbol\eta}(\bm v-\bm u_{\boldsymbol\eta},\bm v-\bm u_{\boldsymbol\eta}).
\end{equation*}

The uniform lower and upper bounds in Eq.~\eqref{eq:analysis-uniform-stability}, followed by integration over $\mathcal P$, yield Eq.~\eqref{eq:analysis-energy-error-equivalence}.
\end{proof}

Corollary~\ref{cor:energy-error} relates the exact linear-elastic population energy to the parameter-averaged mechanical error; the computed-network estimate also includes finite-training discrepancies.

\subsection{Neural approximation in the population space}

Let
\begin{equation*}
    \mathbb V_m=\{\bm u_{\Theta}\mid\Theta\in\mathfrak T_m\}\subset\mathbb V_p,
\end{equation*}
be a CPDEM trial class indexed by $m$, with admissible network parameters $\mathfrak T_m$ and the essential boundary conditions imposed exactly,
where $\bm u_\Theta$ denotes the boundary-treated trial field $\widehat{\bm u}_\Theta$ from the CPDEM architecture.
The best approximation error over $\mathbb V_m$ in the parameter-averaged mechanical norm is
\begin{equation*}
    e_{\mathrm{app},m}
    =\inf_{\bm v\in\mathbb V_m}\|\bm v-\bm u^*\|_{\mathbb V_p}.
\end{equation*}

\begin{proposition}[C\'ea-type CPDEM approximation estimate]
\label{prop:cea-cpdem}
Suppose that $\widetilde{\bm u}_m\in\mathbb V_m$ satisfies
\begin{equation}
    \mathcal J(\widetilde{\bm u}_m)
    \leq\inf_{\bm v\in\mathbb V_m}\mathcal J(\bm v)
    +\varepsilon_{\mathrm{opt}}^{\mathrm{pop}}.
    \label{eq:analysis-population-suboptimality}
\end{equation}

Then
\begin{equation}
    \|\widetilde{\bm u}_m-\bm u^*\|_{\mathbb V_p}^2
    \leq\frac{\beta}{\alpha}e_{\mathrm{app},m}^2
    +\frac{2}{\alpha}\varepsilon_{\mathrm{opt}}^{\mathrm{pop}}.
    \label{eq:analysis-cea-estimate}
\end{equation}
\end{proposition}

\begin{proof}
Apply the lower inequality in Eq.~\eqref{eq:analysis-energy-error-equivalence} to $\widetilde{\bm u}_m$ and the upper inequality to an arbitrary $\bm v\in\mathbb V_m$.
Equation~\eqref{eq:analysis-population-suboptimality} gives
\begin{equation*}
    \frac{\alpha}{2}\|\widetilde{\bm u}_m-\bm u^*\|_{\mathbb V_p}^2
    \leq\frac{\beta}{2}\|\bm v-\bm u^*\|_{\mathbb V_p}^2
    +\varepsilon_{\mathrm{opt}}^{\mathrm{pop}}.
\end{equation*}

Taking the infimum over $\mathbb V_m$ proves Eq.~\eqref{eq:analysis-cea-estimate}.
\end{proof}

Proposition~\ref{prop:cea-cpdem} separates trial-class approximation from population-level optimization suboptimality.
Convergence requires trial classes dense in $\mathbb V_p$ and vanishing optimization suboptimality.

\subsection{Finite-training error estimates}

Parameter sampling approximates the material expectation, and spatial quadrature approximates each potential.
Let $\{\boldsymbol\eta_j\}_{j=1}^{N_\eta}$ be independent samples from $p$ and define
\begin{equation}
    \mathcal J_{N_\eta}(\bm v)
    =\frac{1}{N_\eta}\sum_{j=1}^{N_\eta}
    \Pi_{\boldsymbol\eta_j}(\bm v(\boldsymbol\eta_j)).
    \label{eq:analysis-parameter-empirical-functional}
\end{equation}

Fix the interior and Neumann-boundary quadrature rules for the discrete objective.
Let $\widehat\Pi_{\boldsymbol\eta_j}$ denote the resulting quadrature approximation of the complete potential for material sample $\boldsymbol\eta_j$.
The discrete objective is
\begin{equation}
    \widehat{\mathcal J}(\bm v)
    =\frac{1}{N_\eta}\sum_{j=1}^{N_\eta}
    \widehat\Pi_{\boldsymbol\eta_j}(\bm v(\boldsymbol\eta_j)).
    \label{eq:analysis-fully-discrete-functional}
\end{equation}

As in Eq.~\eqref{eq:cpdem-discrete-objective}, spatial quadrature approximates each mechanical potential, and the outer sample average approximates the expected potential over $\mathcal{P}$.

For the selected architecture, use the trial class $\mathbb V_m$ with parameter set $\mathfrak T=\mathfrak T_m$.
Define
\begin{align}
    \varepsilon_{\rm app}
    &=\inf_{\Theta\in\mathfrak T}
    \left[\mathcal J(\bm u_{\Theta})-\mathcal J(\bm u^*)\right],
\notag \\
    \varepsilon_{\rm disc}
    &=\sup_{\Theta\in\mathfrak T}
    |\mathcal J(\bm u_{\Theta})-\widehat{\mathcal J}(\bm u_{\Theta})|.
    \label{eq:analysis-disc-error}
\end{align}

Let $\widehat{\bm u}=\bm u_{\widehat\Theta}$ denote the trial field at the computed parameters $\widehat\Theta$ and suppose that
\begin{equation}
    \widehat{\mathcal J}(\widehat{\bm u})
    \leq\inf_{\Theta\in\mathfrak T}\widehat{\mathcal J}(\bm u_{\Theta})
    +\varepsilon_{\rm opt}.
    \label{eq:analysis-opt-error}
\end{equation}

\begin{proposition}[Finite-training error bound]
\label{prop:finite-training}
For a trial class satisfying the Dirichlet condition exactly,
\begin{equation*}
    \|\widehat{\bm u}-\bm u^*\|_{\mathbb V_p}
    \leq
    \sqrt{\frac{2}{\alpha}
    \left(\varepsilon_{\rm app}+\varepsilon_{\rm opt}
    +2\varepsilon_{\rm disc}\right)}.
\end{equation*}
\end{proposition}

\begin{proof}
Add and subtract $\widehat{\mathcal J}$ at $\widehat{\bm u}$ and at an arbitrary member of $\mathbb V_m$, apply Eqs.~\eqref{eq:analysis-disc-error} and \eqref{eq:analysis-opt-error}, and take the infimum over the trial class.
This gives
\begin{equation*}
    \mathcal J(\widehat{\bm u})-\mathcal J(\bm u^*)
    \leq\varepsilon_{\rm app}+\varepsilon_{\rm opt}+2\varepsilon_{\rm disc}.
\end{equation*}

The lower bound in Corollary~\ref{cor:energy-error} completes the estimate.
\end{proof}

For exact Dirichlet enforcement, define the two uniform discrepancies
\begin{align*}
    \varepsilon_{\eta}
    &=\sup_{\bm v\in\mathbb V_m}
    |\mathcal J(\bm v)-\mathcal J_{N_\eta}(\bm v)|,
\\
    \varepsilon_{\rm quad}
    &=\sup_{\bm v\in\mathbb V_m}
    |\mathcal J_{N_\eta}(\bm v)-\widehat{\mathcal J}(\bm v)|.
\end{align*}

The triangle inequality applied to Eqs.~\eqref{eq:analysis-parameter-empirical-functional}--\eqref{eq:analysis-fully-discrete-functional} gives
\begin{equation}
    \varepsilon_{\rm disc}
    \leq\varepsilon_{\eta}+\varepsilon_{\rm quad},
    \label{eq:analysis-disc-decomposition}
\end{equation}
where $\varepsilon_{\eta}$ and $\varepsilon_{\rm quad}$ denote constitutive-parameter sampling and spatial and boundary quadrature errors, respectively.
Finite-penalty boundary enforcement adds a consistency contribution $\varepsilon_{\rm BC}$, measured in the squared $\mathbb V_p$ error norm, when comparing with the constrained solution.
The estimate accounts for approximation, optimization, sampling, quadrature, and boundary enforcement, each requiring a separate bound.

For completeness, let
\begin{equation*}
    \mathcal F=\left\{\boldsymbol\eta\mapsto
    \Pi_{\boldsymbol\eta}(\bm u_{\Theta}(\boldsymbol\eta))
    \mid\Theta\in\mathfrak T\right\}.
\end{equation*}

If every member of $\mathcal F$ is measurable and has absolute value at most $M$, and $N_\eta$ material parameter samples are drawn independently from $p$, a standard Rademacher-complexity estimate gives, with probability at least $1-\delta$,
\begin{equation}
    \varepsilon_{\eta}
    \leq 2\mathfrak R_{N_\eta}(\mathcal F)
    +3M\sqrt{\frac{\log(2/\delta)}{2N_\eta}},
    \label{eq:analysis-rademacher}
\end{equation}
where $\delta\in(0,1)$ is the failure probability.
The expected Rademacher complexity is defined using the absolute-supremum convention,
\begin{equation*}
    \mathfrak R_{N_\eta}(\mathcal F)
    =\mathbb E_{\boldsymbol\eta,\epsilon}
    \left[\sup_{f\in\mathcal F}
    \left|\frac{1}{N_\eta}\sum_{j=1}^{N_\eta}\epsilon_j f(\boldsymbol\eta_j)\right|\right],
\end{equation*}
where the independent signs $\epsilon_j\in\{-1,1\}$ have equal probabilities and are independent of the material samples.
Quantitative sampling rates require bounds on the trial-class complexity and finite-energy envelope.

\begin{theorem}[Total CPDEM error for the discrete objective]
\label{thm:analysis-total-cpdem-error}
Assume the hypotheses of Theorem~\ref{thm:population-equivalence} and the conditional sampling estimate in Eq.~\eqref{eq:analysis-rademacher}.
Let $\widehat{\bm u}$ satisfy Eq.~\eqref{eq:analysis-opt-error}.
Then, with probability at least $1-\delta$ over the $N_\eta$ independently sampled material parameter samples,
\begin{equation}
\begin{aligned}
    \|\widehat{\bm u}-\bm u^*\|_{\mathbb V_p}^{2}
    \leq{}&\frac{\beta}{\alpha}e_{\mathrm{app},m}^{2}
    +\frac{4}{\alpha}\left(\varepsilon_\eta+\varepsilon_{\mathrm{quad}}\right)
    +\frac{2}{\alpha}\varepsilon_{\mathrm{opt}}.
\end{aligned}
    \label{eq:analysis-explicit-total-error}
\end{equation}

For penalty enforcement, $\varepsilon_{\mathrm{BC}}$ denotes the boundary-consistency contribution measured in the squared $\mathbb V_p$ error norm and appears on the right-hand side.
This boundary-consistency contribution vanishes identically whenever the trial field satisfies the essential condition exactly.
\end{theorem}

\begin{proof}
For each $\bm v\in\mathbb V_m$, Eqs.~\eqref{eq:analysis-disc-error} and \eqref{eq:analysis-disc-decomposition} give
\begin{equation*}
    |\mathcal J(\bm v)-\widehat{\mathcal J}(\bm v)|
    \leq\varepsilon_\eta+\varepsilon_{\mathrm{quad}}.
\end{equation*}

Apply this bound at $\widehat{\bm u}$ and an arbitrary comparison function in $\mathbb V_m$ and use Eq.~\eqref{eq:analysis-opt-error}.
The resulting population-energy gap is at most the best trial-class energy gap plus $2(\varepsilon_\eta+\varepsilon_{\mathrm{quad}})+\varepsilon_{\mathrm{opt}}$.
Combining both sides of Eq.~\eqref{eq:analysis-energy-error-equivalence} with the definition of $e_{\mathrm{app},m}$ directly yields Eq.~\eqref{eq:analysis-explicit-total-error}.
\end{proof}

Theorem~\ref{thm:analysis-total-cpdem-error} separates approximation, sampling, quadrature, boundary, and optimization errors.
Asymptotic rates require control of neural complexity, quadrature, and optimization accuracy along a specified sequence of calculations.

Parameter coverage gives a deterministic estimate.
Let $\mathcal P_N=\{\boldsymbol\eta_i\}_{i=1}^{N}$ be the training set, with $N=N_\eta$.
The fill distance of $\mathcal P_N$ is
\begin{equation*}
    h_{\mathcal P_N}=\sup_{\boldsymbol\eta\in\mathcal P}
    \min_{\boldsymbol\eta_i\in\mathcal P_N}
    \|\boldsymbol\eta-\boldsymbol\eta_i\|_{\mathrm{par}}.
\end{equation*}

\begin{proposition}[Conditional parameter-coverage estimate]
\label{prop:parameter-coverage}
Suppose the exact solution map and a trained network are Lipschitz continuous in $\boldsymbol\eta$ with constants $L_u$ and $L_\Theta$, respectively.
If
\begin{equation*}
    \max_{1\leq i\leq N_\eta}
    \|\bm u_{\Theta}(\boldsymbol\eta_i)-\bm u^*(\boldsymbol\eta_i)\|_V
    \leq\varepsilon_{\rm train},
\end{equation*}
every $\boldsymbol\eta\in\mathcal P$ satisfies
\begin{equation}
    \|\bm u_{\Theta}(\boldsymbol\eta)-\bm u^*(\boldsymbol\eta)\|_V
    \leq\varepsilon_{\rm train}+(L_u+L_\Theta)h_{\mathcal P_N}.
    \label{eq:analysis-coverage-bound}
\end{equation}
\end{proposition}

\begin{proof}
Choose a nearest sampled point $\boldsymbol\eta_i$ and insert both $\bm u_{\Theta}(\boldsymbol\eta_i)$ and $\bm u^*(\boldsymbol\eta_i)$ into the error.
The triangle inequality and the two Lipschitz bounds give Eq.~\eqref{eq:analysis-coverage-bound}.
Taking the worst-case nearest-point distance gives the stated result.
\end{proof}

\begin{figure}[pos=htbp]
    \centering
    \includegraphics[width=.9\linewidth]{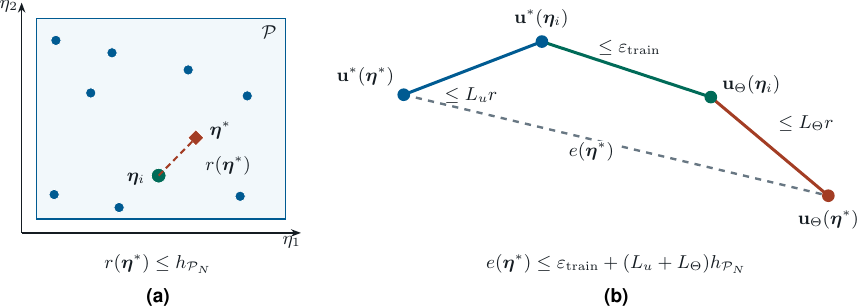}
    \par\vspace{-2pt}
    \caption{Parameter coverage and error bounds for unseen parameters.
    (a) Distance to the nearest training parameter, $r\le h_{\mathcal P_N}$.
    (b) Contributions to the error bound from variation in the exact solution, training error, and variation in the learned solution.}
    \label{fig:parameter-coverage-error}
\end{figure}

Fig.~\ref{fig:parameter-coverage-error} illustrates Proposition~\ref{prop:parameter-coverage} for an unseen vector $\boldsymbol\eta^*$ and its nearest training vector $\boldsymbol\eta_i$.
With $r(\boldsymbol\eta^*)=\|\boldsymbol\eta^*-\boldsymbol\eta_i\|_{\mathrm{par}}$ and $e(\boldsymbol\eta^*)=\|\bm u_\Theta(\boldsymbol\eta^*)-\bm u^*(\boldsymbol\eta^*)\|_V$, the Lipschitz assumptions bound the three contributions by $L_u r$, $\varepsilon_{\mathrm{train}}$, and $L_\Theta r$.

Theorem~\ref{thm:parameter-stability} supplies $L_u$ for linear elasticity; $L_\Theta$ requires numerical quantification from the trained weights and activation functions.
Proposition~\ref{prop:parameter-coverage} motivates evaluating fill distance, held-out prediction errors, and nearest-training-point distance over the parameter domain.

\subsection{Error estimates across constitutive regimes}

The global estimates use the quadratic, uniformly coercive energy of linear elasticity.
For finite-strain hyperelasticity, assume a stable solution branch with uniform energy growth in an admissible neighborhood of $\bm u^*_{\boldsymbol\eta}$:
\begin{equation}
    c_-\operatorname{dist}_{\boldsymbol\eta}(\bm v,\bm u^*_{\boldsymbol\eta})^2
    \leq\Pi_{\boldsymbol\eta}(\bm v)-\Pi_{\boldsymbol\eta}(\bm u^*_{\boldsymbol\eta})
    \leq c_+\operatorname{dist}_{\boldsymbol\eta}(\bm v,\bm u^*_{\boldsymbol\eta})^2,
    \label{eq:analysis-nonlinear-energy-growth}
\end{equation}
with constants uniform over the considered parameter domain.
Under these assumptions, the population and finite-training estimates apply locally with $\operatorname{dist}_{\boldsymbol\eta}$ replacing the linear-elastic norm, consistent with nonlinear Deep Ritz convergence results \cite{dondl2022deepRitzNonlinear,kaltenbach2026parametric}.
For compressible Neo-Hookean materials, Eq.~\eqref{eq:analysis-nonlinear-energy-growth} specifies the local stability assumption, while kinematic admissibility requires $J=\det\bm F>0$ throughout $\Omega$.

Elastoplastic and viscoelastic increments depend on the preceding constitutive state.
Let $\bm s_n$ collect the internal variables at increment $n$ and define the displacement--state error
\begin{equation*}
    e_n=\|\widehat{\bm u}_n-\bm u_n^*\|_V
    +c_s\|\widehat{\bm s}_n-\bm s_n^*\|_{\mathcal S},
\end{equation*}
where $c_s>0$ is a fixed weight and $\|\bm s\|_{\mathcal S}^2=\int_\Omega\sum_a\|\bm s^{(a)}\|_F^2\,d\Omega$, with $a$ indexing the internal-state components and the absolute value used for scalar components.
Assume the reduced incremental solution and update map is locally Lipschitz in the previous state with constant $L_s$ measured in this weighted displacement--state norm.
With $\zeta_{n+1}$ collecting neural approximation, parameter-sampling, quadrature, boundary, and optimization errors at increment $n+1$, the one-step estimate is
\begin{equation*}
    e_{n+1}\leq L_s e_n+\zeta_{n+1}.
\end{equation*}

Repeated application gives
\begin{equation}
    e_n\leq L_s^n e_0+\sum_{j=1}^{n}L_s^{\,n-j}\zeta_j.
    \label{eq:analysis-history-error-propagation}
\end{equation}

Equation~\eqref{eq:analysis-history-error-propagation} motivates assessment over the full loading history.
For standard-linear-solid viscoelasticity, the backward-Euler potential defines the time-discrete update, with viscous strain included in $\bm s_n$.
For associative $J_2$ plasticity, the estimate assumes local stability and a consistent displacement-reduced return mapping through yield-surface transitions.
The common parameter-conditioned variational construction admits global estimates for uniformly stable linear elasticity and local estimates under the stated constitutive assumptions.

\section{Numerical examples}
\label{numerical-examples}
Four groups of numerical benchmarks test CPDEM across different constitutive laws and geometries, beginning with small-strain linear elasticity (Section~\ref{sec:elasticity}) and 2-D and 3-D finite-strain Neo-Hookean hyperelasticity (Section~\ref{sec:neohook}). The remaining groups cover path-dependent $J_2$ elastoplasticity under cyclic loading (Section~\ref{sec:plas}) and rate-dependent viscoelastic creep, stress relaxation, and creep buckling (Section~\ref{sec:visco}).
The order of these examples follows the theoretical structure in Section~\ref{theoretical-analysis}. We verify the uniformly coercive linear-elastic setting, proceed to geometrically nonlinear finite-strain equilibria, and conclude with history-dependent problems governed by incremental potentials and parameter-specific internal-variable updates across loading increments.

Table~\ref{tab:cpdem-config} summarizes the network architectures used across all eight benchmarks.
In every parameterized-pretraining run, material parameters are sampled uniformly over the parameter domain $\mathcal{P}$ in accordance with Section~\ref{problem}, and the interior, Neumann-boundary, and parameter samples enter the discrete expected-energy objective in Eq.~\eqref{eq:cpdem-discrete-objective}.
For the history-dependent benchmarks in Sections~\ref{sec:plas} and~\ref{sec:visco}, the incremental potentials and constitutive state updates (radial return and backward Euler, respectively) replace the static potential energy.
Together, the sampling layout, quadrature rule, boundary treatment, and optimizer schedule define the discrete variational problem solved in each case.

\begin{table}[pos=htbp]
    \centering
    \caption{CPDEM architecture implementation details.
    }
    \label{tab:cpdem-config}
    \small
    \setlength{\tabcolsep}{2pt}
    \resizebox{\textwidth}{!}{%
    \begin{tabular}{lcccccc}
        \toprule
        Problem & $\dim(\boldsymbol{\eta})$ & Components of $\boldsymbol{\eta}$ & $g_{\theta_p}$ & $g_{\theta_c}$ & $g_{\theta_g}$ & Output dim. \\
        \midrule
        2-D linear-elastic beam & 2 & $[E,\nu]$ & $[2,16,16]$ & $[2,16,16]$ & $[32,32,2]$ & 2 \\
        2-D Neo-Hookean beam & 2 & $[E,\nu]$ & $[2,20]$ & $[2,20]$ & $[40,20,2]$ & 2 \\
        3-D Neo-Hookean beam & 2 & $[E,\nu]$ & $[2,30]$ & $[3,30]$ & $[60,30,3]$ & 3 \\
        3-D Neo-Hookean cuboid (torsion) & 2 & $[E,\nu]$ & $[2,30]$ & $[3,30]$ & $[60,30,3]$ & 3 \\
        3-D Plate-with-hole $J_2$ plasticity & 3 & $[E,\nu,\kappa]$ & $[3,32,32]$ & $[3,32,32]$ & $[64,64,3]$ & 3 \\
        2-D viscoelastic creep & 4 & $[E_\infty,E_1,\xi,\nu]$ & $[4,32,32]$ & $[3,32,32]$ & $[64,64,2]$ & 2 \\
        2-D viscoelastic relaxation & 4 & $[E_\infty,E_1,\xi,\nu]$ & $[4,32,32]$ & $[3,32,32]$ & $[64,64,2]$ & 2 \\
        2-D circular viscoelastic creep buckling & 4 & $[E_\infty,E_1,\xi,\nu]$ & $[8,150,150,50]$ & $[3,50,50,50]$ & $[100,100,50,50,2]$ & 2 \\
        \bottomrule
    \end{tabular}%
    }
\end{table}

\subsection{Linear elasticity problems}
\label{sec:elasticity}
\begin{figure}[pos=htbp]
    \centering
    \includegraphics[height=0.125\textwidth]{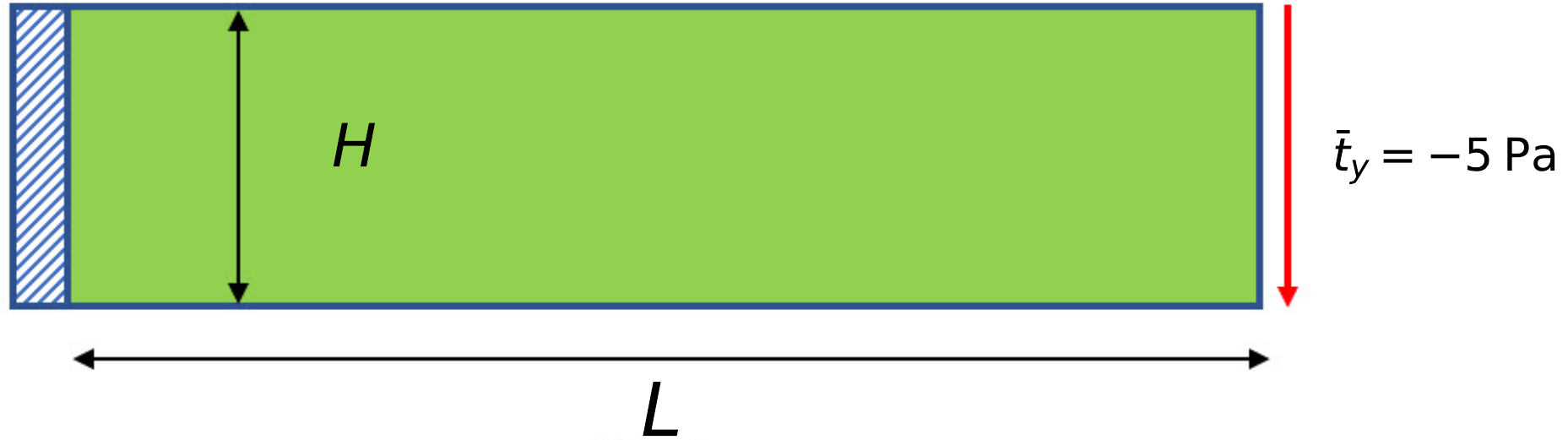}
    \par\vspace{-2pt}
    \caption{Geometry and boundary conditions of the 2-D cantilever beam benchmarks.}
    \label{fig:2Dcase}
\end{figure}

We begin with a two-dimensional plane-strain cantilever beam to establish a baseline in the quadratic, uniformly coercive regime of Section~\ref{theoretical-analysis}.
As illustrated in Fig.~\ref{fig:2Dcase}, the beam has length $L=4\,\mathrm{m}$ and height $H=1\,\mathrm{m}$, is clamped at its left end ($x_1 = 0$) with zero displacement and rotation, and carries a uniform downward traction $\bar{t}_y=-5\,\mathrm{Pa}$ along its right end ($x_1 = L$).
The material is homogeneous, isotropic, and linearly elastic, with Young's modulus $E$ in $\mathrm{Pa}$ and dimensionless Poisson's ratio $\nu$.
Because no closed-form solution is available for this 2-D problem, a converged FEM solution on a $150\times50$-node mesh provides the reference benchmark, while CPDEM learns a single parameterized displacement field $\hat{\bm{u}}(x_1,x_2;E,\nu)$ over the target $(E,\nu)$ domain.

\begin{figure}[pos=htbp]
    \centering
    \includegraphics[width=\textwidth]{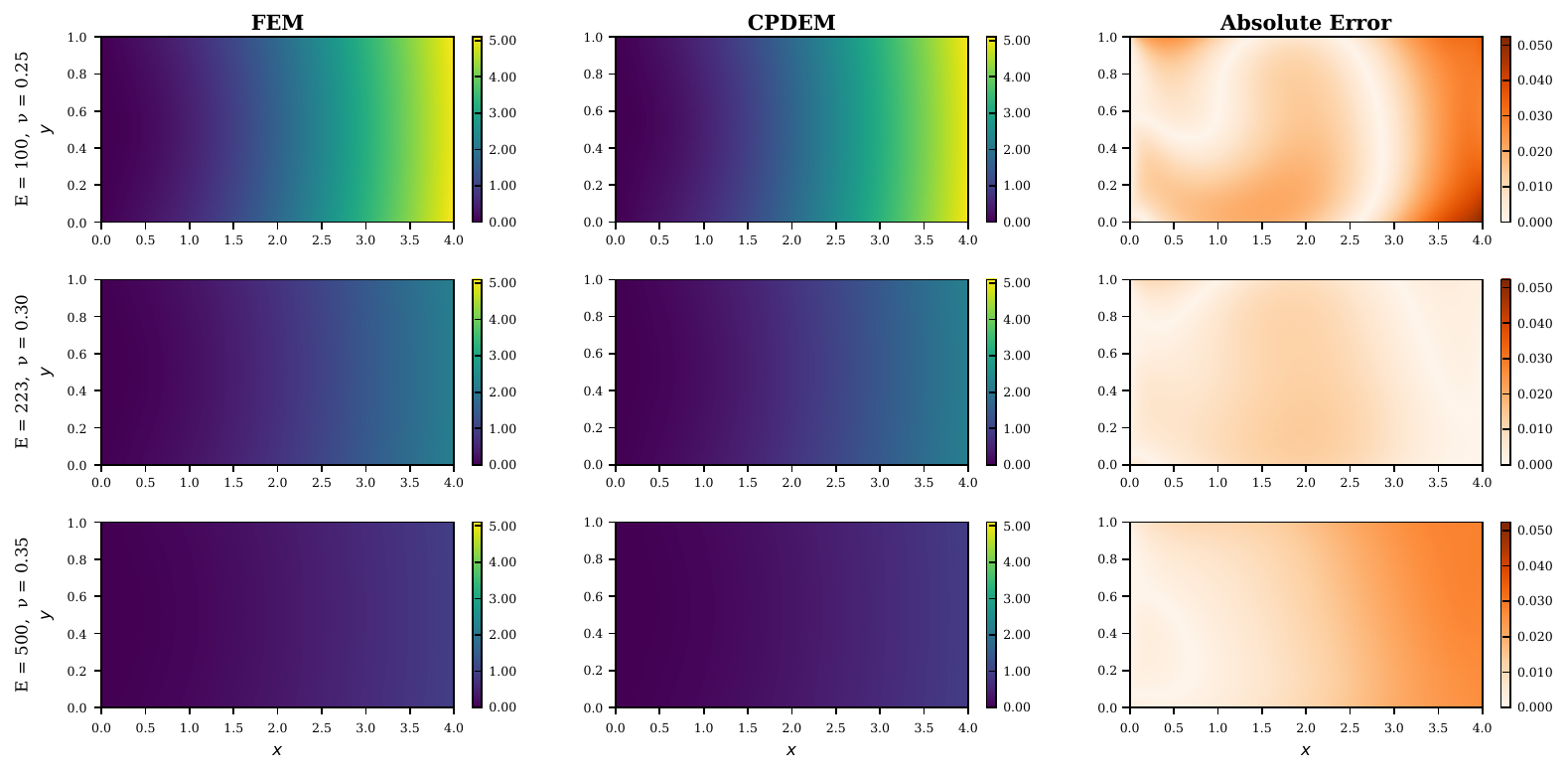}
    \par\vspace{-2pt}
    \caption{Comparison between FEM and CPDEM displacement fields for the 2-D linear-elastic cantilever. Rows use $(E,\nu)=(100,0.25)$, $(223,0.30)$, and $(500,0.35)$ ($E$ in $\mathrm{Pa}$); columns show FEM, CPDEM, and pointwise error ($\mathrm{m}$).}
    \label{fig:beam2d_elasticity_u}
\end{figure}

Fig.~\ref{fig:beam2d_elasticity_u} compares the CPDEM displacement fields with the FEM reference at three unseen in-range material parameter pairs across the stiffness range: $(E,\nu)=(100,0.25)$, $(223,0.30)$, and $(500,0.35)$.
As expected from linear beam bending, increasing $E$ fivefold from $100$ to $500\,\mathrm{Pa}$ under the same end traction reduces the free-end deflection from nearly $5\,\mathrm{m}$ to about $1\,\mathrm{m}$, while the concurrent change in $\nu$ introduces secondary lateral effects.
For all three parameter pairs, the predicted displacement fields agree closely with FEM: against the shared $5\,\mathrm{m}$ displacement scale, the maximum pointwise absolute error remains $0.050\,\mathrm{m}$ (roughly $1\%$ of the peak deflection).
The error contours also show a clear contrast between interior and boundary parameters. For the interior pair $(223,0.30)$, the discrepancy stays below $0.015\,\mathrm{m}$ across almost the entire beam. Larger residual errors appear near the loaded right end for the two boundary pairs and reach about $0.050\,\mathrm{m}$ at the lower-right corner of the softest case $(100,0.25)$, where both bending gradients and deflection amplitudes are largest.
This parameter-space error pattern arises directly from the expected-energy formulation in Eq.~\eqref{eq:cpdem-discrete-objective}. Under uniform sampling over the parameter domain, interior parameter pairs are constrained from all sides by neighboring training samples, whereas extreme stiffness corners lie on the boundary of the parameter support and involve steeper spatial deformation gradients.
In addition, minimizing the first-order potential energy rather than second-order strong-form equilibrium residuals satisfies the traction-free lateral surfaces naturally without tuning boundary penalty weights. The small field discrepancy is also consistent with Corollary~\ref{cor:energy-error}, where a reduced population potential-energy gap bounds the parameter-averaged mechanical error.
This concentration of residual error near high-gradient loaded boundaries is quantified in detail in the next section together with the finite-strain hyperelastic benchmarks.

\subsection{Hyperelasticity problems}
\label{sec:neohook}
We next test CPDEM on three compressible Neo-Hookean problems involving finite deformations, geometric nonlinearity, and progressively more complex boundary conditions: a 2-D cantilever beam, a 3-D cantilever beam, and a 3-D cuboid under combined torsion, lateral traction, and body force.

\textbf{2-D hyperelastic beam.}
Retaining the cantilever geometry of Fig.~\ref{fig:2Dcase} ($L=4\,\text{m}$, $H=1\,\text{m}$), the left end is clamped and the right end carries a uniform downward traction $\bar{t}_y=-5\,\text{Pa}$, while Young's modulus and Poisson's ratio are sampled uniformly over $E\in[800,1200]\,\text{Pa}$ and $\nu\in[0.25,0.35]$.
Both the FEM reference and the CPDEM spatial collocation grid use $150\times50$ points.
To satisfy the clamped condition at $X_1=0$ exactly without boundary penalties, the trial displacement is parameterized as
\begin{equation}
    \label{instance3-2-1}
    \widehat{\bm u}_\Theta(\bm X,\boldsymbol\eta)
    =X_1\bm z_\Theta(\bm X,\boldsymbol\eta),
\end{equation}
where $\bm X\in\mathbb R^2$ and $\bm z_\Theta$ is the raw output of the separate-encoder network in Table~\ref{tab:cpdem-config}.
Training follows the two-stage schedule of Section~\ref{sec:train}: an initial Adam phase (learning rate $0.5$) followed by L-BFGS refinement over $120$ epochs.

\begin{figure}[pos=htbp]
    \centering
    \includegraphics[width=0.82\textwidth]{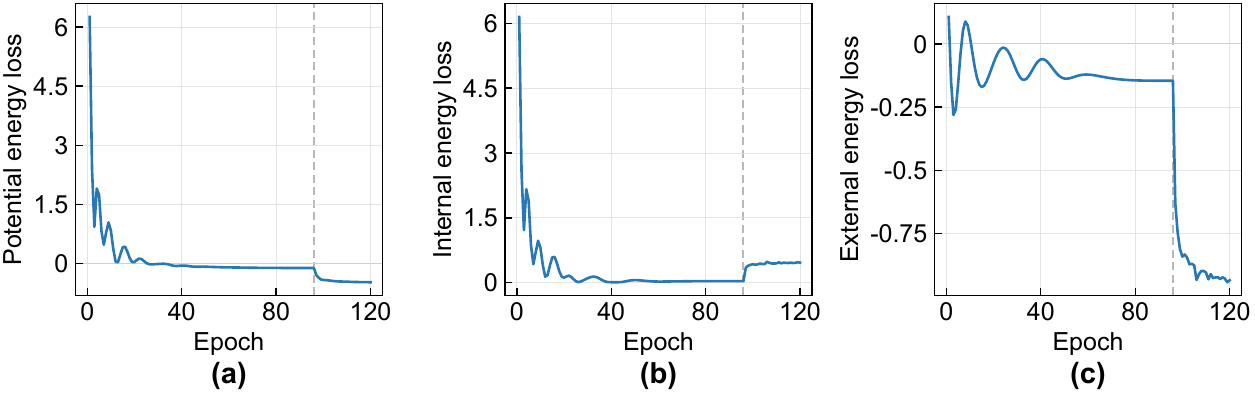}
    \par\vspace{-2pt}
    \caption{Training energy histories for the 2-D Neo-Hookean cantilever: (a) total, (b) internal, and (c) external potential energy per unit thickness ($\mathrm{J/m}$), with the dashed line marking the Adam-to-L-BFGS transition.}
    \label{fig:loss_beam2d_neohook}
\end{figure}

Fig.~\ref{fig:loss_beam2d_neohook} shows that switching from Adam to L-BFGS accelerates convergence of the potential energy. The total potential energy drops sharply to a stable plateau as the stored internal strain energy rises and the external work potential becomes correspondingly more negative.
Because the clamped boundary condition is enforced exactly through Eq.~\eqref{instance3-2-1}, this smooth descent corresponds purely to interior energy equilibration, avoiding gradient competition between boundary penalty terms and bulk strain energy.
In the first stage of parameterized finite-strain training, Adam rapidly aligns the coupled coordinate and material encoders across the sampled parameter family. The quasi-Newton L-BFGS stage complements these first-order updates by using curvature information to converge inside the flat energy valley where the first-order updates slow down.

\begin{figure}[pos=htbp]
    \centering
    \includegraphics[width=\textwidth]{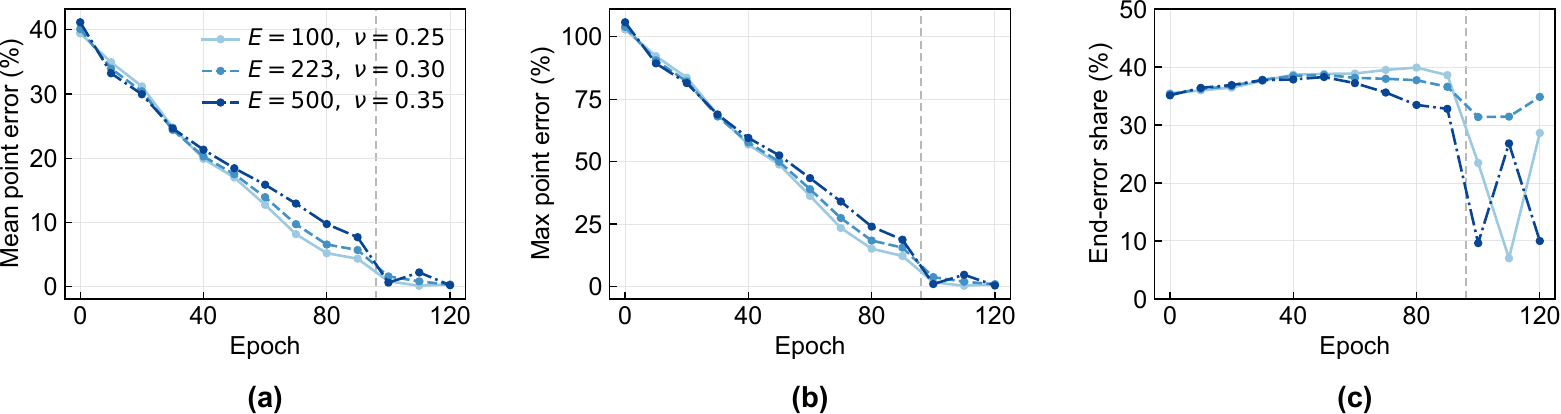}
    \par\vspace{-2pt}
    \caption{Training error histories for the plane-stress linear-elastic cantilever across three pretraining samples: (a) mean point error, (b) maximum point error, and (c) loaded-end squared-error share.}
    \label{fig:linear-error-localization}
\end{figure}

Fig.~\ref{fig:linear-error-localization} reports a complementary plane-stress linear-elastic diagnostic to examine how the two-stage optimization affects pointwise displacement accuracy and where the residual error concentrates, following the observations in Section~\ref{sec:elasticity}. The diagnostic covers $120$ outer optimizer steps for three pretraining samples, $(E,\nu)=(100,0.25)$, $(223,0.30)$, and $(500,0.35)$, with each L-BFGS step potentially executing multiple internal iterations.
For each sample, the pointwise displacement-vector error is normalized by the maximum FEM reference displacement magnitude, yielding the spatial mean and maximum in panels (a) and (b).
To quantify boundary error concentration, panel (c) tracks the loaded-end squared-error share over the terminal tenth of the beam,
\begin{equation}
    R_{\mathrm{end}}
    =
    \frac{
        \displaystyle\int_{\Omega_{\mathrm{end}}}
        \|\widehat{\bm u}-\bm u_{\mathrm{FEM}}\|_2^2\,\mathrm{d}\Omega
    }{
        \displaystyle\int_{\Omega}
        \|\widehat{\bm u}-\bm u_{\mathrm{FEM}}\|_2^2\,\mathrm{d}\Omega
    }
    \times100\%,
    \qquad
    \Omega_{\mathrm{end}}=\{\bm X\in\Omega:X_1/L\geq0.9\},
    \label{eq:loaded-end-squared-error-share}
\end{equation}
evaluated by composite trapezoidal quadrature on the full displacement-error fields.

Consistent with the energy drop in Fig.~\ref{fig:loss_beam2d_neohook}, both the mean and maximum displacement errors fall sharply after the Adam-to-L-BFGS switch, with minor late-stage oscillations.
At epoch $120$, the mean errors reach $0.375\%$, $0.310\%$, and $0.245\%$, while the corresponding maximum point errors reach $0.939\%$, $0.932\%$, and $0.426\%$ across the three pretraining samples.
Although the peak point error always occurs at the loaded tip ($X_1/L=1$), panel (c) shows that the spatial distribution of squared error depends strongly on the paired $(E,\nu)$ values: the loaded-end shares over $\Omega_{\mathrm{end}}$ at epoch $120$ are $28.62\%$, $34.85\%$, and $9.99\%$.
Since $\Omega_{\mathrm{end}}$ covers $10\%$ of the beam area, the third case exhibits nearly uniform squared-error density (a local-to-interior density ratio near $1$), whereas the first two cases retain concentrated residual error near the loaded end with density ratios of about $3.61$ and $4.81$, respectively.

\begin{figure}[pos=htbp]
    \centering
    \includegraphics[width=\textwidth]{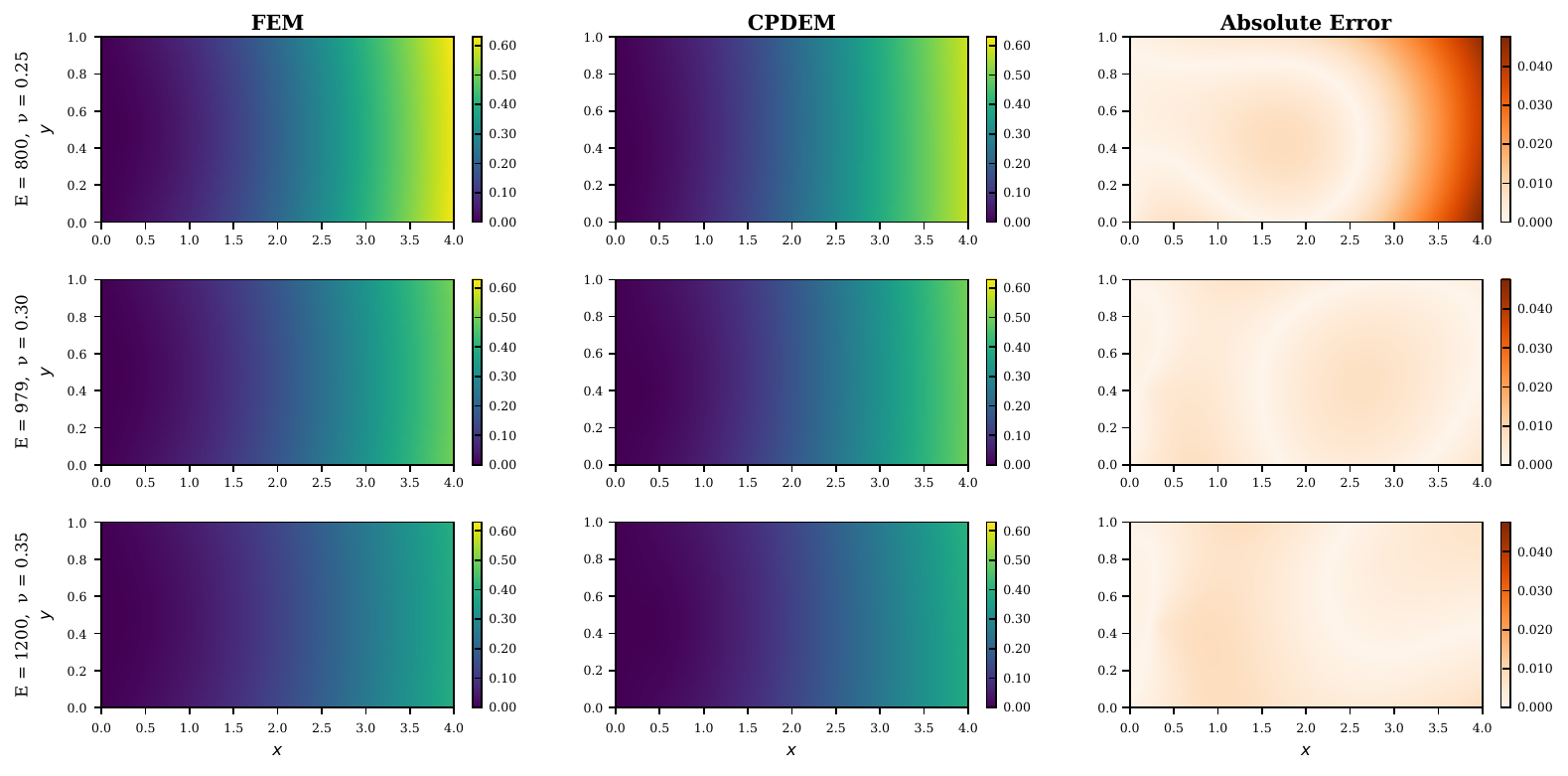}
    \par\vspace{-2pt}
    \caption{Comparison between FEM and CPDEM displacement magnitudes for the 2-D Neo-Hookean cantilever. Rows use $(E,\nu)=(800,0.25)$, $(979,0.30)$, and $(1200,0.35)$; columns show FEM, CPDEM, and pointwise error ($\mathrm{m}$).}
    \label{fig:beam2d_neohook_u}
\end{figure}

For the 2-D Neo-Hookean beam, Fig.~\ref{fig:beam2d_neohook_u} compares the CPDEM displacement magnitudes against FEM at three unseen in-range material pairs: $(E,\nu)=(800,0.25)$, $(979,0.30)$, and $(1200,0.35)$.
As $E$ and $\nu$ increase together under the same end traction, the free-end deflection decreases steadily from nearly $0.60\,\mathrm{m}$ to roughly $0.35\,\mathrm{m}$ owing to the stiffer finite-strain bending response.
The pointwise error follows the same parameter-space pattern observed in linear elasticity. In the softest boundary case $(800,0.25)$, peak errors near $0.040\,\mathrm{m}$ localize tightly along the loaded right edge ($x=4$) where nonlinear bending curvature is highest. For both the interior pair $(979,0.30)$ and the stiffer boundary pair $(1200,0.35)$, the discrepancies stay uniformly low across the beam (largely below $0.015\,\mathrm{m}$).

\begin{figure}[pos=htbp]
    \centering
    \includegraphics[width=0.60\textwidth]{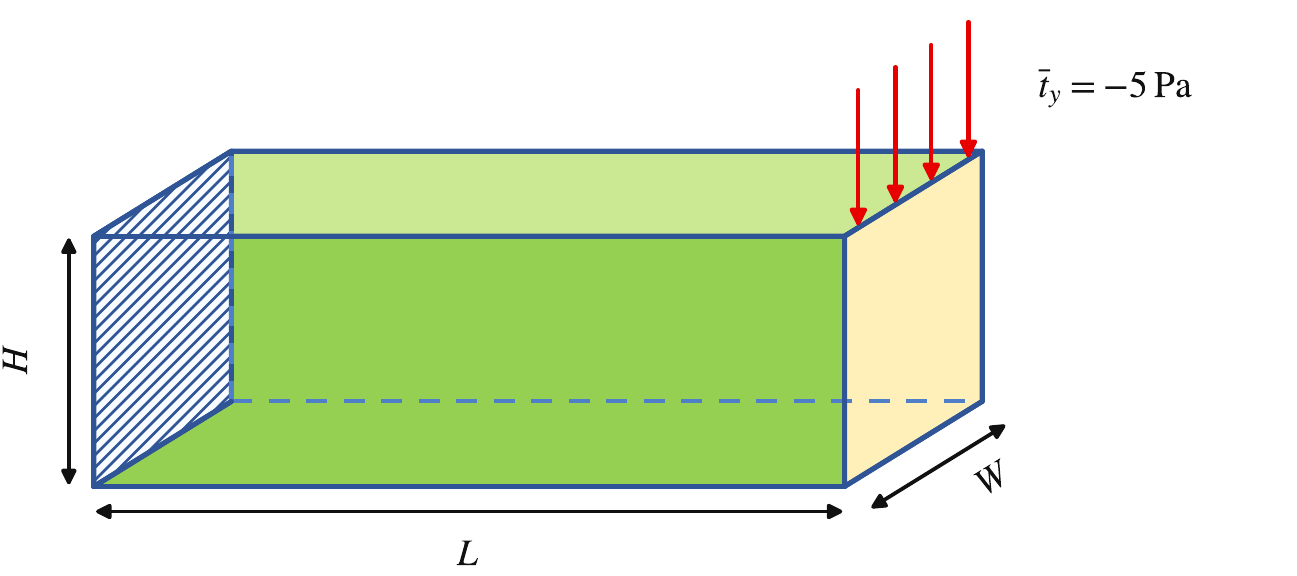}
    \par\vspace{-2pt}
    \caption{Geometry and loading of the 3-D Neo-Hookean cantilever beam ($L=4\,\mathrm{m}$, $H=W=1\,\mathrm{m}$), clamped on the left face with traction $\bar t_y=-5\,\mathrm{Pa}$ on the right face.}
    \label{fig:3Dcase-beam}
\end{figure}

\textbf{3-D hyperelastic beam.}
We next extend the Neo-Hookean bending problem to three dimensions ($L=4\,\text{m}$, $H=W=1\,\text{m}$; Fig.~\ref{fig:3Dcase-beam}), keeping the left face clamped and applying a uniform traction $\bar{t}_y=-5\,\text{Pa}$ on the right face.
Using the 3-D separate-encoder architecture in Table~\ref{tab:cpdem-config}, the trial field again enforces the clamped root exactly via
\begin{equation}
    \label{eq:beam3d-dirichlet-ansatz}
    \widehat{\bm u}_\Theta(\bm X,\boldsymbol\eta)
    =X_1\bm z_\Theta(\bm X,\boldsymbol\eta),
\end{equation}
with $\bm X\in\mathbb R^3$.
For systematic coverage of the parameter domain $E\in[800,1200]\,\text{Pa}$ and $\nu\in[0.25,0.35]$, reference FEM solutions are computed on a structured $40\times10\times10$-element mesh across a $21\times11$ grid of $231$ $(E,\nu)$ pairs.

\begin{figure}[pos=htbp]
    \centering
    \includegraphics[width=\textwidth]{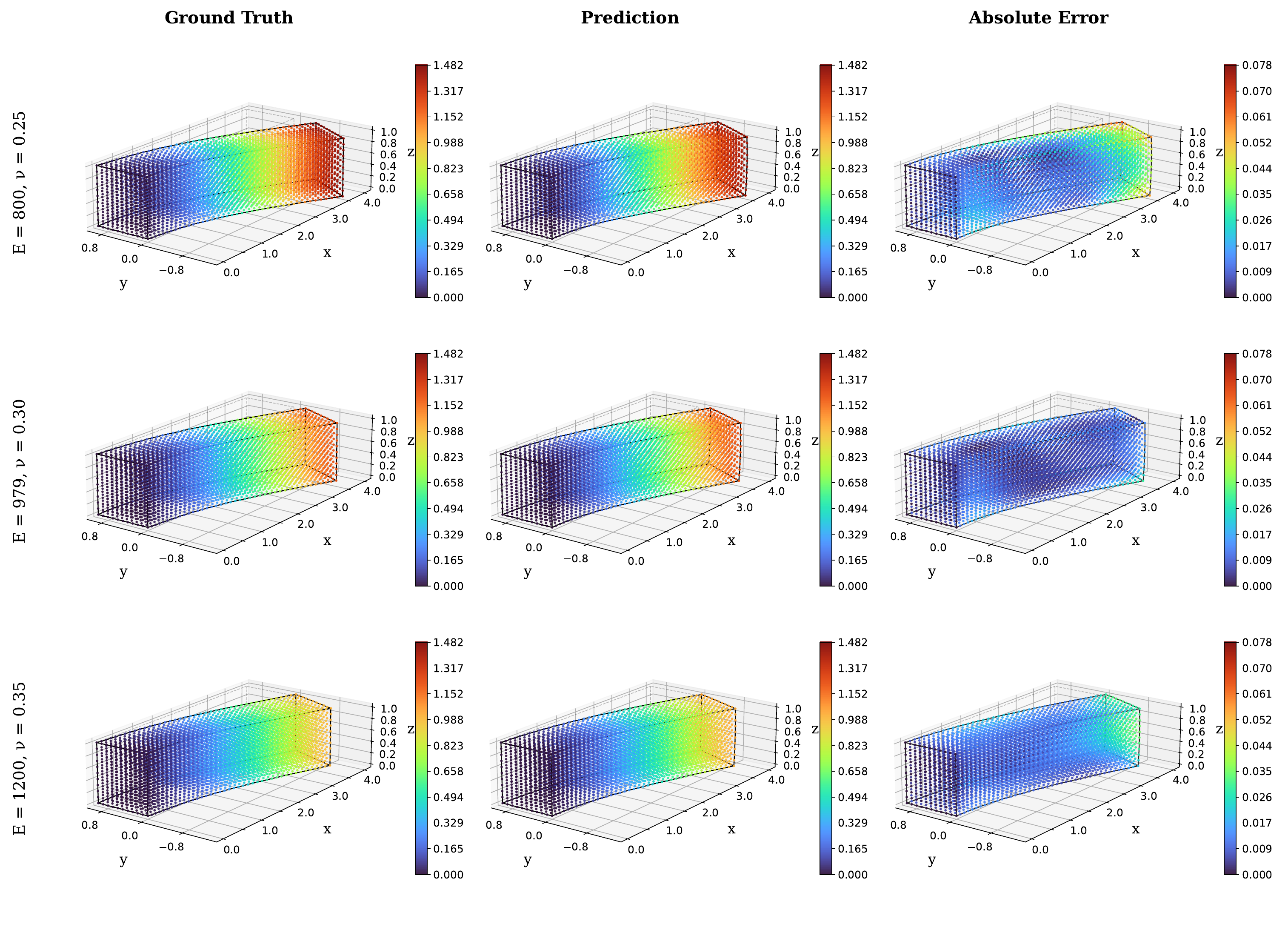}
    \par\vspace{-2pt}
    \caption{Comparison between FEM and CPDEM displacements for the 3-D Neo-Hookean cantilever in deformed coordinates. Rows use $(E,\nu)=(800,0.25)$, $(979,0.30)$, and $(1200,0.35)$; columns show FEM, CPDEM, and pointwise error ($\mathrm{m}$).}
    \label{fig:beam3d_neohook_u}
\end{figure}

Fig.~\ref{fig:beam3d_neohook_u} compares direct CPDEM predictions with FEM at three unseen in-range material parameter pairs spanning the $(E,\nu)$ domain.
As in the 2-D cantilever, the tip deflection stiffens steadily as $E$ and $\nu$ increase together, and CPDEM accurately reproduces the full 3-D deformed configuration across all three unseen material pairs.
In the most compliant case $(800,0.25)$, the peak displacement magnitude is $1.482\,\mathrm{m}$. The maximum pointwise absolute error reaches $0.078\,\mathrm{m}$, roughly $5\%$ of the peak displacement, along the loaded free-end edges where large 3-D bending rotations and strains develop.
As seen in the 2-D benchmarks, the discrepancy becomes much smaller for the interior pair $(979,0.30)$, remaining mostly below $0.025\,\text{m}$ throughout the beam, and stays moderate at the stiffer boundary corner $(1200,0.35)$ across the 3-D volume.

\begin{figure}[pos=htbp]
    \centering
    \includegraphics[width=0.35\textwidth]{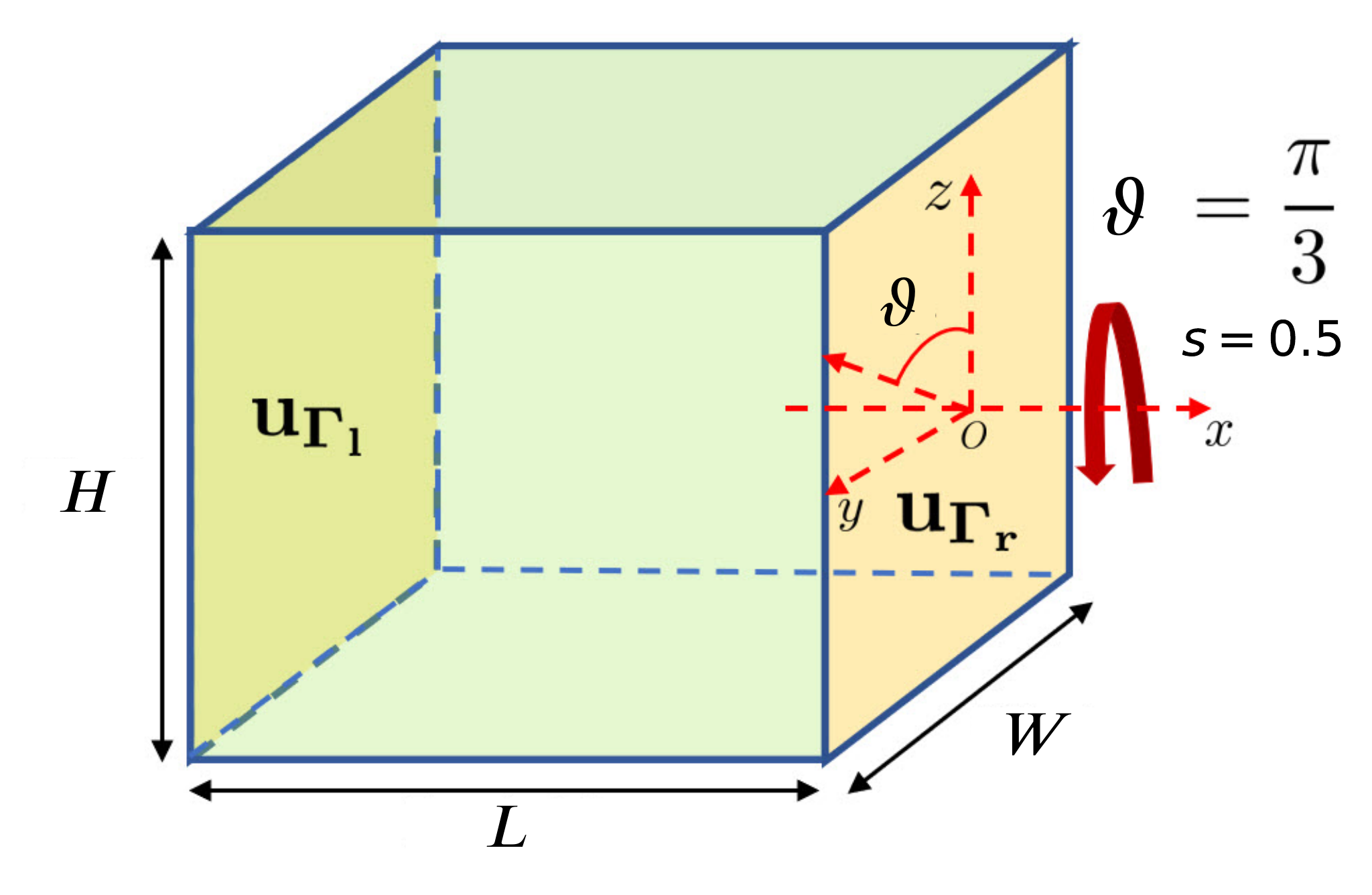}
    \par\vspace{-2pt}
    \caption{Geometry and boundary conditions of the 3-D Neo-Hookean cuboid under torsion: fixed left face $\Gamma_{\mathrm l}=\{\bm X\in\partial\Omega:X_1=0\}$ and imposed rotational displacement Eq.~\eqref{eq:cuboid3d-torsion-bc} on the right face $\Gamma_{\mathrm r}$.}
    \label{fig:3Dtorsion-cuboid}
\end{figure}

\textbf{3-D hyperelastic cuboid under torsion.}
Whereas the 2-D and 3-D beams involve purely traction-driven bending with a single clamped face, a more demanding test combines finite 3-D rotation, mixed Dirichlet boundaries, lateral tractions, and body forces.
As depicted in Fig.~\ref{fig:3Dtorsion-cuboid}, a cuboid of dimensions $L=1.25\,\mathrm{m}$ and $W=H=1\,\mathrm{m}$ is clamped on its left face $\Gamma_{\mathrm l}=\{\bm X\in\partial\Omega:X_1=0\}$, while its right face $\Gamma_{\mathrm r}=\{\bm X\in\partial\Omega:X_1=L\}$ undergoes a scaled rotational displacement ($\vartheta=\pi/3$, $s=0.5$) given by
\begin{equation}
    \label{eq:cuboid3d-torsion-bc}
    \bm{u}|_{\Gamma_{\mathrm r}} =
    s\begin{bmatrix}
        0 \\[2pt]
        0.5+(X_2-0.5)\cos\vartheta-(X_3-0.5)\sin\vartheta-X_2 \\[2pt]
        0.5+(X_2-0.5)\sin\vartheta+(X_3-0.5)\cos\vartheta-X_3
    \end{bmatrix}.
\end{equation}
This centered end-face motion combines a $30^{\circ}$ rotation about the fixed cross-section center $O$ with a uniform in-plane contraction by $\sqrt{3}/2$, reaching a corner displacement magnitude of $0.354\,\mathrm{m}$.
Concurrently, the lateral faces experience a traction $\bar{\bm t}=[1,0,0]^{\top}\,\mathrm{Pa}$ and the volume carries a body force $\bm f_b=[0,-0.5,0]^{\top}\,\mathrm{N/m^3}$, with material parameters spanning $E\in[8\times10^{5},1.2\times10^{6}]\,\mathrm{Pa}$ and $\nu\in[0.25,0.35]$.
Because the imposed end-face motion is fixed while the interior deformation and reaction stresses vary with $(E,\nu)$, this setup tests how CPDEM balances hard and weak kinematic constraints: the clamp at $X_1=0$ is imposed strictly by the ansatz
\begin{equation}
    \widehat{\bm u}_\Theta(\bm X,\boldsymbol\eta)
    =X_1\bm z_\Theta(\bm X,\boldsymbol\eta),
    \label{eq:cuboid3d-dirichlet-ansatz}
\end{equation}
with $\bm X\in\mathbb R^3$, whereas the rotational displacement on $\Gamma_{\mathrm r}$ enters the objective through a boundary penalty alongside the lateral traction and body-force work.

\begin{figure}[pos=htbp]
    \centering
    \includegraphics[width=\textwidth]{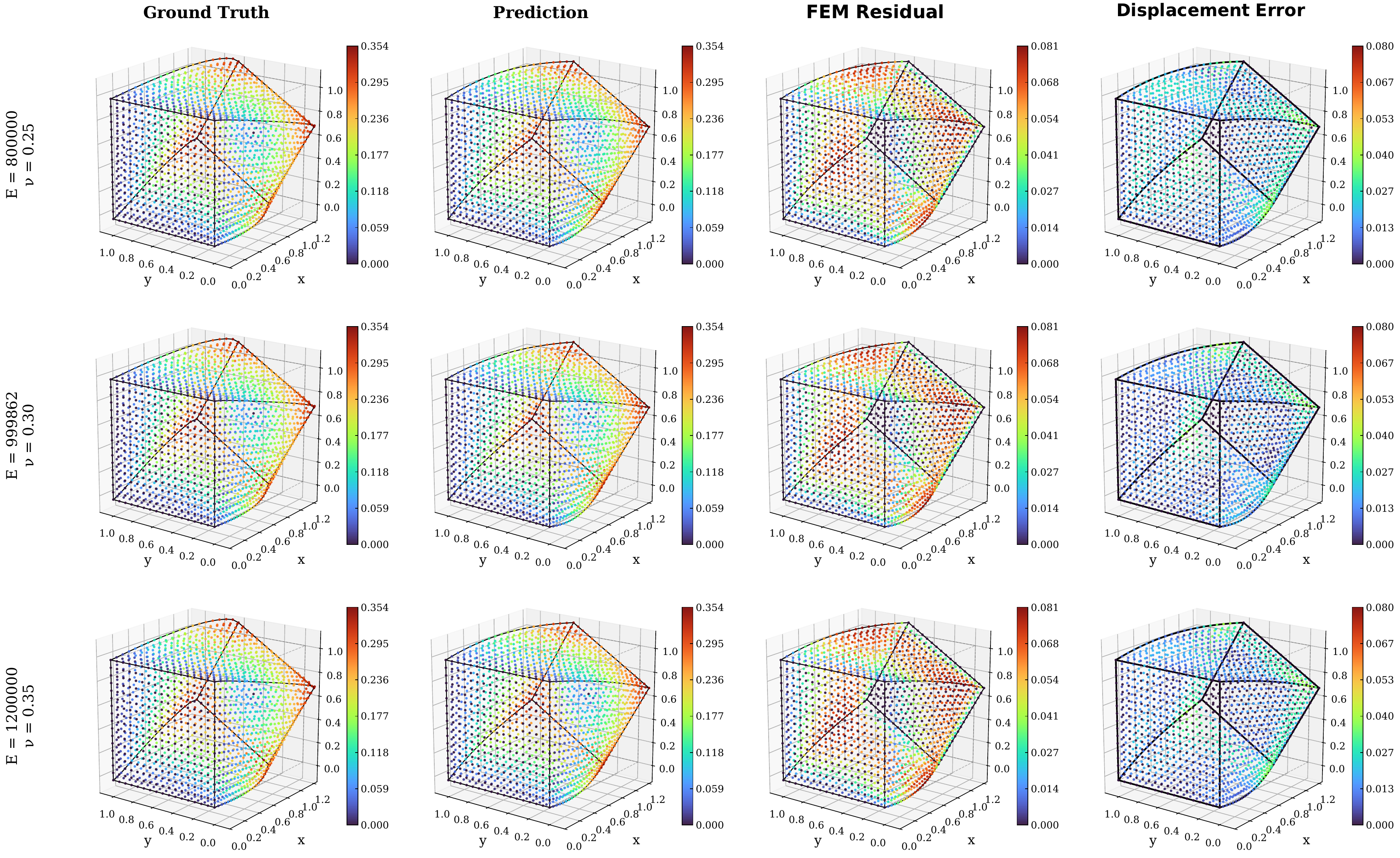}
    \par\vspace{-2pt}
    \caption{Comparison between FEM and CPDEM displacement magnitudes for 3-D hyperelastic torsion in deformed coordinates $(x,y,z)$. Rows correspond to $(E,\nu)=(8\times10^{5},0.25)$, $(9.99862\times10^{5},0.30)$, and $(1.2\times10^{6},0.35)$; columns show FEM, CPDEM, FEM nonlinear residual, and pointwise absolute error ($\mathrm{m}$).}
    \label{fig:cuboid3d_neohook_u}
\end{figure}

Fig.~\ref{fig:cuboid3d_neohook_u} compares the FEM and CPDEM fields at three unseen in-range material parameter pairs using shared column color scales: $0.354\,\mathrm{m}$ for displacement magnitude, $0.081\,\mathrm{m}$ for the FEM nonlinear residual relative to a linear axial extension of Eq.~\eqref{eq:cuboid3d-torsion-bc}, and $0.080\,\mathrm{m}$ for the pointwise Euclidean error $\|\widehat{\bm u}_{\Theta}(\bm X,\boldsymbol\eta)-\bm u_{\mathrm{FEM}}(\bm X,\boldsymbol\eta)\|_2$.
Because the imposed motion on $\Gamma_{\mathrm r}$ governs the overall displacement envelope, the displacement magnitude grows smoothly from the clamped face toward the twisted end with a similar global distribution across all three rows, while the FEM residual column isolates the finite-strain departure from purely linear twist along the lateral faces.
Over the full stiffness and compressibility range ($E\in[8\times10^{5},1.2\times10^{6}]\,\mathrm{Pa}$ and $\nu$ up to $0.35$), direct forward evaluation of the frozen pretrained weights predicts the 3-D torsional deformation without parameter-specific retraining. Interior discrepancies stay predominantly below $0.020\,\mathrm{m}$ with no degradation at larger $\nu$. Peak errors localize along the outer edges of the penalty-enforced loaded face $\Gamma_{\mathrm r}$, similar to the loaded-boundary error concentration seen in the cantilever benchmarks.
These three Neo-Hookean examples illustrate two practical numerical properties of CPDEM in finite elasticity.
First, whereas incremental nonlinear FEM relies on load stepping and repeated tangent-stiffness assembly to maintain Newton convergence under large rotations, CPDEM resolves the full parameter family directly through variational energy minimization without forming tangent matrices.
Second, exact trial-field constructions eliminate essential-boundary drift whenever the geometry permits this treatment, as in the hard-constrained cantilevers (Figs.~\ref{fig:beam2d_neohook_u} and~\ref{fig:beam3d_neohook_u}). In the mixed hard/penalty cuboid under torsion (Fig.~\ref{fig:cuboid3d_neohook_u}), weak penalty terms accommodate the more complex imposed end motion at the cost of slightly elevated discrepancies along the penalized face $\Gamma_{\mathrm r}$.

\subsection{Elastic-plasticity problems}
\label{sec:plas}

\begin{figure}[pos=htbp]
    \centering
    \includegraphics[width=0.72\textwidth]{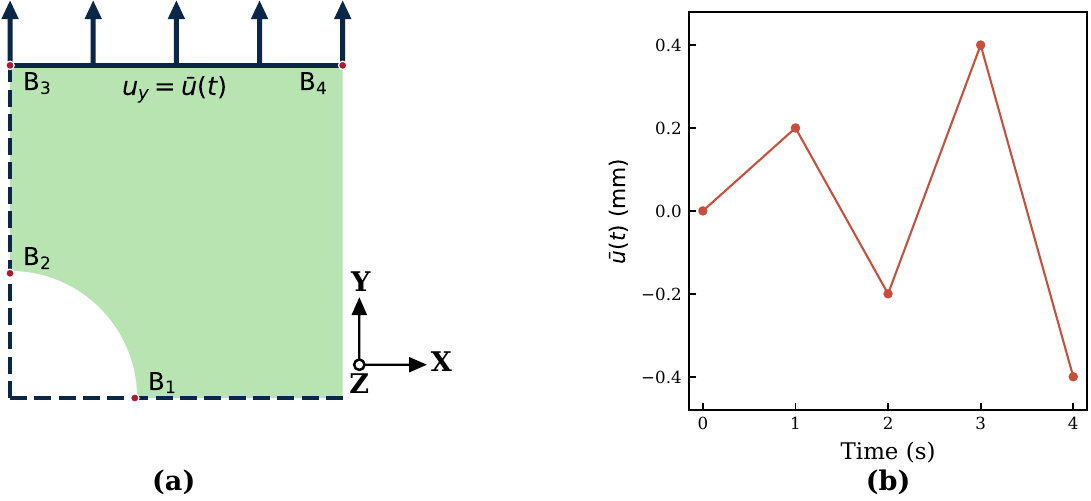}
    \par\vspace{-2pt}
    \caption{Geometry and loading of the perforated plate: (a) domain and monitors $B_1$--$B_4$; (b) cyclic loading $\bar u(t)$.}
    \label{fig:cyclic-plate-hole}
\end{figure}

We next examine irreversible, path-dependent elastoplastic deformation by considering a perforated plate subjected to cyclic loading across varying material parameters.
The full plate measures $8\,\mathrm{mm}\times8\,\mathrm{mm}\times2\,\mathrm{mm}$ with a central circular hole of radius $r_h=1.5\,\mathrm{mm}$.
Exploiting threefold symmetry, we model one-eighth of the plate (Fig.~\ref{fig:cyclic-plate-hole}), with boundary monitors $B_1$--$B_2$ placed at the hole-arc endpoints and $B_3$--$B_4$ at the upper-edge endpoints; the axes $(X,Y,Z)$ in Fig.~\ref{fig:cyclic-plate-hole}(a) map directly to $(x,y,z)$ on the reduced eighth-plate domain, with all spatial coordinates expressed in $\mathrm{mm}$.
The computational domain of the reduced plate is
\begin{equation}
    \label{eq:plate-hole-domain}
    \Omega_{\mathrm{ph}}=\{(x,y,z)\mid 0\le x\le4,\;0\le y\le4,\;0\le z\le1,\;x^2+y^2\ge r_h^2\}.
\end{equation}

Symmetry constraints are imposed on the three coordinate planes, while a cyclic vector displacement is applied on the upper boundary face $\Gamma_{y=4}$ as follows:
\begin{equation}
    \label{eq:plate-hole-bcs}
    \begin{alignedat}{2}
        u_x &= 0 &\quad& \text{on}\ \Gamma_{x=0},\\
        u_y &= 0 &\quad& \text{on}\ \Gamma_{y=0},\\
        u_z &= 0 &\quad& \text{on}\ \Gamma_{z=0},\\
        \bm{u} &= \bar{u}(t)\bm{e}_y &\quad& \text{on}\ \Gamma_{y=4},
    \end{alignedat}
\end{equation}
where the cyclic loading path in Fig.~\ref{fig:cyclic-plate-hole}(b) has five loading stations and an amplitude of $\bar u_{\max}=0.4\,\mathrm{mm}$,
\begin{equation}
    \label{eq:plate-hole-loading}
    \bar u(t_n)=\lambda_n\bar u_{\max},\qquad
    (\lambda_n)_{n=0}^{4}=(0,\,0.5,\,-0.5,\,1,\,-1).
\end{equation}
Fixing $u_y=\bar u(t)$ together with $u_x=u_z=0$ on $\Gamma_{y=4}$ enforces a full-vector kinematic constraint that induces a strongly non-homogeneous multiaxial stress state around the hole; the FEM reference applies the identical Dirichlet condition.
Over $\Omega_{\mathrm{ph}}$, the 3-D displacement field is written as
\begin{equation}
    \label{eq:plate-hole-displacement-field}
    \boldsymbol{u}(\boldsymbol{x},t_n)=
    \left[u_x(\boldsymbol{x},t_n),u_y(\boldsymbol{x},t_n),u_z(\boldsymbol{x},t_n)\right]^{\mathrm{T}},
    \quad \boldsymbol{x}\in\Omega_{\mathrm{ph}}.
\end{equation}
To compare spatial fields on the mid-plane section $z=0$, we evaluate the in-plane displacement magnitude $\bm u_\parallel=(u_x,u_y)^\top$,
\begin{equation}
    \label{eq:plate-hole-displacement-magnitude}
    \|\bm u_\parallel\|_2=\sqrt{u_x^2+u_y^2},
\end{equation}
and the pointwise absolute difference between the CPDEM and FEM in-plane displacement magnitudes,
\begin{equation}
    \label{eq:plate-hole-displacement-error}
    e_u(\bm x,t_n)=
    \left|\|\bm u_\parallel^{\mathrm{CPDEM}}(\bm x,t_n)\|_2
    -\|\bm u_\parallel^{\mathrm{ref}}(\bm x,t_n)\|_2\right|.
\end{equation}

\begin{figure}[pos=htbp]
    \centering
    \includegraphics[width=\textwidth]{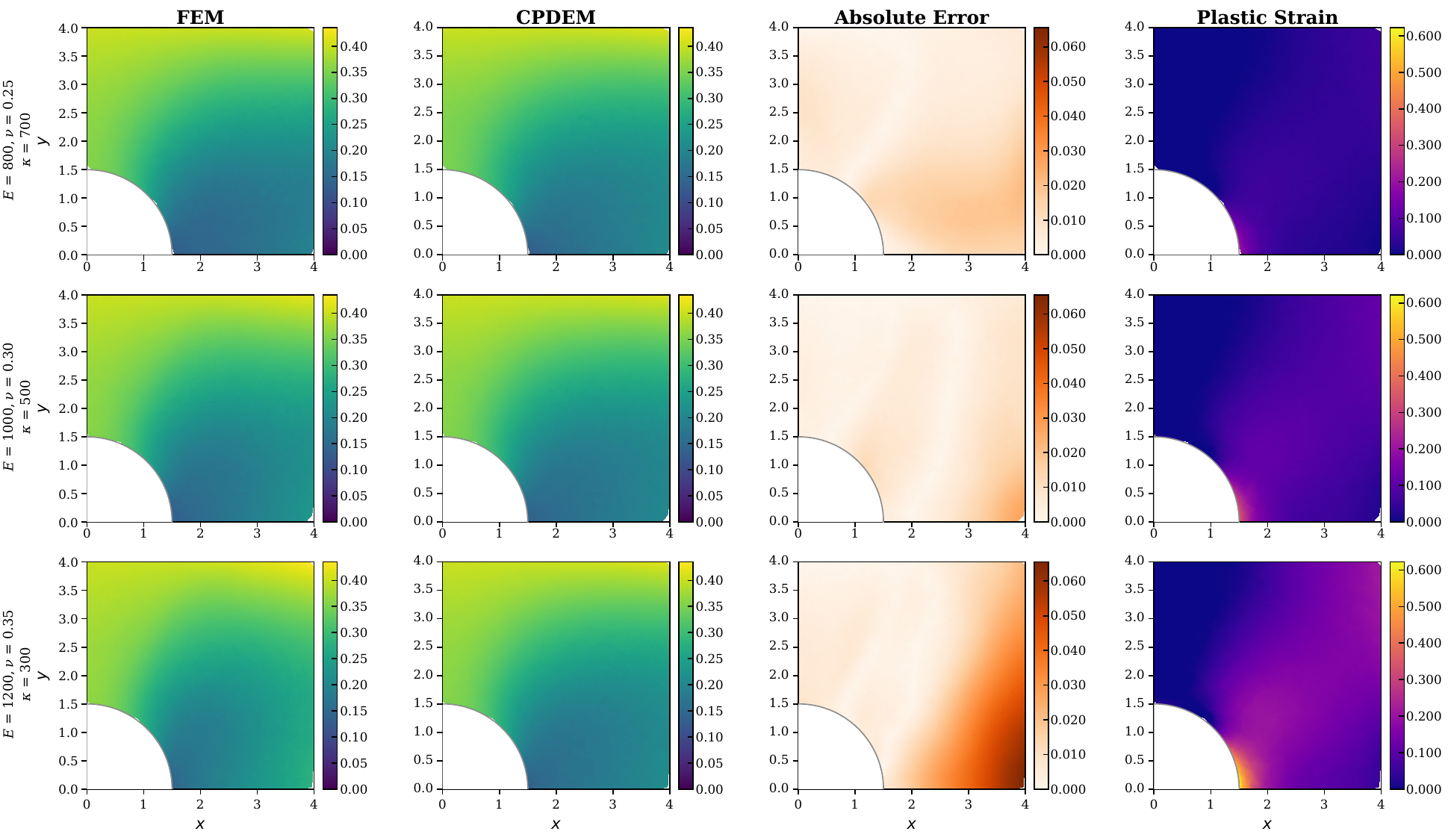}
    \par\vspace{-2pt}
    \caption{Comparison between FEM and CPDEM for the perforated plate at step 4 ($\bar u=-0.4\,\mathrm{mm}$). Rows use cases (a)--(c); columns show FEM and CPDEM in-plane displacements, pointwise error ($\mathrm{mm}$), and FEM plastic strain $\bar\varepsilon^p$.}
    \label{fig:plate-hole-result}
\end{figure}

\begin{figure}[pos=htbp]
    \centering
    \includegraphics[width=0.84\textwidth]{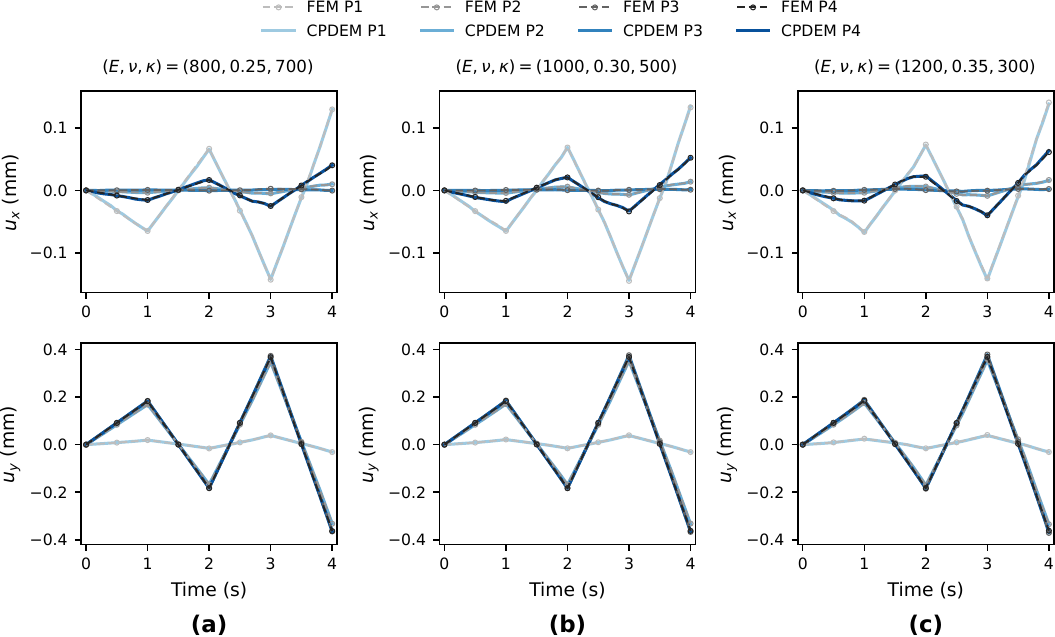}
    \par\vspace{-2pt}
    \caption{Comparison between FEM and CPDEM cyclic displacements at monitors $P_1$--$P_4$: $u_x$ (top) and $u_y$ (bottom).}
    \label{fig:plastic-displacement-history}
\end{figure}

\begin{figure}[pos=htbp]
    \centering
    \includegraphics[width=0.84\textwidth]{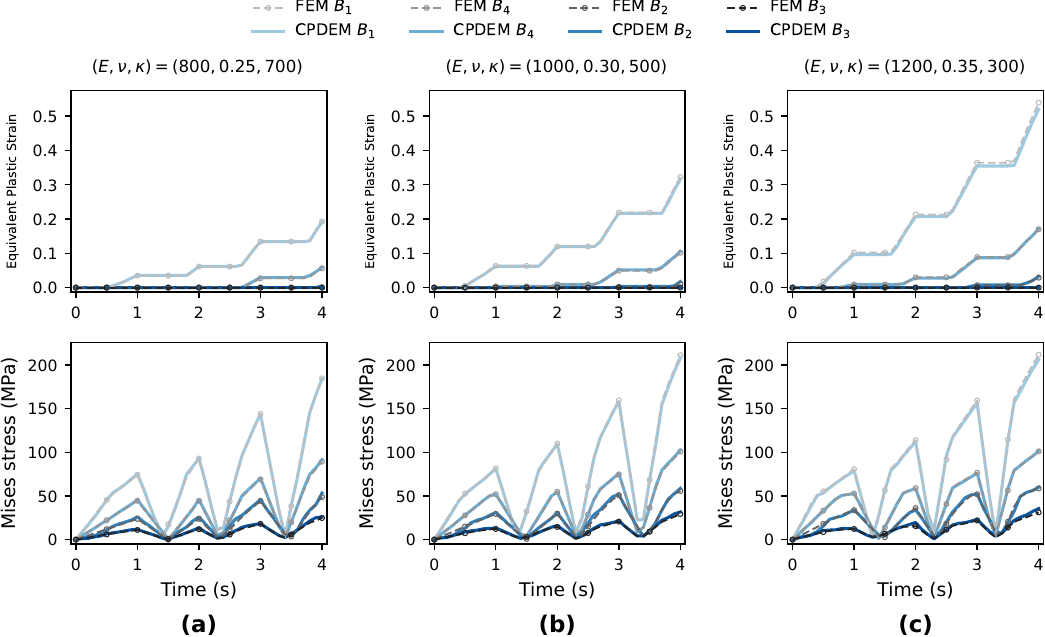}
    \par\vspace{-2pt}
    \caption{Comparison between FEM and CPDEM cyclic response histories at boundary monitors $B_1$--$B_4$: equivalent plastic strain $\bar\varepsilon^p$ (top) and von Mises stress (bottom).}
    \label{fig:plastic-strain-stress-history}
\end{figure}

The plate dimensions and hole radius are fixed across the material cases.
The material vector $\boldsymbol\eta=(E,\nu,\kappa)$ spans $\mathcal P_{\mathrm{ep}}=[800,1200]\,\mathrm{MPa}\times[0.25,0.35]\times[300,700]\,\mathrm{MPa}$.
The isotropic hardening modulus $\kappa$ sets the slope of the yield stress with respect to equivalent plastic strain in Eq.~\eqref{eq:plastic-hardening-law}.
The supplementary cyclic-history test uses an initial yield stress of $\sigma_{y,0}=50\,\mathrm{MPa}$.
Three unseen in-range material cases (a)--(c) use $(E,\nu,\kappa)=(800,0.25,700)$, $(1000,0.30,500)$, and $(1200,0.35,300)$, respectively, with $E$ and $\kappa$ in $\mathrm{MPa}$.
To assess both global kinematics and localized yielding, we first examine the full-field snapshot at peak compressive reversal (Fig.~\ref{fig:plate-hole-result}). The cyclic response across load reversals is examined through the interior displacement trajectories (Fig.~\ref{fig:plastic-displacement-history}) and the evolution of equivalent plastic strain and von Mises stress at boundary monitors $B_1$--$B_4$ (Fig.~\ref{fig:plastic-strain-stress-history}).

At step 4 of Fig.~\ref{fig:cyclic-plate-hole}(b) under the full-vector constraint in Eq.~\eqref{eq:plate-hole-bcs}, Fig.~\ref{fig:plate-hole-result} compares CPDEM and FEM in-plane displacement fields for cases (a)--(c).
Against a maximum displacement magnitude of roughly $0.45\,\mathrm{mm}$, the differences in cases (a) and (b) are largely below $0.02\,\mathrm{mm}$.
In case (c), which combines the largest $E$ and $\nu$ with the smallest $\kappa$, the equivalent plastic strain exceeds $0.50$ near the hole rim at $(x,y)=(1.5,0)\,\mathrm{mm}$ and the peak displacement error reaches about $0.06\,\mathrm{mm}$ near the lower-right region.

We assess the predicted response along the full cyclic loading path, where errors in the internal state can accumulate across load reversals (Eq.~\eqref{eq:analysis-history-error-propagation}).
Fig.~\ref{fig:plastic-displacement-history} compares the full displacement histories over $t\in[0,4]\,\mathrm{s}$ in a supplementary test where only $u_y$ is imposed on the upper boundary via the piecewise-linear sequence $0$, $0.2$, $-0.2$, $0.4$, and $-0.4\,\mathrm{mm}$ at $t=0,1,2,3,4\,\mathrm{s}$.
Four interior monitors, $P_1=(1.596,0.157,0.500)$, $P_2=(1.603,2.996,0.500)$, $P_3=(0.987,3.252,0.500)$, and $P_4=(2.996,3.645,0.500)$ in millimeters, probe the near-hole, transition, background, and near-loaded-edge zones, respectively.
At $P_2$--$P_4$, the vertical histories $u_y$ closely follow the imposed boundary amplitude (up to $\pm 0.37\,\mathrm{mm}$), whereas $P_1$ near the symmetry plane $y=0$ remains within $\pm 0.04\,\mathrm{mm}$.
The horizontal component $u_x$ is more sensitive to spatial location, with lateral reversals of up to $\pm0.14\,\mathrm{mm}$ at the unconstrained hole rim near $P_1$. The effect of the material constants is evident at $P_4$, where the amplitude increases steadily across the three material combinations from about $0.04\,\mathrm{mm}$ in case (a) to $0.06\,\mathrm{mm}$ in case (c).
For all three unseen in-range material sets, the CPDEM displacement histories agree well with FEM in both amplitude and reversal phase at all four interior monitors across the cycle without noticeable drift.

To examine the evolution of the internal variables, Fig.~\ref{fig:plastic-strain-stress-history} plots equivalent plastic strain and von Mises stress along the same cyclic path at the boundary stations $B_1$--$B_4$.
Consistent with the fourth column of Fig.~\ref{fig:plate-hole-result}, plastic flow concentrates strongly at the hole-rim monitor $B_1$, where cumulative $\bar\varepsilon^p$ at $t=4\,\mathrm{s}$ rises from about $0.20$ in case (a) to $0.32$ in case (b) and $0.54$ in case (c). Moderate secondary yielding develops at the upper-right corner $B_4$ ($\bar\varepsilon^p\approx 0.06$--$0.17$), whereas $B_2$ and $B_3$ remain nearly elastic throughout the cycle.
The histories distinguish irreversible plastic accumulation from elastic unloading, with $\bar\varepsilon^p$ increasing in discrete yielding steps separated by flat unloading plateaus. At $B_1$, the von Mises stress drops nearly to zero at each load reversal and rises across successive peaks owing to isotropic cyclic hardening, from about $75$--$80\,\mathrm{MPa}$ at $t=1\,\mathrm{s}$ to about $185$--$210\,\mathrm{MPa}$ at $t=4\,\mathrm{s}$.
CPDEM captures both the plastic-strain plateaus and the cyclic stress envelope across all three material sets, with small discrepancies confined mainly to the final high-strain peak at $B_1$ in the lowest-hardening case (c).
This cyclic agreement tests the checkpoint-based incremental scheme in Algorithm~\ref{alg:cpdem-incremental} under repeated loading and unloading reversals.
In path-dependent $J_2$ plasticity, any strain-gradient error at an earlier increment perturbs the trial stress entering the radial-return update and can propagate into irreversible drift across subsequent reversals (Eq.~\eqref{eq:analysis-history-error-propagation}).
\enlargethispage{6pt}
Separating the material-shared incremental displacement network from parameter-specific internal-state storage preserves consistent yielding transitions and elastic-unloading plateaus across varying $(E,\nu,\kappa)$ combinations without parameter-specific retraining. The local peak discrepancy at $B_1$ arises in the narrow elastic--plastic transition zone where smooth global neural bases encounter the steepest spatial strain gradients.

\subsection{Viscoelasticity problems}
\label{sec:visco}

Whereas $J_2$ plasticity is rate-independent, viscoelastic solids exhibit explicit time dependence driven by internal viscous dissipation.
We evaluate this four-parameter constitutive regime across three complementary loading and geometric configurations: traction-controlled tensile creep of a 2-D beam, displacement-controlled stress relaxation of the same beam, and external-pressure creep buckling of a curved circular shell.
In the incremental model, $\mathcal{L}_{ve}^{n+1}$ in Eq.~\eqref{eq:viscoelastic-incremental-potential} governs backward-Euler viscous updates with $\mathbb{L}^{\infty}$, $\mathbb{L}^{1}$, and $\mathbb{M}$ parameterized by $\boldsymbol{\eta}=[E_{\infty},E_{1},\xi,\nu]^{\top}$ ($\tau=\xi/E_{1}$), while the snapshot and history tests below note their effective-modulus or shear-relaxation forms explicitly.

\begin{figure}[pos=htbp]
    \centering
    \includegraphics[width=0.65\textwidth]{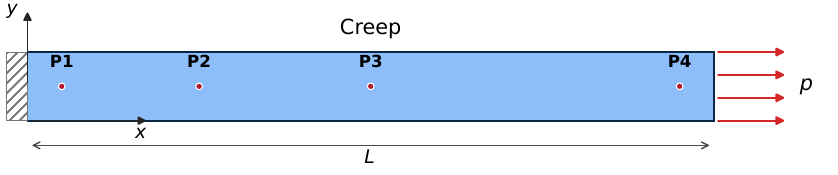}
    \par\vspace{-2pt}
    \caption{Geometry and loading of the 2-D viscoelastic creep cantilever ($L=10\,\mathrm{m}$, $h=1\,\mathrm{m}$), clamped on the left edge with constant tensile traction $p=0.4\,\mathrm{Pa}$ on the right edge for $t\ge 0$.}
    \label{fig:visco-creep}
\end{figure}

\textbf{Viscoelastic creep of a 2-D beam.}
The 2-D plane-strain cantilever beam in Fig.~\ref{fig:visco-creep} occupies $\Omega_{\mathrm{cr}}=\{(x,y)\mid 0\le x\le L,\;0\le y\le h\}$, with $L=10\,\mathrm{m}$ and $h=1\,\mathrm{m}$.
For a clamped left edge and a constant horizontal tensile traction $p$ on the right edge for $t\ge 0$, the boundary conditions are
\begin{equation}
    \label{eq:visco-creep-bcs}
    \begin{array}{l}
        \bm{u}(0,y,t) = \mathbf{0} \quad \text{on}\ \Gamma_{x=0},\\
        \bm{\sigma}(L,y,t)\cdot\bm{e}_{x} = p\,\bm{e}_{x} \quad \text{on}\ \Gamma_{x=L},\quad t\ge 0,\\
        \bm{\sigma}\cdot\bm{n} = \mathbf{0} \quad \text{on}\ \Gamma_{y=0}\cup\Gamma_{y=h}.
    \end{array}
\end{equation}
To separate kinematic enforcement from traction equilibration, the clamped root at $x=0$ is imposed strictly by the trial ansatz
\begin{equation}
    \label{eq:visco-creep-ansatz}
    \begin{split}
        \hat{u}_{x}(x,y,t;\boldsymbol{\eta}) &= x\,\hat{z}_{x}(x,y,t;\boldsymbol{\eta}),\\
        \hat{u}_{y}(x,y,t;\boldsymbol{\eta}) &= x\,\hat{z}_{y}(x,y,t;\boldsymbol{\eta}),
    \end{split}
\end{equation}
where $(\hat z_x,\hat z_y)$ are the components of the raw network output $\bm z_\Theta(\bm X,t;\boldsymbol\eta)$, while the right-edge traction enters variationally through the external potential $\int_{\Gamma_{x=L}} p\,u_{x}\,d\Gamma$ in Eq.~\eqref{eq:viscoelastic-incremental-potential}.
Advancing along a uniform time grid ($\Delta t=1.0\,\mathrm{s}$) couples spatial quadrature of $\mathcal{L}_{ve}^{n+1}$ with backward-Euler viscous integration using consistent material parameters at each increment. Time-step refinement isolates temporal discretization error from neural approximation error. The FEM reference solves the identical constant-traction history for direct comparison.

The four-dimensional parameter domain is $\mathcal P_{\mathrm{ve}}=[2,6]\,\mathrm{Pa}\times[1,10]\,\mathrm{Pa}\times[10,60]\,\mathrm{Pa\cdot s}\times[0.20,0.40]$, with nominal point $(5\,\mathrm{Pa},3\,\mathrm{Pa},50\,\mathrm{Pa\cdot s},0.30)$. The instantaneous modulus is $E_0=E_\infty+E_1$, and the relaxation modulus is $E(t)=E_\infty+E_1\exp(-tE_1/\xi)$. The supplementary effective-modulus benchmark uses $\tau=\xi/E_0$ and defines the three parameter sets in terms of $E_0$ as
\begin{equation}
    \label{eq:visco-creep-realisations}
    \begin{aligned}
        (E_0,E_\infty,\xi,\nu)^{(a)} &= (10\,\mathrm{Pa},\,5\,\mathrm{Pa},\,50\,\mathrm{Pa\!\cdot\!s},\,0.35),\quad \tau^{(a)}=5\,\mathrm{s},\\
        (E_0,E_\infty,\xi,\nu)^{(b)} &= (10\,\mathrm{Pa},\,2\,\mathrm{Pa},\,50\,\mathrm{Pa\!\cdot\!s},\,0.25),\quad \tau^{(b)}=5\,\mathrm{s},\\
        (E_0,E_\infty,\xi,\nu)^{(c)} &= (10\,\mathrm{Pa},\,2\,\mathrm{Pa},\,10\,\mathrm{Pa\!\cdot\!s},\,0.25),\quad \tau^{(c)}=1\,\mathrm{s}.
    \end{aligned}
\end{equation}
This pairing is deliberately structured: cases (a) and (b) share $\tau=5\,\mathrm{s}$ while varying the long-term stiffness $E_\infty$ and Poisson's ratio $\nu$, whereas cases (b) and (c) share all elastic constants and differ solely in viscosity $\xi$.
Each snapshot uses $E_{\mathrm{eff}}(t)=E_\infty+(E_0-E_\infty)\exp(-t/\tau)$, and field accuracy is measured at common spatial points via the absolute horizontal displacement error
\begin{equation}
    \label{eq:visco-creep-error}
    e_{u_x}(\bm{x},t;\boldsymbol{\eta})=
    \left|u_{x,\mathrm{CPDEM}}(\bm{x},t;\boldsymbol{\eta})-u_{x,\mathrm{ref}}(\bm{x},t;\boldsymbol{\eta})\right|.
\end{equation}

\begin{figure}[pos=htbp]
    \centering
    \includegraphics[width=\textwidth]{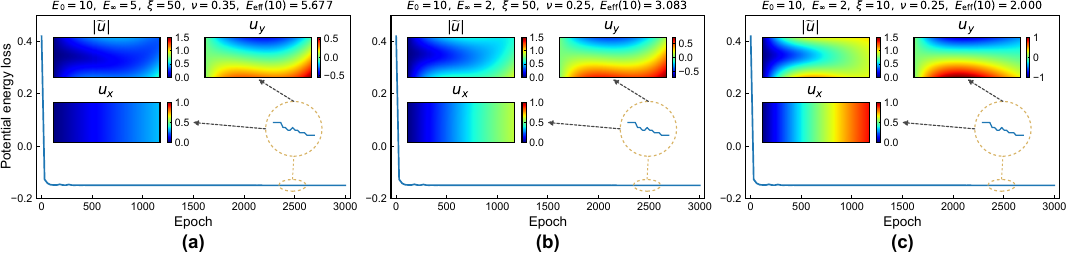}
    \par\vspace{-2pt}
    \caption{Training loss histories (rescaled) for 2-D viscoelastic creep. Panels (a)--(c) correspond to the material sets in Eq.~\eqref{eq:visco-creep-realisations}; insets show normalized $u_x$, $u_y$, and approximate magnitude $|\widetilde{u}|$ at $t=10\,\mathrm{s}$.}
    \label{fig:visco-creep-train}
\end{figure}

\begin{figure}[pos=htbp]
    \centering
    \includegraphics[width=\textwidth]{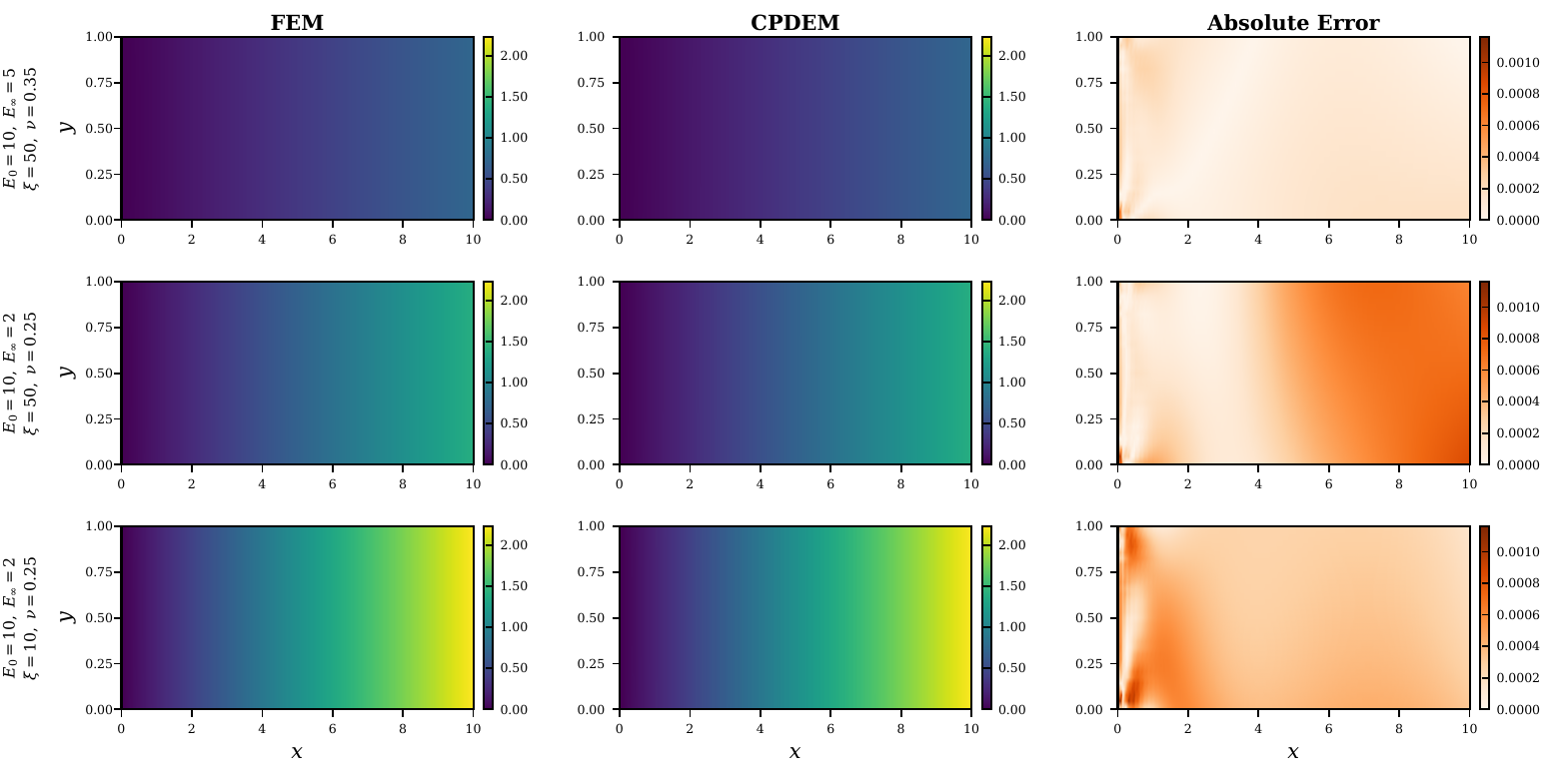}
    \par\vspace{-2pt}
    \caption{Comparison between FEM and CPDEM horizontal displacements $u_x$ for viscoelastic creep at $t=10\,\mathrm{s}$. Rows correspond to the material sets in Eq.~\eqref{eq:visco-creep-realisations}; columns show FEM, CPDEM, and pointwise absolute error $e_{u_x}$ ($\mathrm{m}$).}
    \label{fig:visco-creep-result}
\end{figure}

Fig.~\ref{fig:visco-creep-train} shows the convergence of the shared training loss, with epochs 2190--2700 enlarged on the rescaled axis in all three panels. At $t=10\,\mathrm{s}$, the three material cases exhibit axial elongation toward the loaded edge together with a smaller transverse contraction.
Fig.~\ref{fig:visco-creep-result} compares the converged $u_x$ fields with FEM at $t=10\,\mathrm{s}$ and illustrates the separate roles of long-term stiffness and relaxation time. Case (a) ($E_\infty=5\,\mathrm{Pa}$) remains the stiffest with free-end elongation near $0.70\,\mathrm{m}$. Case (b) ($E_\infty=2\,\mathrm{Pa}$, $\tau=5\,\mathrm{s}$) reaches about $1.30\,\mathrm{m}$ because viscous relaxation is still incomplete ($t/\tau=2$), whereas case (c) ($E_\infty=2\,\mathrm{Pa}$, $\tau=1\,\mathrm{s}$) has already relaxed to its long-term compliance ($t/\tau=10$) and stretches to roughly $2.10\,\mathrm{m}$.
Pointwise absolute errors remain below $1\times 10^{-3}\,\mathrm{m}$ (under $0.1\%$ of peak elongation) in all three rows: errors stay well below $4\times 10^{-4}\,\mathrm{m}$ in case (a) and increase only mildly near the clamped root corners and loaded tip in the two larger-strain creep cases (b) and (c).

\begin{figure}[pos=htbp]
    \centering
    \includegraphics[width=0.86\textwidth]{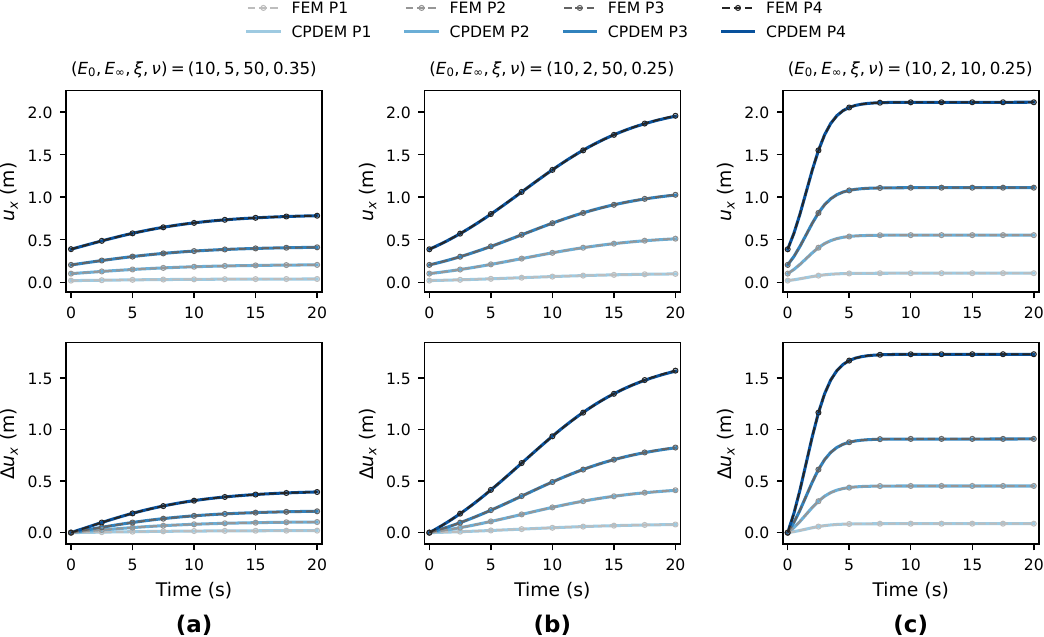}
    \par\vspace{-2pt}
    \caption{Comparison between FEM and CPDEM creep histories at $P_1$--$P_4$: $u_x$ (top) and increment $\Delta u_x$ (bottom).}
    \label{fig:creep-displacement-increment-history}
\end{figure}

To examine how viscosity governs the transient evolution toward equilibrium beyond a single snapshot, Fig.~\ref{fig:creep-displacement-increment-history} plots the plane-stress effective-modulus response over $t\in[0,20]\,\mathrm{s}$ under $p=0.4\,\mathrm{Pa}$. The four midline monitors are $P_1=(0.5,0.5)$, $P_2=(2.5,0.5)$, $P_3=(5.0,0.5)$, and $P_4=(9.5,0.5)$, with coordinates in metres, spanning from near the clamp to near the loaded tip without lying on either end boundary.
Subtracting the instantaneous elastic response gives the cumulative creep increment $\Delta u_x(t)=u_x(t)-u_x(0^+)$, separating the time-dependent deformation from the initial elastic stretch.
Because all three cases share the instantaneous modulus $E_0=10\,\mathrm{Pa}$, they start from essentially the same elastic stretch at $t=0^+$, with $u_x(0^+)\approx0.40\,\mathrm{m}$ at $P_4$. Both $u_x$ and $\Delta u_x$ vary almost linearly with axial position from $P_1$ to $P_4$, consistent with nearly uniform uniaxial creep strain away from the clamp.
Over the $20\,\mathrm{s}$ window, case (a) approaches $u_x\approx 0.80\,\mathrm{m}$ ($\Delta u_x\approx 0.40\,\mathrm{m}$ at $P_4$). With the same elastic constants ($E_0$, $E_\infty$, and $\nu$), cases (b) and (c) tend toward the same long-term asymptote ($u_x\approx 2.10\,\mathrm{m}$, $\Delta u_x\approx 1.70\,\mathrm{m}$). However, the lower-viscosity case (c) ($\tau=1\,\mathrm{s}$) reaches the plateau by $t\approx 6\,\mathrm{s}$, whereas case (b) ($\tau=5\,\mathrm{s}$) climbs gradually to $u_x\approx 1.95\,\mathrm{m}$ ($\Delta u_x\approx 1.55\,\mathrm{m}$) without flattening out by $t=20\,\mathrm{s}$.
The CPDEM histories agree closely with the FEM transient growth rates and equilibrium amplitudes at all four midline monitors for all three seen parameter sets.

\begin{figure}[pos=htbp]
    \centering
    \includegraphics[width=0.56\textwidth]{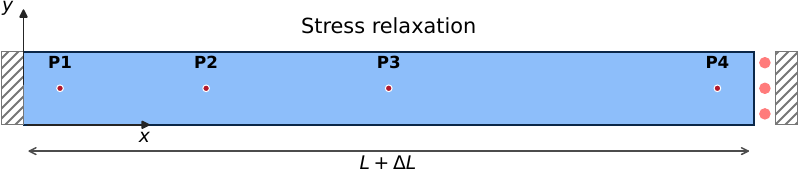}
    \par\vspace{-2pt}
    \caption{Geometry and loading of the 2-D viscoelastic stress-relaxation cantilever ($L=10\,\mathrm{m}$, $h=1\,\mathrm{m}$), clamped on the left edge with imposed elongation $\Delta L=0.1\,\mathrm{m}$ on the right edge for $t\ge 0$.}
    \label{fig:visco-relax}
\end{figure}

\textbf{Stress relaxation of a 2-D beam.}
We next replace the constant right-edge traction with an imposed axial elongation to test stress relaxation on the same domain $\Omega_{\mathrm{cr}}$ (Fig.~\ref{fig:visco-relax}) and midline monitors ($y=0.5\,\mathrm{m}$, $x=0.5, 2.5, 5.0, 9.5\,\mathrm{m}$).
Here the right edge is displaced by $\Delta L=0.1\,\mathrm{m}$ at $t=0$ (nominal axial strain $\Delta L/L=0.01$) and held fixed thereafter:
\begin{equation}
    \label{eq:visco-relax-bcs}
    \begin{split}
        \bm{u}(0,y,t) &= \mathbf{0} \quad \text{on}\ \Gamma_{x=0},\\
        u_{x}(L,y,t) &= \Delta L,\quad u_{y}(L,y,t)=0 \quad \text{on}\ \Gamma_{x=L},\quad t\ge 0,\\
        \bm{\sigma}\cdot\bm{n} &= \mathbf{0} \quad \text{on}\ \Gamma_{y=0}\cup\Gamma_{y=h}.
    \end{split}
\end{equation}
Because both ends now carry Dirichlet constraints, the trial field embeds both boundaries simultaneously,
\begin{equation}
    \label{eq:visco-relax-ansatz}
    \begin{split}
        \hat{u}_{x}(x,y,t;\boldsymbol{\eta}) &= \hat{z}_{x}(x,y,t;\boldsymbol{\eta})\,x\,(x-L)+\frac{\Delta L}{L}\,x,\\
        \hat{u}_{y}(x,y,t;\boldsymbol{\eta}) &= \hat{z}_{y}(x,y,t;\boldsymbol{\eta})\,x\,(x-L),
    \end{split}
\end{equation}
so that with traction-free lateral edges the incremental potential in Eq.~\eqref{eq:viscoelastic-incremental-potential} reduces purely to bulk free energy and dissipation, minimized at $\Delta t=1.0\,\mathrm{s}$ while end reactions are recovered from the stress field.

\begin{figure}[pos=htbp]
    \centering
    \includegraphics[width=\textwidth]{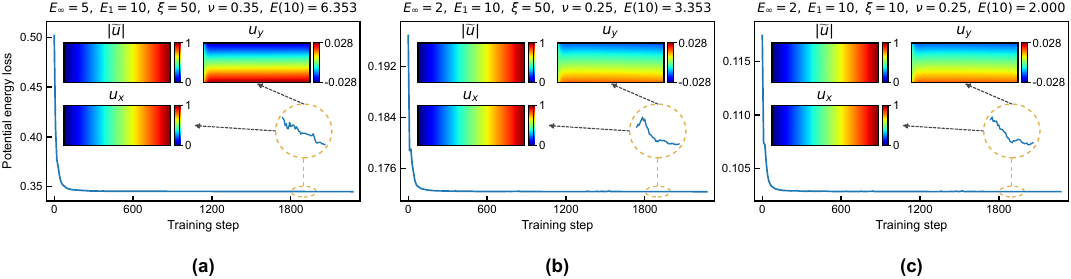}
    \par\vspace{-2pt}
    \caption{Training energy histories per unit thickness ($\mathrm{J/m}$) for 2-D stress relaxation. Panels (a)--(c) correspond to $(E_\infty,E_1,\xi,\nu)=(5,10,50,0.35)$, $(2,10,50,0.25)$, and $(2,10,10,0.25)$; insets show $|\widetilde{u}|$, $u_x$, and $u_y$ at $t=10\,\mathrm{s}$ ($\Delta L=1\,\mathrm{m}$).}
    \label{fig:visco-relax-train}
\end{figure}

\begin{figure}[pos=htbp]
    \centering
    \includegraphics[width=\textwidth]{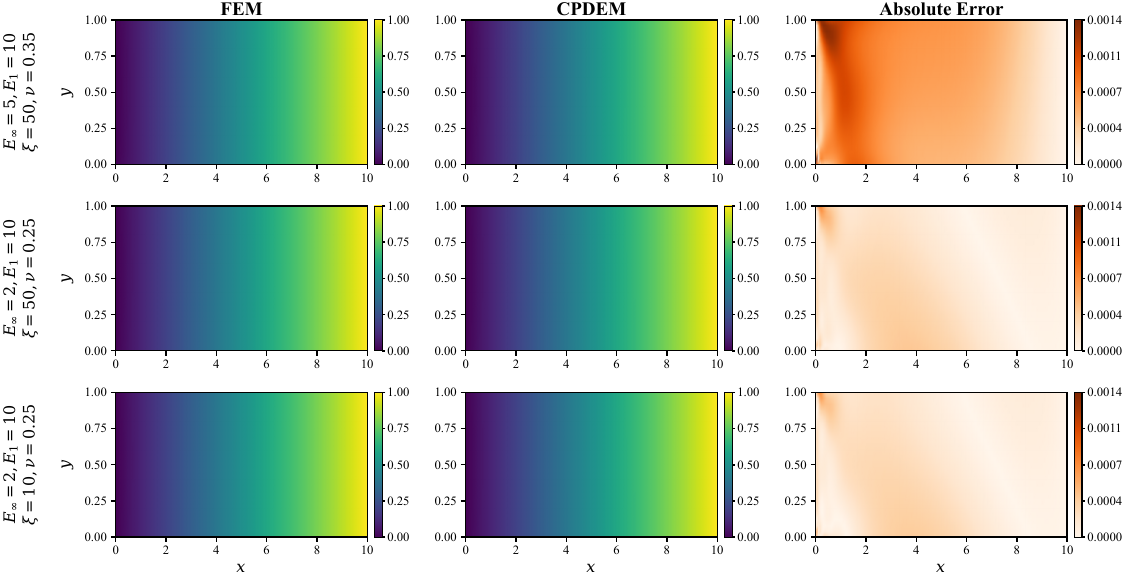}
    \caption{Comparison between FEM and CPDEM $u_x$ in stress relaxation at $t=10\,\mathrm{s}$ ($\Delta L=1\,\mathrm{m}$). Rows use $(E_\infty,E_1,\xi,\nu)=(5,10,50,0.35)$, $(2,10,50,0.25)$, and $(2,10,10,0.25)$; columns show FEM, CPDEM, and error ($\mathrm{m}$).}
    \label{fig:visco-relax-result}
\end{figure}

We first assess full-field spatial accuracy in a supplementary plane-strain snapshot at $t=10\,\mathrm{s}$ under $\Delta L=1\,\mathrm{m}$, combining the Neo-Hookean energy with the relaxation modulus $E(t)=E_\infty+E_1\exp(-t/\tau)$ ($\tau=\xi/E_1$) for $(E_\infty,E_1,\xi,\nu)=(5,10,50,0.35)$, $(2,10,50,0.25)$, and $(2,10,10,0.25)$ (moduli in $\mathrm{Pa}$, viscosity in $\mathrm{Pa\!\cdot\!s}$).
The training energies in Fig.~\ref{fig:visco-relax-train} decrease rapidly before leveling off. The enlargements show the minor fluctuations over training steps 1880--2271 in all three panels.
Compared against an independently equilibrated FEM reference on a $160\times24$ mesh in Fig.~\ref{fig:visco-relax-result}, CPDEM achieves maximum absolute $u_x$ errors of $1.44\times10^{-3}$, $8.06\times10^{-4}$, and $7.54\times10^{-4}\,\mathrm{m}$. For the same three cases, the relative $L^2$ errors of the full displacement vector are $0.154\%$, $0.076\%$, and $0.083\%$, respectively.
In this fixed-extension snapshot, the three relaxation moduli at $t=10\,\mathrm{s}$ are $6.353$, $3.353$, and $2.000\,\mathrm{Pa}$. Because the scalar modulus rescales only the energy and stress levels, the equilibrium displacement field depends solely on $\nu$. Consequently, the last two parameter sets produce identical FEM displacement fields at distinct energy levels, and CPDEM recovers the same decoupled response.

\begin{figure}[pos=htbp]
    \centering
    \includegraphics[width=0.86\textwidth]{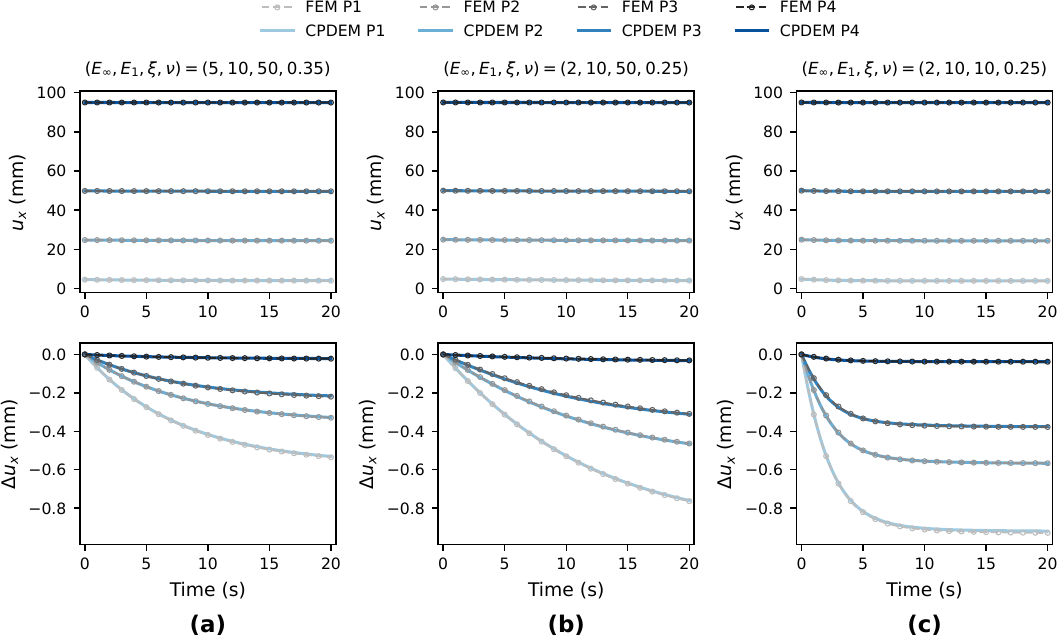}
    \caption{Comparison between FEM and CPDEM relaxation histories at $P_1$--$P_4$: $u_x$ (top) and increment $\Delta u_x$ (bottom).}
    \label{fig:relaxation-ux-increment-history}
\end{figure}

Unlike that single-modulus snapshot, a full hereditary standard-linear-solid model with constant bulk modulus $K=(E_\infty+E_1)/[3(1-2\nu)]$ and relaxing shear modulus $G(t)=[E_\infty+E_1\exp(-t/\tau)]/[2(1+\nu)]$ causes the effective Poisson's ratio to increase over time, driving internal displacement redistribution even under fixed elongation.
Fig.~\ref{fig:relaxation-ux-increment-history} compares the four-point histories over $0$--$20\,\mathrm{s}$ under $\Delta L=0.1\,\mathrm{m}$ ($100\,\mathrm{mm}$) for these three unseen in-range material parameter sets using the energy-trained CPDEM model.
The total displacement $u_x$ remains close to the linear axial profile ($\approx 5, 25, 50, 95\,\mathrm{mm}$ at $P_1$--$P_4$). Subtracting $u_x(0^+)$, however, reveals negative submillimeter redistribution ($\Delta u_x<0$) along the midline. This redistribution is largest at $P_1$ near the clamped root where transverse contraction is constrained, reaching about $-0.53$, $-0.77$, and $-0.92\,\mathrm{mm}$ at $t=20\,\mathrm{s}$ in cases (a)--(c), respectively, and decays smoothly to below $0.04\,\mathrm{mm}$ at $P_4$.
Between cases (a) and (b), a smaller equilibrium-to-relaxing modulus ratio amplifies the redistribution magnitude. Between cases (b) and (c), the shorter shear relaxation time ($\tau=1\,\mathrm{s}$ versus $5\,\mathrm{s}$) drives case (c) to a plateau of $-0.92\,\mathrm{mm}$ by $t\approx 6\,\mathrm{s}$, while case (b) continues relaxing through $t=20\,\mathrm{s}$. CPDEM agrees with FEM on both the submillimeter amplitudes and the relaxation rates across all four monitoring points.
In the traction-controlled creep benchmark, the four-parameter material encoder represents large-amplitude creep governed by $t/\tau$. In the displacement-controlled relaxation benchmark, the same encoder and variational structure represent the smaller internal displacement redistribution driven by the time-dependent relaxation of the shear modulus relative to the bulk modulus.

\begin{figure}[pos=htbp]
    \centering
    \includegraphics[width=.86\textwidth]{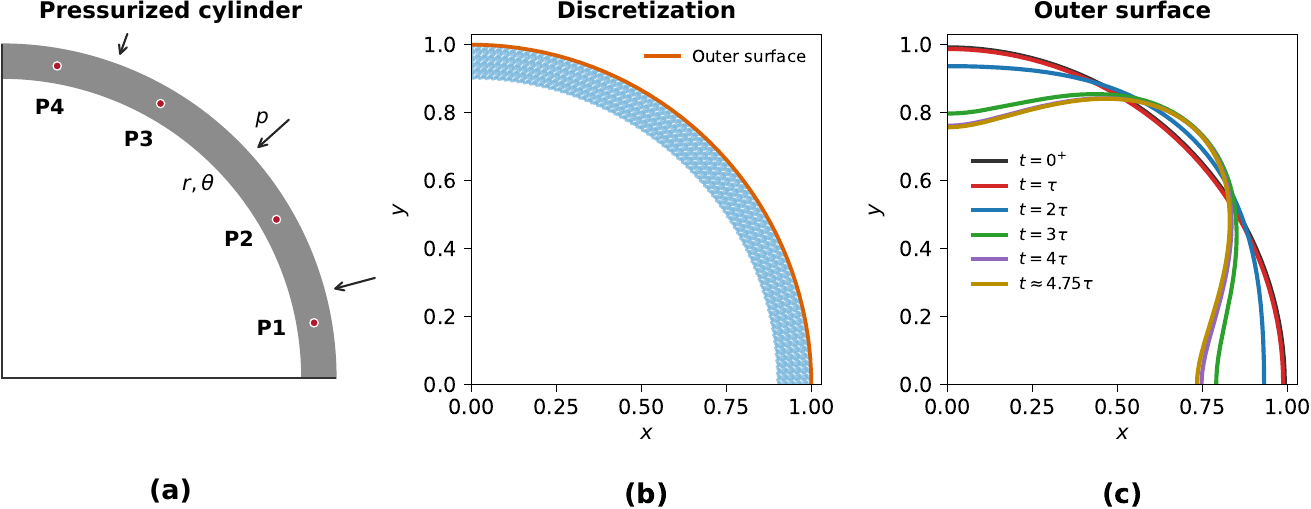}
    \par\vspace{-2pt}
    \caption{Geometry and loading of the quarter annular shell under external pressure: (a) geometry and pressure loading; (b) collocation points and outer surface; (c) outer-surface profiles at $t=0^+$ and multiples of $\tau$.}
    \label{fig:circle-creep-buckling}
\end{figure}

\begin{figure}[pos=htbp]
    \centering
    \includegraphics[width=\textwidth]{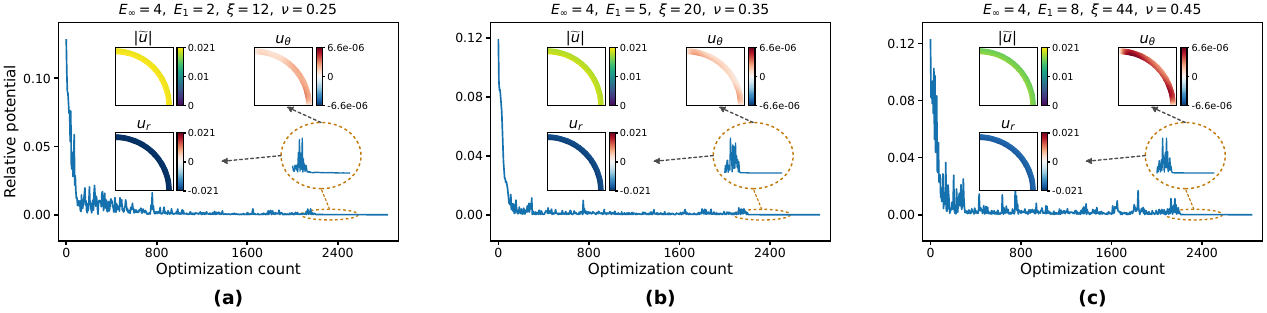}
    \par\vspace{-2pt}
    \caption{Training energy histories for the quarter annular shell in order of increasing $E_1$. Insets show the displacement magnitude and the radial and circumferential displacement components at $t=5\tau$.}
    \label{fig:circle-creep-buck-train}
\end{figure}

\begin{figure}[pos=htbp]
    \centering
    \includegraphics[width=0.82\textwidth]{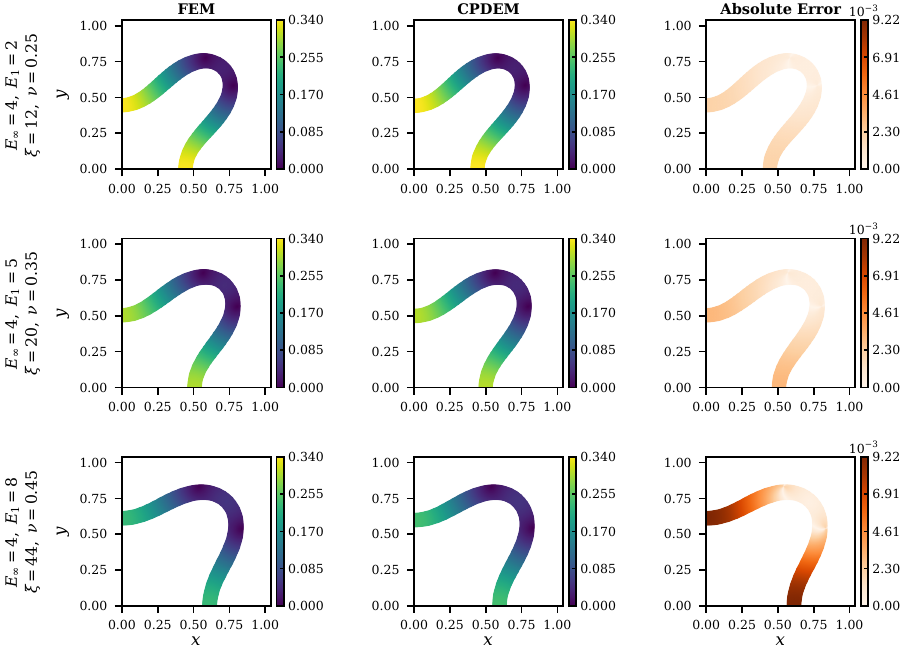}
    \par\vspace{-2pt}
    \caption{Comparison between FEM and CPDEM postbuckling displacement magnitudes for the quarter annular shell at $t=4\tau$ and $p=0.011\,\mathrm{Pa}$. Columns show FEM, CPDEM, and pointwise error on the FEM configuration ($\mathrm{m}$).}
    \label{fig:circle-creep-buck-result}
\end{figure}

\begin{figure}[pos=htbp]
    \centering
    \includegraphics[width=0.86\textwidth]{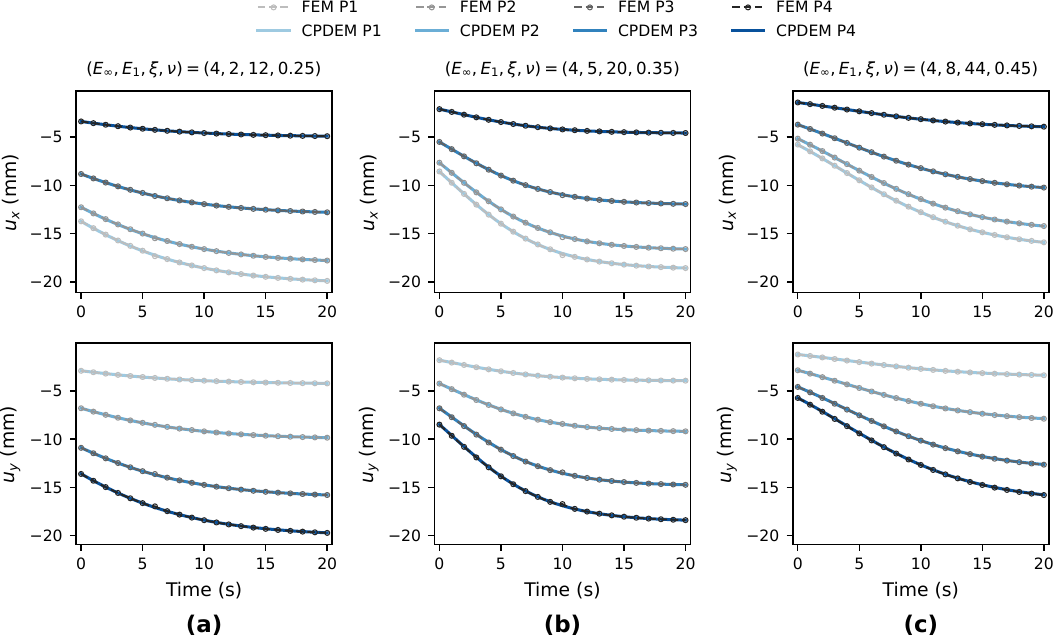}
    \par\vspace{-2pt}
    \caption{Comparison between FEM and CPDEM displacements at $P_1$--$P_4$ on the quarter shell: $u_x$ (top) and $u_y$ (bottom).}
    \label{fig:circle-displacement-history}
\end{figure}

\begin{figure}[pos=htbp]
    \centering
    \includegraphics[width=.88\linewidth]{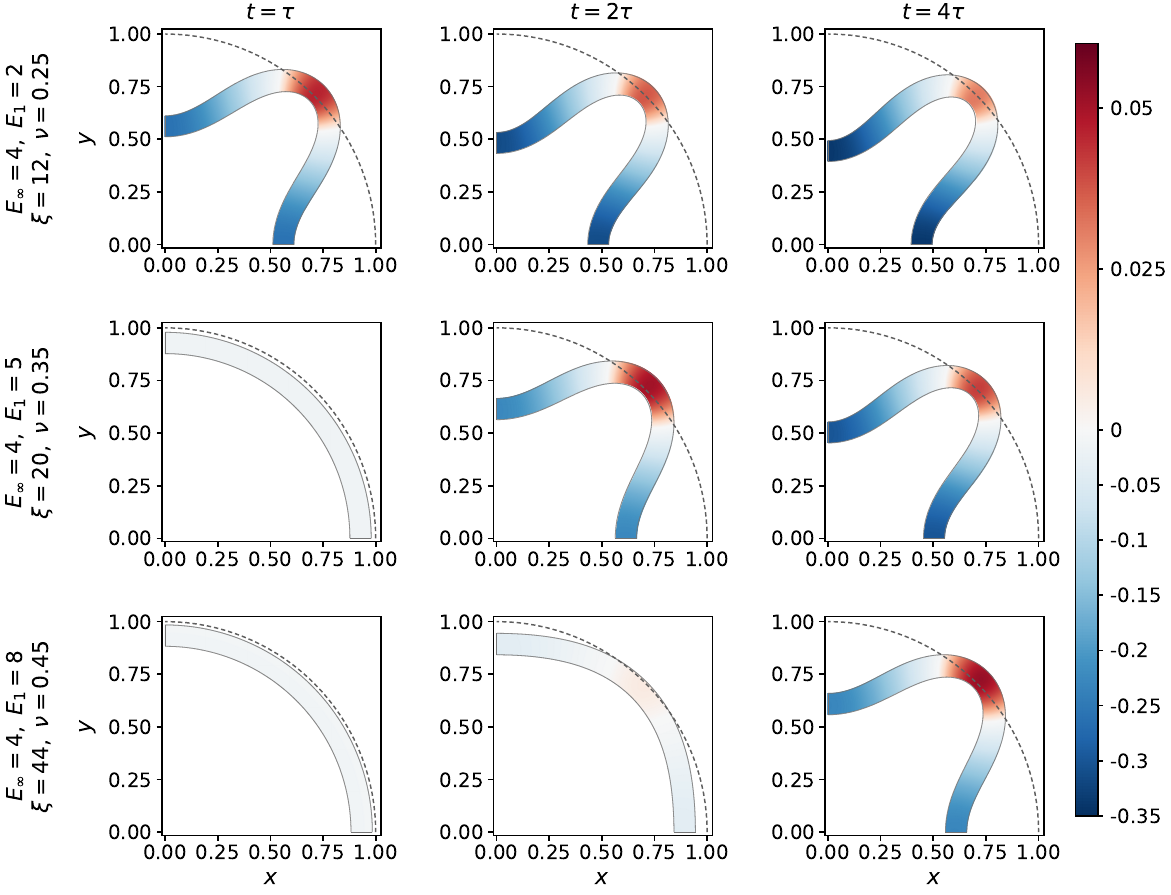}
    \par\vspace{-2pt}
    \caption{FEM postbuckling deformation of the quarter annular shell at $t/\tau=1$, $2$, and $4$ (columns) for $E_1=2$, $5$, and $8\,\mathrm{Pa}$ (rows). Color indicates signed radial displacement ($\mathrm{m}$); the gray dashed curve marks the undeformed outer surface.}
    \label{fig:circle-material-deformation}
\end{figure}

\textbf{Viscoelastic creep buckling of a 2-D circle.}
Finally, we extend the viscoelastic benchmark from straight beams to a curved structural geometry under external pressure: a quarter annular shell of outer radius $r_o=1\,\mathrm{m}$ and thickness $a=0.1\,\mathrm{m}$ loaded by uniform external pressure $p=10^{-2}\,\mathrm{Pa}$ over $\theta\in[0,\pi/2]$ (Fig.~\ref{fig:circle-creep-buckling}).
The material comparisons in Figs.~\ref{fig:circle-creep-buck-train}--\ref{fig:circle-material-deformation} use $E_\infty=4\,\mathrm{Pa}$ and the three material combinations $(E_1,\xi,\nu)=(2,12,0.25)$, $(5,20,0.35)$, and $(8,44,0.45)$, with $E_1$ in $\mathrm{Pa}$ and $\xi$ in $\mathrm{Pa\,s}$. Their relaxation times $\tau=\xi/E_1$ are $6$, $4$, and $5.5\,\mathrm{s}$, respectively. Four points $P_1$--$P_4$ on the mid-thickness arc at $r=0.95\,\mathrm{m}$ and $\theta=12^\circ,29^\circ,51^\circ,76^\circ$ monitor the displacement history.
For the $0.010\,\mathrm{Pa}$ calculations in Figs.~\ref{fig:circle-creep-buck-train} and~\ref{fig:circle-displacement-history}, the trial field combines the plane-strain Lam\'e solution for a pressurized annulus with a learned correction,
\begin{equation}
    \label{eq:circle-creep-theta-ansatz}
    \widehat u_i=X_i\left(\frac{u_r^{\mathrm{L}}}{r}+0.003\,z_{\Theta,i}\right),\qquad i=1,2,
\end{equation}
where $u_r^{\mathrm{L}}$ is the linear-elastic radial displacement evaluated with the effective modulus at time $t$, $r=\|\bm X\|_2$, and $z_{\Theta,i}$ are the Cartesian network outputs. The factors $X_i$ enforce zero normal displacement on the symmetry planes. The material encoder uses eight features derived from the four material parameters, and the coordinate encoder takes the two spatial coordinates together with the normalized time $t/(30\,\mathrm{s})$.

A shared network is trained by energy minimization on 24 material combinations, including the three displayed here. Fig.~\ref{fig:circle-creep-buck-train} tracks the potential of each displayed case at $t=5\tau$ during optimization. Each ordinate is the change relative to that case's final potential in units of $10^{-6}E_\infty r_o^2$ per unit thickness. The enlargements cover counts 2075--2580; the count includes Adam steps followed by L-BFGS objective evaluations.

Fig.~\ref{fig:circle-creep-buck-result} compares postbuckling fields at a pressure of $0.011\,\mathrm{Pa}$ and $t=4\tau$. To follow the mirror-symmetric branch, we train a separate CPDEM trial field by minimizing the potential energy over these three material states. The polar displacement components combine a neural correction with a Fourier expansion in angle and a cubic Chebyshev expansion through the thickness. The modal coefficients depend quadratically on $E_1$. An inward perturbation at both ends selects the bending mode, and the trial field enforces symmetry about the diagonal of the quarter annulus throughout training. FEM displacements enter only in the subsequent comparison.
Both methods predict that the ends move inward and the arc bends outward at midspan relative to the ends. Comparing displacement magnitudes at matching reference nodes gives peak differences of $2.14$, $3.13$, and $9.22\,\mathrm{mm}$ in order of increasing $E_1$, or $0.63\%$, $1.05\%$, and $4.00\%$ of the respective FEM peak magnitudes.

At the baseline pressure of $0.010\,\mathrm{Pa}$, Fig.~\ref{fig:circle-displacement-history} follows the four monitoring points over a common interval of $0$--$20\,\mathrm{s}$. Averaged over the four points, the FEM displacement magnitude increases from 14.03 to $20.33\,\mathrm{mm}$ in the first case, from 8.76 to $18.97\,\mathrm{mm}$ in the second, and from 5.91 to $16.27\,\mathrm{mm}$ in the third. The largest CPDEM--FEM displacement-vector difference across the four points and all sampled times is $0.113$, $0.170$, and $0.027\,\mathrm{mm}$, respectively.

To trace the development of the bent profiles in Fig.~\ref{fig:circle-creep-buck-result}, Fig.~\ref{fig:circle-material-deformation} shows the signed radial displacement on the same mirror-symmetric FEM branch at three normalized times. At $t=2\tau$, the difference between the largest and smallest radii of the deformed outer surface is $0.347$, $0.273$, and $0.042\,\mathrm{m}$ for the three material combinations in order of increasing $E_1$. The first shell bends most strongly, while the third retains a nearly circular profile. By $t=4\tau$, both ends have moved inward and the middle of the arc bulges outward relative to the ends in all three cases. At this time, the maximum height of the outer profile above its left endpoint is $0.149$, $0.122$, and $0.074\,\mathrm{m}$, respectively.

Across Sections~\ref{sec:elasticity}--\ref{sec:visco}, the same dual-encoder variational formulation handles linear-elastic, finite-strain hyperelastic, cyclic $J_2$ elastoplastic, and four-parameter viscoelastic benchmarks without tangent-matrix assembly or per-query retraining.
A consistent parametric and spatial error pattern appears across all four constitutive regimes: interior parameter sets generally attain lower discrepancies than extreme domain corners, while local residual errors concentrate in high-gradient zones such as loaded beam tips, the yielded hole rim, and penalty-enforced boundary faces.
After offline pretraining, CPDEM predicts the response at unseen in-range material parameters with the trained weights held fixed. For history-dependent problems, the local constitutive update advances the internal state at each increment. The fusion network can be fine-tuned when greater accuracy is needed for an individual query (Section~\ref{sec:tune}).

\section{Conclusions}
\label{conclusions}

In this work, we developed the \textbf{C}onstitutive \textbf{P}arameterized \textbf{D}eep \textbf{E}nergy \textbf{M}ethod (\textbf{CPDEM}) to solve solid mechanics problems across continuous material-parameter domains. The method applies to linear, nonlinear, and history-dependent regimes.
CPDEM couples separate coordinate and material encoders through a shared fusion network to construct a unified displacement trial field. Training minimizes the expected potential energy using only first-order spatial derivatives and without labeled FEM solutions.
For uniformly stable linear-elastic families, product-space analysis establishes Lipschitz continuity of the parameter-to-solution map. The expected energy gap is equivalent to the squared parameter-averaged mechanical error, which justifies using energy minimization to control that error.
The finite-training bounds separate the contributions of approximation, parameter sampling, quadrature, boundary enforcement, and optimization to the error. Local stability analysis extends the framework to finite-strain hyperelasticity and incremental history-dependent problems. In the latter, parameter-specific constitutive updates govern internal-state evolution and error propagation.

Numerical benchmarks across linear elasticity, finite-strain Neo-Hookean hyperelasticity, cyclic $J_2$ elastoplasticity, and standard-linear-solid viscoelasticity corroborate these theoretical properties against FEM references.
CPDEM reproduces the main 2-D and 3-D elastic and hyperelastic bending and torsion fields at unseen parameters. These predictions require no tangent stiffness assembly. For the perforated elastoplastic plate, CPDEM captures cyclic displacement reversals. The predictions also capture plastic-strain accumulation near the hole and stress drops during elastic unloading.
In the four-parameter viscoelastic beam benchmarks, the same variational formulation captures transient creep and internal displacement redistribution during stress relaxation. For the externally pressurized quarter annular shell, CPDEM captures baseline transient deformation and the mirror-symmetric postbuckling branch.
Single-parameter deep energy solvers and repeated finite-element analyses solve a new problem for each material set. CPDEM shares the cost of offline training across the parameter space. New in-range queries use direct evaluation of the frozen network. The fusion network can be fine-tuned for queries that require higher accuracy.

Several limitations of CPDEM remain to be addressed.
First, the neural energy minimization problem is non-convex, and offline training can cost more than a single classical FEM solve. With global smooth bases, residual errors concentrate near loaded boundaries, sharp corners, and steep plastic-strain gradients. Adaptive collocation, higher-order quadrature, or mixed stress--displacement energy formulations could improve local accuracy.
Second, the present formulation varies finite-dimensional material parameters while the geometry and loading remain fixed. In postbuckling, the formulation follows a selected symmetry branch. Future work includes joint geometric--constitutive parameterization and automated tracking of multiple bifurcation modes. Further extensions would address more general nonlinear hardening or anisotropic constitutive laws.
Overall, constitutive parameterization within energy minimization provides a label-free basis for multi-query simulation, stochastic response analysis, and inverse material identification.

\section*{Acknowledgment}
The authors gratefully acknowledge support from the National Natural Science Foundation of China under Grant~12572138 for the development and numerical evaluation of the proposed method.

\section*{Funding}
This work was supported by the National Natural Science Foundation of China [grant number 12572138].
The funder had no role in study design, data collection and analysis, decision to publish, or preparation of this article.

\printcredits

\section*{Data and code availability}
The material parameters, loads, and boundary conditions for the numerical examples are given in this article.
The CPDEM source code is publicly available at \mbox{\url{https://github.com/liangzhangyong/CPDEM.git}}.

\section*{Declaration of competing interests}
The authors declare that they have no known competing financial interests or personal relationships that could have appeared to influence the work described in this paper.

\bibliographystyle{cas-model2-names}
\bibliography{cas-refs}

\end{document}